%% file: paper.tex
\documentclass[twocolumn]{svjour3}

\usepackage{multirow}
\usepackage{appendix}
\usepackage[normalem]{ulem}

\input{preamble}
\input{abbrev}

\begin{document}

%%%%% TITLE
\title{Understanding and Improving Kernel Local Descriptors}

\author{Arun Mukundan \and Giorgos Tolias \and Andrei Bursuc \and Herv\'e J\'egou \and Ond{\v r}ej Chum}
\institute{A. Mukundan,
G.Tolias ,
O. Chum \at VRG, FEE, CTU in Prague
\\\email{{arun.mukundan,giorgos.tolias,chum@felk.cvut.cz}} \and
A. Bursuc \at Valeo.ai
\\\email{andrei.bursuc@valeo.com}\and
H. J\'egou \at  Facebook AI Research
\\\email{rvj@fb.com}}

\maketitle
\thispagestyle{empty}

%%%%%%%%% ABSTRACT
\begin{abstract}
We propose a multiple-kernel local-patch descriptor based on efficient match
kernels from pixel gradients. It combines two parametrizations of gradient
position and direction, each parametrization provides robustness to a different
type of patch mis-registration: polar parametrization for noise in the patch
dominant orientation detection, Cartesian for imprecise location of the feature
point.
Combined with whitening of the descriptor space, that is learned with or without supervision,
the performance is significantly improved.
We analyze the effect of the whitening on patch similarity and demonstrate its semantic meaning.
Our unsupervised variant is the best performing descriptor constructed without the need of labeled data.
Despite the simplicity of the proposed descriptor, it competes well with deep learning
approaches on a number of different tasks.

\end{abstract}

%%%%%%%%% BODY TEXT

\input{intro}
\input{related}
\input{preliminaries}

\input{method}

\input{experiments}

\section{Conclusions}

We have proposed a multiple-kernel local-patch descriptor combining two
parametrizations of gradient position and direction. Each parametrization
provides robustness to a different type of patch miss-registration: polar
parametrization for noise in the dominant orientation, Cartesian for imprecise
location of the feature point.
We have performed descriptor whitening and have shown that its effect on patch
similarity is semantically meaningful.
The lessons learned from analyzing the similarity after whitening can be exploited
for further improvements of kernel, or even CNN-based, descriptors.
Learning the whitening in a supervised or unsupervised way boosts the performance.
Interestingly, the latter generalizes better and sets the best so far performing hand-crafted
 descriptor that is competing well even with CNN-based descriptors.
Unlike the currently best performing CNN-based approaches, the proposed descriptor is easy to implement and interpret.

\section{Acknowledgments}
The authors were supported by the OP VVV funded project CZ.02.1.01/0.0/0.0/16\_019/0000765 ``Research Center
for Informatics'' and MSMT LL1303 ERC-CZ grant. Arun Mukundan was supported by the CTU student grant SGS17/185/OHK3/3T/13.
We would like to thank Karel Lenc and Vassileios Balntas for their valuable help with the HPatches benchmark,
and Dmytro Mishkin for providing the L2Net and HardNet descriptors for the HPatches dataset.

\input{nomixparam}

{\small
\bibliographystyle{spmpsci}
\bibliography{egbib}
}

\end{document}

%% file: preamble.tex
\usepackage{xcolor}
\usepackage{amsmath}
\usepackage{amsfonts}
\usepackage{booktabs}
\usepackage{tabularx,booktabs}
\usepackage{caption}
\usepackage{array}
\usepackage{graphicx}
\usepackage{nicefrac}
\usepackage{xspace}
\usepackage{relsize}
\usepackage{lmodern}

\usepackage{pgfplots}
\usepackage{pgfplotstable}
\usepackage{pgf,tikz}
\usetikzlibrary{shapes,arrows}
\usetikzlibrary{positioning}
\usetikzlibrary{decorations.pathreplacing}
\usetikzlibrary{decorations.markings}

\pgfplotsset{compat=newest}
\usetikzlibrary{plotmarks}
\usetikzlibrary{patterns}
\usepgfplotslibrary{patchplots}
\pgfplotsset{
	tick label style={font=\scriptsize},
	label style={font=\scriptsize},
	legend style={font=\scriptsize},
	title style={font=\scriptsize}}

\usepackage{scrextend}
\usepackage{hyperref}
\usepackage{bigfoot}
\usepackage{url}

\def\addlegendimage{\csname pgfplots@addlegendimage\endcsname}

\newcolumntype{L}{>{$}l<{$}}
\newcolumntype{C}{>{$}c<{$}}
\newcolumntype{R}{>{$}r<{$}}

\renewcommand\footnotemark{}

\makeatletter
\ifcase \@ptsize \relax% 10pt
  \newcommand{\miniscule}{\@setfontsize\miniscule{4}{5}}% \tiny: 5/6
\or% 11pt
  \newcommand{\miniscule}{\@setfontsize\miniscule{5}{6}}% \tiny: 6/7
\or% 12pt
  \newcommand{\miniscule}{\@setfontsize\miniscule{5}{6}}% \tiny: 6/7
\fi
\makeatother

%% file: abbrev.tex
\def\real{\mathbb{R}}

\def\l2{\ensuremath{\ell_2}\xspace}

\def \I{\mathbf{I}}

\def\des{KD\xspace}

\def\P{\ensuremath{\mathcal{P}}\xspace}
\def\Q{\ensuremath{\mathcal{Q}}\xspace}
\def\Pb{\ensuremath{\mathbb{P}}\xspace}
\def\Mb{\ensuremath{\mathbb{M}}\xspace}

\def\p{\ensuremath{p}\xspace}
\def\q{\ensuremath{q}\xspace}

\def\pm{\ensuremath{\p_m}\xspace}
\def\px{\ensuremath{\p_x}\xspace}
\def\py{\ensuremath{\p_y}\xspace}
\def\pt{\ensuremath{\p_\theta}\xspace}
\def\ptt{\ensuremath{\p_{\tilde{\theta}}}\xspace}
\def\pp{\ensuremath{\p_\phi}\xspace}
\def\pr{\ensuremath{\p_\rho}\xspace}
\def\pg{\ensuremath{\p_g}\xspace}

\def\qm{\ensuremath{\q_m}\xspace}
\def\qx{\ensuremath{\q_x}\xspace}
\def\qy{\ensuremath{\q_y}\xspace}
\def\qt{\ensuremath{\q_\theta}\xspace}
\def\qtt{\ensuremath{\q_{\tilde{\theta}}}\xspace}
\def\qp{\ensuremath{\q_\phi}\xspace}
\def\qr{\ensuremath{\q_\rho}\xspace}
\def\qg{\ensuremath{\q_g}\xspace}

\def\n{\ensuremath{\gamma}\xspace}

\def\k{\ensuremath{k}\xspace}
\def\kx{\ensuremath{k_x}\xspace}
\def\ky{\ensuremath{k_y}\xspace}
\def\kr{\ensuremath{k_\rho}\xspace}
\def\kt{\ensuremath{k_\theta}\xspace}
\def\ktt{\ensuremath{k_{\tilde{\theta}}}\xspace}
\def\kp{\ensuremath{k_\phi}\xspace}

\def\VM{\mathsf{VM}}
\def\V{\mathbf{V}}
\def\M{\mathcal{M}}

\def\map{\ensuremath{\psi}\xspace}

\def\mapx{\ensuremath{\psi_x}\xspace}
\def\mapy{\ensuremath{\psi_y}\xspace}
\def\mapr{\ensuremath{\psi_\rho}\xspace}
\def\mapt{\ensuremath{\psi_\theta}\xspace}
\def\maptt{\ensuremath{\psi_{\tilde{\theta}}}\xspace}
\def\mapp{\ensuremath{\psi_\phi}\xspace}
\def\mappa{\ensuremath{\psi_{\prma}}\xspace}
\def\mappb{\ensuremath{\psi_{\prmb}}\xspace}

\def\prma{\ensuremath{\phi\rho\tilde{\theta}}\xspace}
\def\prmb{\ensuremath{xy\theta}\xspace}

\newcommand{\ve}[1][x]{\ensuremath{\mathbf{#1}}\xspace}

\def\x{\ve[x]}

\def\spl{\hspace{-1pt}+\hspace{-1pt}}
\def\cart{C}
\def\pol{P}
\def\lw{W\textsubscript{S}\xspace}
\def\pcaw{W\xspace}
\def\pcawt{W\textsubscript{UA}\xspace}
\def\pcaws{W\textsubscript{US}\xspace}

\def\sssp{\hspace{2pt}}
\def\ssp{\hspace{3pt}}
\def\msp{\hspace{5pt}}

\renewcommand{\paragraph}[1]{\vspace{.0\baselineskip}\noindent{\bf #1}\xspace}

\def\etal{\emph{et al.}\xspace}
\def\ie{\emph{i.e.}\xspace}
\def\eg{\emph{e.g.}\xspace}
\def\etc{\emph{etc.}\xspace}
\def\wrt{\emph{w.r.t.}\xspace}

\def\xya{{\it cartes}\xspace}
\def\baseline{{\it polar}\xspace}

\def\half{\ensuremath{\nicefrac{1}{2}}\xspace}

\newcommand{\leg}[1]{\addlegendentry{#1}}

\renewcommand{\paragraph}[1]{{\medskip \noindent \bf #1}}

\def\sssp{\hspace{2pt}}
\def\ssp{\hspace{3pt}}
\def\msp{\hspace{5pt}}

%% file: intro.tex
\section{Introduction}
\label{sec:intro}
Representing and matching local features is an essential step of several
computer vision tasks. It has attracted a lot of attention in the last decades,
when local features still were a required step of most approaches.
Despite the large focus on Convolutional Neural Networks (CNN) to process whole
images, local features still remain important and necessary for tasks such as
Structure-from-Motion (SfM)~\cite{FGGJR10,HSD+15,SF16}, stereo matching~\cite{MMPL15}, or
retrieval under severe change in viewpoint or scale~\cite{SRCF15,ZZS+17}.

Classical approaches involve hand-crafted design of a local descriptor,
which has been the practice for more than a decade with some widely used
examples~\cite{L04,BETV09,MS05,TLF10,CLSF10,LCS11}. Such descriptors do
not require any training data or supervision. This kind of approach
allows to easily inject domain expertise, prior knowledge or even the result
of a thorough analysis~\cite{DS15}. Learning methods have been also employed
in order to learn parts of the hand-crafted design, \eg the pooling regions~\cite{WB07,SVZ13},
 from the training data.

Recently, the focus has shifted from hand-crafted descriptors to CNN-based
descriptors. Learning such descriptors relies on large training sets of
patches, that are commonly provided as a side-product of SfM~\cite{WB07}.
Integrating domain expertise has been mostly so far neglected in this kind of approaches.
Nevertheless, remarkable performance is achieved on a standard benchmark~\cite{BRPM16,TW17,MMR+17}.
On the other hand, recent work~\cite{BLVM17,SLJS17} shows that many CNN-based approaches do
not necessarily generalize equally well on different tasks or different datasets.
Hand-crafted descriptors still appear an attractive alternative.

In this work, we choose to work with a particular family of hand-crafted descriptors,
the so called \emph{kernel descriptors}~\cite{BS09,BRF11,TBFJ15}. They provide a
quite flexible framework for matching sets, patches in our case, by encoding different properties
of the set elements, pixels in our case.  In particular, we build upon the hand-crafted kernel descriptor
 proposed by Bursuc~\etal~\cite{BTJ15} that is shown to have good performance, even compared
to learned alternatives. Its few parameters are easily tuned on a validation
set, while it is shown to perform well on multiple tasks, as we confirm in our
experiments.

Further post-processing or descriptor normalization, such as Principal Component Analysis (PCA)
 and power-law normalization, is shown to be effective on different tasks~\cite{DGJP13,BTJ15,MM07,TTO16}.
We combine our descriptor with such post-processing that is learned from the data
in unsupervised or supervised ways. We show how to reduce the estimation error
and significantly improve results even without any supervision.

The hand-crafted nature and simplicity of our descriptor allows to
visualize and analyze its parametrization, and finally understand its advantages and disadvantages.
It leads us to propose a simple combination of parametrizations each offering robustness to
different types of patch miss-registrations. Interestingly, the same analysis
is possible even for the learned post-processing. We observe that its effect
on the patch similarity is semantically meaningful. The feasibility of such
analysis and visualization is an advantage or our approach, and hand-crafted
approaches in general, compared to CNN-based methods. 
Several insightful ablation and visualization studies~\cite{ZF14,YCN+15,MV16,BZK+17}
reveal what a CNN has learned. 
This typically provides only a partial view, \ie for a small number of neurons, on their behavior, 
while our approach enables visualization of the overall learned similarity in a general way that is not restricted to particular examples.

This work is an extension of our earlier conference publication~\cite{MTC17}.
In addition to the earlier version, we propose unsupervised whitening with shrinkage,
give extra insight about its effect on patch similarity, present extended comparisons
of different whitening variants and provide a proof justifying the absence of
regularized concatenation.

The manuscript is organized as follows. Related work is discussed
in Section~\ref{sec:related}, and background knowledge for kernel descriptors is
presented in Section~\ref{sec:prem}. Our descriptor, the different whitening
variants, and their interpretation are described in Section~\ref{sec:method}.
Finally, the experimental validation on two patch benchmarks is presented in Section~\ref{sec:exp}.

%% file: related.tex
\section{Related work}
\label{sec:related}

We review prior work on local descriptors, covering both hand-crafted and learned ones.

\subsection{Hand-crafted descriptors}
Hand-crafted descriptors have dominated the research landscape and a variety of approaches and methodologies exists. There are different variants on descriptors building features from filter-bank responses~\cite{BETV09,BSW05,KY08,OT01,SMo97}, pixel gradients~\cite{L04,MS05,TLF10,AY11}, pixel intensities~\cite{SI07,CLSF10,LCS11,RRKB11}, ordering or ranking of pixel intensities~\cite{OPM02,HPS09}, local edge shape~\cite{FL07}.
Some approaches focus on particular aspects of the local descriptors, such as a injecting invariance in the patch descriptor~\cite{OPM02,LSP05,AMHP09,TTO16}, computational efficiency~\cite{TLF10,AY11}, binary descriptors~\cite{CLSF10,LCS11,AOV12}.

A popular direction is that of gradient histogram-based descriptors,
where the most popular representative is SIFT~\cite{L04}. SIFT is a
long-standing top performer on multiple benchmarks and tasks across
the years. Multiple improvements for SIFT have been subsequently proposed:
PCA-SIFT~\cite{KeS04}, ASIFT~\cite{YM09}, OpponentSIFT~\cite{SGS10},
3D-SIFT~\cite{SAS07}, RootSIFT~\cite{AZ12}, DSP-SIFT~\cite{DS15}, \etc
A simple and effective improvement of SIFT is brought by the RootSIFT
descriptor~\cite{AZ12}, which uses Hellinger kernel as similarity measure.
DSP-SIFT~\cite{DS15} counters the aliasing effects caused by the binned
quantization in SIFT by pooling gradients over multiple scales instead of
only the scale selected by SIFT. Our \emph{kernelized descriptor} also
deals with quantization artifacts by embedding each pixel in a continuous
space and the aggregating pixels per patch by sum-pooling.

Kernel descriptors based on the idea of Efficient Match Kernels
(EMK)~\cite{BS09} encode entities inside a patch (such a gradient, color, \etc)
in a continuous domain, rather than as a histogram. The kernels and their few
parameters are often hand-picked and tuned on a validation set. Kernel
descriptors are commonly represented by a finite-dimensional explicit feature
maps~\cite{VZ12}. Quantized descriptors, such as SIFT, can be also interpreted as kernel
descriptor~\cite{BTJ15,BRF10}. Furthermore, the widely used RootSIFT descriptor~\cite{AZ12} can be also thought of as an explicit feature map from the original SIFT space to the RootSIFT space, such that the Hellinger kernel is \emph{linearised}, \ie the linear kernel (\ie dot product) in RootSIFT space is equivalent to the Hellinger kernel in the original SIFT space. In this case, the feature mapping is performed by \ensuremath{\ell_1}-normalization and square-rooting, without any expansion in dimensionality.

In this work we build upon EMK by integrating multiple pixel attributes in the patch descriptor. Unlike EMK which relies on features from random projections that require subsequent learning, we leverage instead explicit feature maps to approximate a kernel behavior directly. These representations can be further improved by minimal learning.

\subsection{Learned descriptors}
Learned descriptors commonly require annotation at patch level. Therefore,
research in this direction is facilitated by the release of datasets that
originate from an SfM system~\cite{WB07,PDHM+15}. Such training datasets allow
effective learning of local descriptors, and in particular, their pooling
regions~\cite{WB07,SVZ13}, filter banks~\cite{WB07}, transformations for
dimensionality reduction~\cite{SVZ13} or embeddings~\cite{PISZ10}.

Kernelized descriptors are formulated within a supervised framework by
Wang \etal~\cite{WWZX+13}, where image labels enable kernel learning and
dimensionality reduction. In this work, we rather focus on learning
discriminative projections with minimal or no supervision.
 This is several orders of magnitude faster to learn than other learning approaches.

Recently, local descriptor learning is dominated by deep learning. The proposed network
architectures mimic the ones for full-image processing. They have fewer parameters, however they still use a large amount of training patches.

Among representative examples is the work of Simo-Serra \etal~\cite{STFK+15} training
with hard positive and negative examples or the work of Zagoruyko~\cite{ZK15} where a central-surround
representation is found to be immensely beneficial. Such CNN-based approaches are seen as joint feature,
filter bank, and metric learning~\cite{HLJS+15} since both the convolutional filters, patch descriptor
and metrics are learned end-to-end. Going further towards an end-to-end pipeline for patch detection and
description, LIFT~\cite{YTL+16} advances a multi-step architecture with several spatial transformer
modules~\cite{JSZ15} that detects interest points and crops them, identifies their dominant orientation and
rotates them accordingly and finally extract a patch descriptor.
Paulin \etal~\cite{PMD+16} propose a deep patch descriptor from unsupervised learning. They consider a convolutional
kernel network~\cite{MKHS14} with feature maps compatible with the Gaussian kernel and which require layer-wise training.

Recent works in deep patch descriptors lean towards more compact architectures with more carefully designed training strategies and loss functions. Balntas \etal~\cite{BRPM16,BJTM16} advance shallower architectures with improved triplet ranking
loss~\cite{BRPM16,BJTM16}. In L2-Net~\cite{TW17} supervision is imposed on intermediate feature maps, the loss function integrates multiple attributes, while sampling of training data is done progressively to better balance positive and negative pairs at each step. In HardNet~\cite{MMR+17}, Mishchuk \etal extend L2-Net with a loss that mimics Lowe's matching criterion by maximizing the distance between the closest positive and closest negative example in the batch. HardNet is currently a top perfomer on most benchmarks.
Despite obtaining impressive results on standard benchmarks, the generalization of CNN-based local descriptors to other datasets is not always the case~\cite{SLJS17}.

\subsection{Post-processing}

A post-processing step is common to both hand-crafted and learned descriptors.
This post-processing ranges from simple \l2 normalization, PCA dimensionality
reduction, to transformations learned on annotated data~\cite{RTC16,BHW11,BLVM17,JC12}.

%% file: preliminaries.tex
\section{Preliminaries}
\label{sec:prem}
%
% \textbf{Kernelized descriptors.}
\subsection{Kernelized descriptors.}
In general lines we follow the formulation of Bursuc \etal~\cite{BTJ15}.
We represent a patch \P as a set of pixels $\p \in \P$ and compare two patches
\P and \Q via a match kernel
\begin{align}
\M(\P,\Q) = \sum_{\p\in \P} \sum_{\q\in \Q} \k(\p,\q),
\end{align}
where kernel $\k: \real^n \times \real^n \rightarrow \real$ is a similarity
function, typically non-linear, comparing two pixels or their corresponding feature vectors. The evaluation of this match kernel is costly as it computes exhaustively similarities between all pairs of pixels from the two sets.
Match kernel $\M(\P,\Q)$ can be approximated with
EMK~\cite{BS09}. It uses an explicit feature map $\map: \real^n \rightarrow \real^d$ to
approximate this result as
\begin{equation}
\begin{split}
\M(\P,\Q) & = \sum_{\p\in \P} \sum_{\q\in \Q} \k(\p,\q) \\
          & \approx \sum_{\p\in \P} \sum_{\q\in \Q} \map(\p)^\top \map(\q) \\
          & = \sum_{\p\in \P} \map(\p)^\top \sum_{\q\in \Q} \map(\q).
\end{split}
\end{equation}
Vector $\V(\P) = \sum_{\p\in \P} \map(\p)$ is a \emph{kernelized descriptor} (\des),
associated with patch \P, used to approximate $\M(\P,\Q)$, whose explicit
evaluation is costly. The approximation is given by a dot product
$\V(\P)^\top \V(\Q)$, where $\V(\P)\in \real^d$.
To ensure a unit self similarity, \l2-normalization by a factor \n is introduced.
The normalized \des is then given by $\bar{\V}(\P) = \gamma(\P) \V(\P)$, where
$\gamma(\P) = (\V(\P)^\top \V(\P))^{-\half}$.

Kernel \k comprises products of kernels, each kernel acting on a different scalar pixel attribute
\begin{equation}
\k(\p,\q) = \k_1(\p_1,\q_1) \k_2(\p_2,\q_2) \ldots \k_n(\p_n,\q_n),
\end{equation}
where kernel $\k_n$ is pairwise similarity function for scalars and $\p_n$ are
pixel attributes such as position and gradient orientation. Feature map $\map_n$
corresponds to kernel $\k_n$ and feature map $\map$ is constructed via Kronecker
product of individual feature maps
$\map(\p) = \map_1(\p_1) \otimes \map_2(\p_2) \otimes \ldots \otimes \map_n(\p_n)$.
It is straightforward to show that
\begin{equation}
\map(\p)^\top \map(\q) \approx \k_1(\p_1,\q_1) \k_2(\p_2,\q_2) \ldots \k_n(\p_n,\q_n).
\end{equation}

\subsection{Feature maps.}
As non-linear kernel for scalars we use the normalized Von Mises probability
density function\footnote{Also known as the periodic normal distribution},
which is used for image~\cite{TBFJ15} and patch~\cite{BTJ15} representations.
It is parametrized by $\kappa$ controlling the shape of the kernel, where lower
$\kappa$ corresponds to wider kernel, \ie less selective kernel. We use a stationary (shift invariant) kernel that, by
definition, depends only on the difference
$\Delta_{n} = \p_n-\q_n$, \ie $\k_\VM(\p_n, \q_n) := \k_\VM(\Delta_n)$.
We approximate this probability density function with Fourier series with $N$ frequencies that produces a
feature map $\map_\VM: \real \rightarrow \real^{2N+1}$. It has the property that
\begin{equation}
\k_\VM(\p_n, \q_n) \approx \map_\VM(\p_n)^\top \map_\VM(\q_n).
\end{equation}

In particular we approximate the Fourier series by the sum of the first $N$ terms as
\begin{equation}
\k_\VM(\Delta_n) \approx \sum_{i=0}^{N}\gamma_{i} \cos(i\Delta_n).
\end{equation}
The feature map $\map_\VM(\p_n)$ is designed as follows:
\begin{multline}
\small
\label{eq:FeatureMap}
\map_\VM(\p_n) = (\sqrt{\gamma_{0}}, \sqrt{\gamma_{1}}\cos(\p_n),\ldots,\sqrt{\gamma_{N}}\cos(N\p_n), \\
\sqrt{\gamma_{1}}\sin(\p_n),\ldots,\sqrt{\gamma_{N}}\sin(N\p_n))^{\top}.
\end{multline}
This vector has $2N+1$ components. It is now easy to show that the inner product of two feature maps
is approximating the kernel . That is,
\begin{align}
\map_\VM(\p_n)^{\top}\map_\VM(\q_n) &=\gamma_{0} + \sum_{i=1}^{N}\gamma_{i}(\cos(i\p_n)\cos(i\q_n) \\ \nonumber
                                                                        &~~~~~~~~~~~~~~~~ + \sin(i\p_n)\sin(i\q_n))\\ \nonumber
&=\sum_{i=0}^{N} \gamma_{i}\cos(i(\p_n-\q_n))\\ \nonumber
& \approx \k_\VM(\Delta_n).
\end{align}
The reader is encouraged to read prior work for details on these feature maps~\cite{VZ10,Chum15}, which are previously used in various contexts~\cite{TBFJ15,BTJ15}.

\subsection{Descriptor post-processing.}
It is known that further descriptor post-processing~\cite{RTC16,BL15,BTJ15} is
beneficial. In particular, \des is further centered and projected as
\begin{equation}
\hat{\V}(\P) = A^\top(\bar{\V}(\P) -\mu),
\label{equ:norm}
\end{equation}
where $\mu \in \real^d$ and $A\in \real^{d\times d}$ are the mean vector and
the projection matrix. These are commonly learned by PCA~\cite{JC12} or with
supervision~\cite{RTC16}. The final descriptor is always \l2-normalized in the end.

%% file: method.tex
\section{Method}
\label{sec:method}
In this section we consider different patch parametrizations and kernels that
result in different patch similarity. We discuss the benefits of each and
propose how to combine them. We further learn descriptor transformation with
or without supervision and provide useful insight on how patch similarity is affected.

\subsection{Patch attributes.}
We consider a pixel \p to be associated with coordinates \px, \py in Cartesian
coordinate system, coordinates \pr, \pp in polar coordinate system, pixel
gradient magnitude \pm, and pixel gradient angle \pt.
Angles \pt, \pp $\in [0, 2\pi]$, distance from the center \pr is normalized to
$[0, 1]$, while coordinates \px, \py $\in \{1, 2, \ldots, W\}$ for $W \times W$
patches. In order to use feature map $\map_\VM$, attributes \pr, \px, and \py
are linearly mapped to $[0, \pi]$. The gradient angle is expressed \wrt the
patch orientation, \ie \pt directly, or \wrt to the position of the pixel.
The latter is given as $\ptt = \pt - \pp$.

\input{figs/tex/figkernels}
\subsection{Patch parametrizations.}
Composing patch kernel \k as a product of kernels over different attributes
enables easy design of various patch similarities. Correspondingly, this
defines different \des.
All attributes \px, \py, \pr, \pt, \pp, and \ptt are matched by the Von Mises
kernel, namely, \kx, \ky, \kr, \kt, \kp, and \ktt parameterized by
$\kappa_x$, $\kappa_y$, $\kappa_\rho$, $\kappa_\theta$, $\kappa_\phi$,
and $\kappa_{\tilde{\theta}}$, respectively.
In a similar manner to SIFT, we apply a Gaussian mask by $\pg = \exp(-\pr^2)$ which gives more importance to
central pixels.

In this work we focus on the two following match kernels over patches.
One in \emph{polar} coordinates
\begin{multline}
\M_{\prma}(\P,\Q) = \sum_{\p\in \P} \sum_{\q\in \Q} \pg\qg \sqrt{\pm}\sqrt{\qm}\kp(\pp,\qp) \\
\kr(\pr,\qr)\ktt(\ptt,\qtt),
\end{multline}
and one in \emph{Cartesian} coordinates
\begin{multline}
\M_{\prmb}(\P,\Q) = \sum_{\p\in \P} \sum_{\q\in \Q} \pg\qg \sqrt{\pm}\sqrt{\qm}\kx(\px,\qx) \\
\ky(\py,\qy)\kt(\pt,\qt).
\end{multline}

The \des for the two cases are given by
\begin{equation}
\begin{aligned}
\V_{\prma}(\P) &= \sum_{\p\in \P} \pg\sqrt{\pm} \mapp(\pp)\otimes \mapr(\pr) \otimes \maptt(\ptt) \\
               &= \sum_{\p\in \P} \pg\sqrt{\pm} \mappa(\p)
\end{aligned}
\end{equation}

\begin{equation}
\begin{aligned}
\V_{\prmb}(\P) &= \sum_{\p\in \P} \pg \sqrt{\pm} \mapx(\px)\otimes \mapy(\py) \otimes \mapt(\pt) \\
               &= \sum_{\p\in \P} \pg\sqrt{\pm}\mappb(\p).
\end{aligned}
\end{equation}
The $\V_{\prma}$ variant is exactly the one proposed by Bursuc \etal~\cite{BTJ15},
considered as a baseline in this work. Different parametrizations result in
different patch similarity, which is analyzed in the following. In
Figure~\ref{fig:kernels} we present the approximation of kernels used per attribute.

\input{figs/tex/parametrizations.tex}
\input{figs/tex/rotins.tex}

\subsection{Post-processing learned with or w/o supervision.}
\label{sec:whitening}
We detail different ways to learn the projection matrix $A$ of (\ref{equ:norm}) to perform the descriptor post-processing. Let us consider a learning set of patches \Pb and the corresponding set of descriptors $V_\Pb = \{V(\P),~\P \in \Pb\}$.
Let $C$ be the covariance matrix of $V_\Pb$. Vector $\mu$ is the mean descriptor vector, and different ways to compute $A$ are as follows.

\textbf{Supervised whitening.}
We further assume that supervision is available in the form of pairs of matching patches.
This is given by set $\Mb = \{(\P,\Q) \in \Pb \times \Pb,~\P\sim \Q\}$, where $\sim$ denotes matching patches.
We follow the work of Mikolajczyk and Matas~\cite{MM07} to learn discriminative projections using the available supervision.
The discriminative projection is composed of two parts, a whitening part and a rotation part. The whitening part is obtained from the intraclass (matching pairs) covariance matrix $C_\Mb$, while the rotation part is the PCA of the interclass (non-matching pairs) covariance matrix in the whitened space. We set the interclass one to be equal to $C$ as this is dominated by non-matching pairs, while
the intraclass one is given by
\begin{equation}
C_\Mb = \sum_{(\P,\Q) \in \Mb} \left(V(\P) - V(\Q)\right)\left(V(\P) - V(\Q)\right)^\top.
\end{equation}

The projection matrix is now given by
\begin{equation} \label{eq:lw}
A = C_\Mb^{-\nicefrac{1}{2}}\mbox{eig}( C_\Mb^{-\nicefrac{1}{2}} C C_\Mb^{-\nicefrac{1}{2}} ),
\end{equation}
where $\mbox{eig}$ denotes the eigenvectors of a matrix into columns. To reduce the descriptor dimensionality, only eigenvectors corresponding to the largest eigenvalues are used. The same holds for all cases that we perform PCA in the rest of the paper. We refer to this transformation as supervised whitening (\lw).

\textbf{Unsupervised whitening.} There is no supervision in this case and the projection is learned via PCA on set $V_\Pb$. In particular, projection matrix is given by
\begin{equation} \label{eq:pcaw}
A = \mbox{eig}(C) \mbox{diag}(\lambda_1^{-\nicefrac{1}{2}}, \ldots, \lambda_d^{-\nicefrac{1}{2}})^\top,
\end{equation}
where $\mbox{diag}$ denotes a diagonal matrix with the given elements on its diagonal, and $\lambda_i$ is the $i$-th eigenvalue of matrix $C$. This method is called PCA whitening and we denote simply by \pcaw~\cite{JC12}.

\textbf{Unsupervised whitening with shrinkage.}
We extend the PCA whitening scheme by introducing parameter $t$ controlling the extent of whitening and the projection matrix becomes
\begin{equation} \label{eq:pcawp}
A = \mbox{eig}(C) \mbox{diag}(\lambda_1^{-\nicefrac{t}{2}}, \ldots, \lambda_d^{-\nicefrac{t}{2}})^\top,
\end{equation}
where $t \in [0, 1]$, with $t=1$ corresponding to the standard PCA whitening and $t=0$ to simple rotation without whitening.

Equivalently, $t=0$ imposes the covariance matrix to be identity.
We call this method attenuated PCA whitening and denote it by \pcawt.

The aforementioned process resembles covariance estimation with shrinkage~\cite{LW04,LW04b}.
The sample covariance matrix is known to be a noise estimator, especially when
the available samples are not sufficient relatively to the number of dimensions~\cite{LW04b}.
 Ledoit and Wolf~\cite{LW04b} propose to replace this by a linear combination of the sample covariance
matrix and a structured estimator. Their solution is well conditioned and is
shown to reduce the effect of noisy estimation in eigen decomposition.
The imposed condition is simply that all variances are the same and all covariances are zero.
The shrunk covariance is
\begin{equation}
\tilde{C} = (1-\beta)C + \beta \I_d,
\label{eq:cshrink}
\end{equation}
where $\I_d$ is the identity matrix and $\beta$ the shrinking parameter.
This process ``shrinks'' extreme (too large or too small) eigenvalues to intermediate ones.
In our experiments we show that a simple tuning of parameter $\beta$ performs well across different context and datasets.
The projection matrix is now
\begin{equation} \label{eq:pcaws}
\small
A = \mbox{eig}(C) \mbox{diag}((\alpha\lambda_1+\beta)^{-\nicefrac{1}{2}}, \ldots, (\alpha\lambda_d+\beta)^{-\nicefrac{1}{2}})^\top,
\end{equation}
where $\alpha = 1- \beta$. We call this method PCA whitening with shrinkage and denote it by \pcaws.
We set parameter $\beta$ equal to the $i$-th eigenvalue.
A method similar to ours is used in the work of Brown \etal~\cite{BHW11}, but does not allow dimensionality reduction since descriptors are projected back to the original space after the eigenvalue clipping.

\subsection{Visualizing and understanding patch similarity.}
We define pixel similarity $\M(\p,\q)$ as kernel response between pixels
\p and \q, approximated as $\M(\p,\q) \approx \map(\p)^\top \map(\q)$. To show
a spatial distribution of the influence of pixel \p, we define a \emph{patch map}
of pixel \p (fixed \px, \py, and \pt). The patch map has the same size as the image patches;
for each pixel \q of the patch, map $\M(\p,\q)$ is evaluated for some constant value of
$\q_\theta$.

For example, in Figure~\ref{fig:parametrizations} patch maps for different
kernels are shown. The position of \p is denoted by $\times$ symbol. Then, $\p_\theta = 0$,
while $\q_\theta = 0$ for all spatial locations of \q in the
top row and $\q_\theta = -\pi/8$ in the bottom row. This example shows the
toy patches and their gradient angles in arrows to be more explanatory. The toy
patches are directly defined by $\p_{\theta}$, and $\q_{\theta}$.
Only $\p_{\theta}$ and $\q_{\theta}$ are used in later examples, while the toy patches are
 skipped from the figures.

The example in Figure~\ref{fig:parametrizations} reveals a discontinuity
near the center of the patch when pixel similarity is given by $\V_{\prma}$ descriptor.
It  is caused by the polar coordinate system where a small difference in the position near the origin
causes large difference in $\phi$ and $\tilde{\theta}$.
The patch maps reveal weaknesses of kernel descriptors, such the aforementioned
discontinuity, but also advantages of each parametrization.
It is easy to observe that the kernel parametrized by Cartesian coordinates and absolute angle of the
gradient ($\V_{\prmb}$, third column) is insensitive to small translations, \ie feature point displacement.
Moreover, in the bottom row we see that using the relative gradient direction $\tilde{\theta}$ allows to
compensate for imprecision caused by small patch rotation, \ie the most similar
pixel is not the one at the location of \p with different $\tilde{\theta}$, but
a rotated pixel with more similar value of  $\tilde{\theta}$.
This effect is further analyzed in Figure~\ref{fig:rotins}.
The final similarity involves the product of two kernels that both depend on
angle $\phi$. They are both maximized at the same point if $\Delta\theta=0$, otherwise
not. The larger $\Delta\theta$ is, the maximum value moves further (in the patch) from $\p$.

We additionally construct patch maps in the case of descriptor post-processing
by a linear transformation, \eg descriptor whitening.
Now the contribution of a pixel pair is given by
\begin{equation}
\begin{aligned}
\hat{\M}(\p,\q) &= (A^\top(\map(\p)-\mu))^\top (A^\top(\map(\q)-\mu))\\
                &= (\map(\p)-\mu)^\top AA^\top(\map(\q)-\mu)\\
                &= \map(\p)^\top AA^\top \map(\q) - \map(\p)^\top AA^\top\mu \\
                &\quad - \map(\q)^\top AA^\top\mu + \mu^\top AA^\top \mu.
\end{aligned}
\end{equation}
The last term is constant and can be ignored. If $A$ is a rotation matrix
then the similiarity is affected just by shifting by $\mu$. After the transformation,
the similarity is no longer shift-invariant. Non-linear post-processing,
such as power-law normalization or simple \l2 normalization cannot be visualized,
as it acts after the pixel aggregation.

\input{figs/tex/joint}

\input{figs/tex/learning}
\input{figs/tex/sim1D}
\subsection{Combining kernel descriptors.}
We propose to take advantage of both parametrizations $\V_{\prma}$ and
$\V_{\prmb}$, by summing their contribution. This is performed by simple
concatenation of the two descriptors. Finally, whitening is jointly learned
and dimensionality reduction is performed.

In Figure~\ref{fig:joint} we show patch maps for the individual and combined
representation, for different pixels $\p$. Observe how the
combined one better behaves around the center. The combined descriptor inherits
reasonable behavior around the patch center and insensitivity to position misalignment
from the Cartesian parametrization, while insensitivity to dominant orientation
misalignment from the polar parametrization, as shown earlier.

\input{figs/response/sample_resp}

\subsection{Understanding the whitened patch similarity.}
\label{sec:under}
We learn the different whitening variants of Section~\ref{sec:whitening} and
visualize their patch maps in Figure~\ref{fig:learning}. All examples shown are for
$\Delta\theta=0$ but gradient angles \pt and \qt jointly vary.
We initially observe that the similarity is shift invariant only in the fist column
of patch maps where no whitening is applied. This is expected by definition.
Projecting by matrix $A$ does not allow to reconstruct the shift invariant kernels anymore;
the similarity does not only depend on $\Delta\theta$, which is 0, but also on \pt and \qt.

The patch similarity learned by whitening exhibits an interesting property.
The shape of the 2D similarity becomes anisotropic and gets aligned
with the orientation of the gradient.
Equivalently, it becomes perpendicular to the edge on which the pixel lies.
This is a semantically meaningful effect.
It prevents over-counting of pixel matching along aligned edges of the two patches.
In the case of a blob detector this can provide tolerance to errors in the scale
 estimation, \ie the similarity remains large towards the direction
 that the blob edges shift in case of scale estimation error.

We presume that this is learned by pixels with similar gradient angle that co-occur frequently.
A similar effect is captured by both the supervised and the unsupervised whitening
with covariance shrinkage, it is, though, less evident in the case of \pcaws.
Moreover, we see that it is mostly the Cartesian parametrization that allows this kind of deformation.

According to our interpretation, supervised whitening~\cite{MM07,RTC16} owes its success
to covariance estimation that is more noise free. The noise removal comes from supervision,
but we show that standard approaches for well-conditioned and accurate covariance estimation
have similar effect on the patch similarity even without supervision.
The observation that different parametrizations allow for different types of co-occurrences
to be captured is related to other domains too.
For instance. CNN-based image retrieval exhibits improvements after whitening~\cite{BL15},
but this is very unequal between average and max pooling.
However, observing the differences is not as easy as in our case with the visualized patch similarity.

Finally, we obtain slices from the 2D patch maps and present the 1D similarity kernels in Figure~\ref{fig:sim1D}.
It is the similarity between pixel \p and all pixels $\q \in \Q$ that lie on the vertical and horizontal lines
drawn on the patch map at the left of Figure~\ref{fig:sim1D}.
We present the case for which $\pt = 0$ and $\qt = 0, \forall \q \in \Q$ (top row in Figure~\ref{fig:learning}).
The gradient angle is fully horizontal in this case and the 2D similarity kernel tends to get aligned with that,
while the chosen slices are aligned in this fashion too.
In our experiments we show that all \lw, \pcawt, and \pcaws \ provide significant performance improvements.
However, herein, we observe that the underlined patch similarity demonstrates some differences.
Supervised whitening \lw \ is not a decreasing function, which might be an outcome of over-fitting to the training data.
This is further validated in our experiments. Finally, \pcawt \ and \lw \ are not maximized on point \p, which does not seem a desired
property. \pcaws \ is maximized similarly to the raw descriptor without any post-processing.

Patch maps are a way to visualize and study the general shape of the underlined similarity function.
In a similar manner, we visualize the kernel responses for a particular pair of patches to reflect which are the pixels contributing the most to the patch similarity.
This is achieved by assigning strength $\sum_{\q \in \Q}\hat{\M}(\p,\q)$ to pixel \p. In cases without whitening, $\M(\p,\q)$ is used.
We present such heat maps in Figure~\ref{fig:heatmaps}. 
Whitening significantly affects the contribution of most pixels. The over-counting phenomenon described in Section~\ref{sec:under} is also visible; some of the long edges are suppressed.

%% file: figs/tex/figkernels.tex
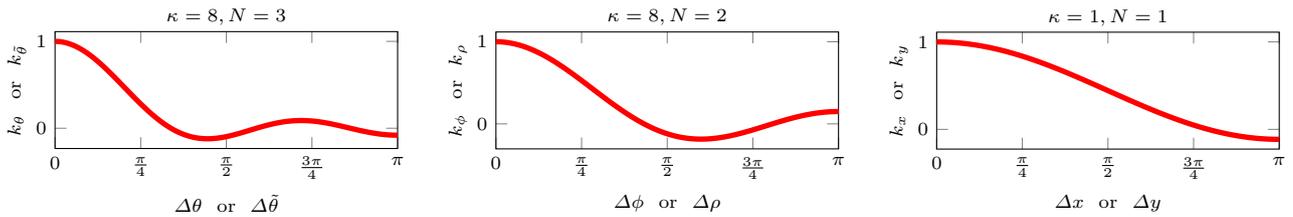
\begin{figure*}

\begin{tabular}{c}

\begin{tikzpicture}
 \tikzstyle{every node}=[font=\normalsize]
   \begin{axis}[%
        title={$\kappa = 8, N = 3$},
      height=0.18\textwidth,
      width=0.35\textwidth,
      xlabel={$\Delta\theta$~~or~~$\Delta\tilde{\theta}$},
      ylabel={$\kt$~~or~~$\ktt$},
      legend pos=north east,
      legend cell align=left,
      legend style={font=\tiny},
      ytick = {0, 1},
      xtick={0, 0.7854, 1.5708, 2.3562, 3.14159},
      xmin = 0, xmax = 3.14159,
      xticklabels={$0$,$\frac{\pi}{4}$,$\frac{\pi}{2}$,$\frac{3\pi}{4}$, $\pi$},
      y label style={at={(axis description cs:-0.06,0.5)}},
      title style={at={(axis description cs:0.5,0.85)}},
   ]
   \addplot[color=red, style=solid, line width=2pt] table[x index=0,y index=1]{figs/method/n3kappa8.dat};
   \end{axis}
\end{tikzpicture}
\hspace{5pt}
% \\
\begin{tikzpicture}
 \tikzstyle{every node}=[font=\normalsize]
   \begin{axis}[%
        title={$\kappa = 8, N = 2$},
      height=0.18\textwidth,
      width=0.35\textwidth,
      xlabel={$\Delta\phi$~~or~~$\Delta\rho$},
      ylabel={$\kp$~~or~~$\kr$},
      legend pos=north east,
      legend cell align=left,
      legend style={font=\tiny},
      xtick={0, 0.7854, 1.5708, 2.3562, 3.14159},
      ytick = {0, 1},
      xmin = 0, xmax = 3.14159,
      xticklabels={$0$,$\frac{\pi}{4}$,$\frac{\pi}{2}$,$\frac{3\pi}{4}$, $\pi$},
      y label style={at={(axis description cs:-0.06,0.5)}},
      title style={at={(axis description cs:0.5,0.85)}},
   ]
   \addplot[color=red, style=solid, line width=2pt] table[x index=0,y index=1]{figs/method/n2kappa8.dat};
   \end{axis}
\end{tikzpicture}
\hspace{5pt}
% \\
\begin{tikzpicture}
 \tikzstyle{every node}=[font=\normalsize]
   \begin{axis}[%
      title={$\kappa = 1, N = 1$},
      height=0.18\textwidth,
      width=0.35\textwidth,
      xlabel={$\Delta x$~~or~~$\Delta y$},
      ylabel={$\kx$~~or~~$\ky$},
      legend pos=north east,
      legend cell align=left,
      legend style={font=\tiny},
      xtick={0, 0.7854, 1.5708, 2.3562, 3.14159},
      ytick = {0, 1},
      xmin = 0, xmax = 3.14159,
      xticklabels={$0$,$\frac{\pi}{4}$,$\frac{\pi}{2}$,$\frac{3\pi}{4}$, $\pi$},
      y label style={at={(axis description cs:-0.06,0.5)}},
      title style={at={(axis description cs:0.5,0.85)}},
   ]
   \addplot[color=red, style=solid, line width=2pt] table[x index=0,y index=1]{figs/method/n1kappa1.dat};
   \end{axis}
\end{tikzpicture}

\end{tabular}
% \vspace{5pt}
\caption{Kernel approximations that we use for pixel attributes.
Parameter $\kappa$ and the number of frequencies $N$ define the final shape.
The choice of kernel parameters is guided by~\cite{BTJ15}.
\label{fig:kernels}}
\end{figure*}

%% file: figs/tex/parametrizations.tex
\begin{figure*}
\centering
\begin{tabular}{cccccc}
  Pixel \p & Patch \Q &
  \multirow{2}{*}{\kp\hspace{-2pt}\kr\hspace{-2pt}\kt} &
  \multirow{2}{*}{\kp\hspace{-2pt}\kr\hspace{-2pt}\ktt} &
  \multirow{2}{*}{\kx\hspace{-2pt}\ky\hspace{-2pt}\kt} &
  \multirow{2}{*}{\kx\hspace{-2pt}\ky\hspace{-2pt}\ktt} \\[2pt]
$\p_{\theta} = 0$ &  $\q_{\theta} = 0, \forall \q \in \Q$ & & & & \\
  \input{figs/tex/pixel_arrow_1}
    & \input{figs/tex/patch_arrow_1}
       & \includegraphics[height=60pt]{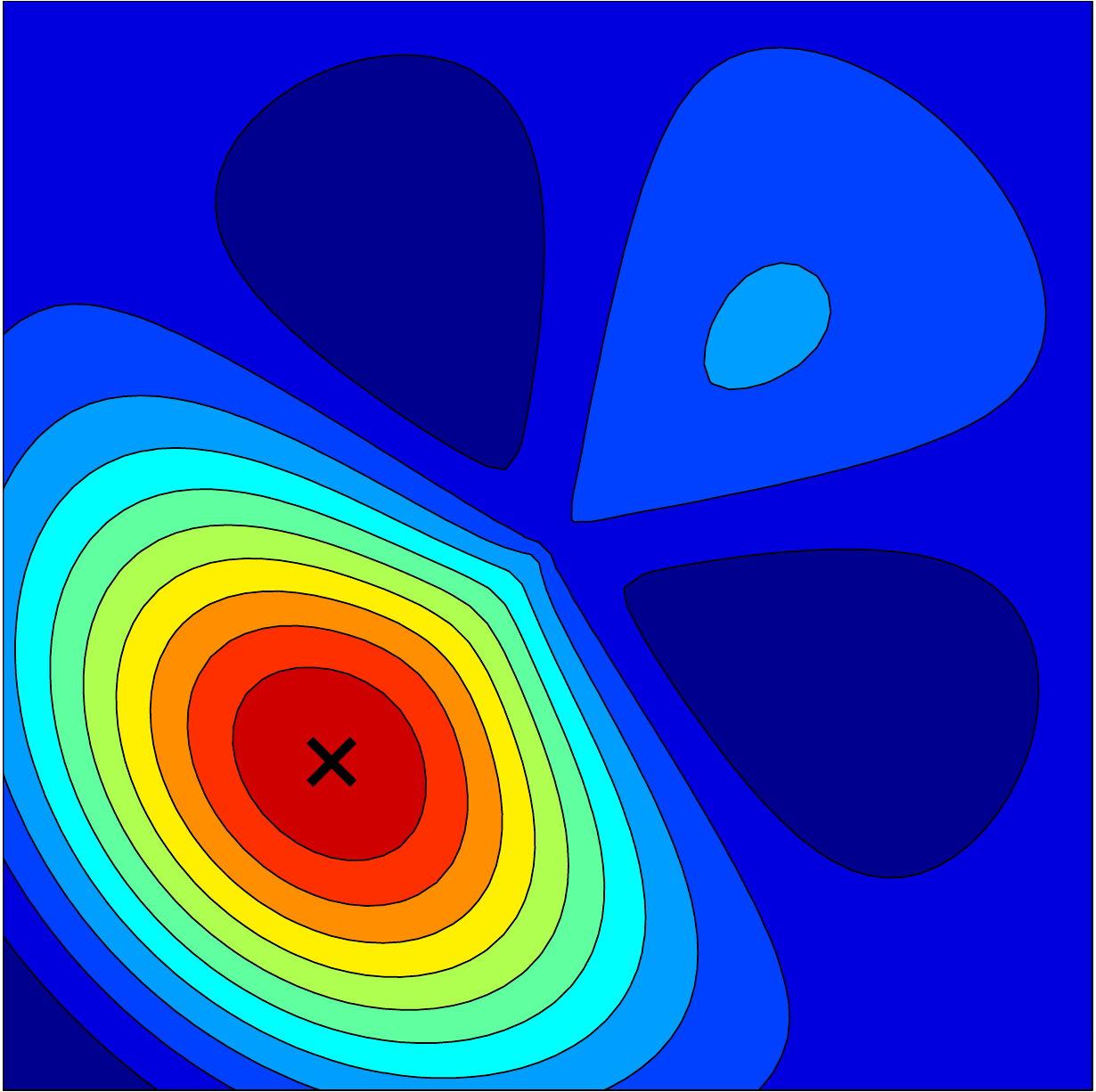}
         & \includegraphics[height=60pt]{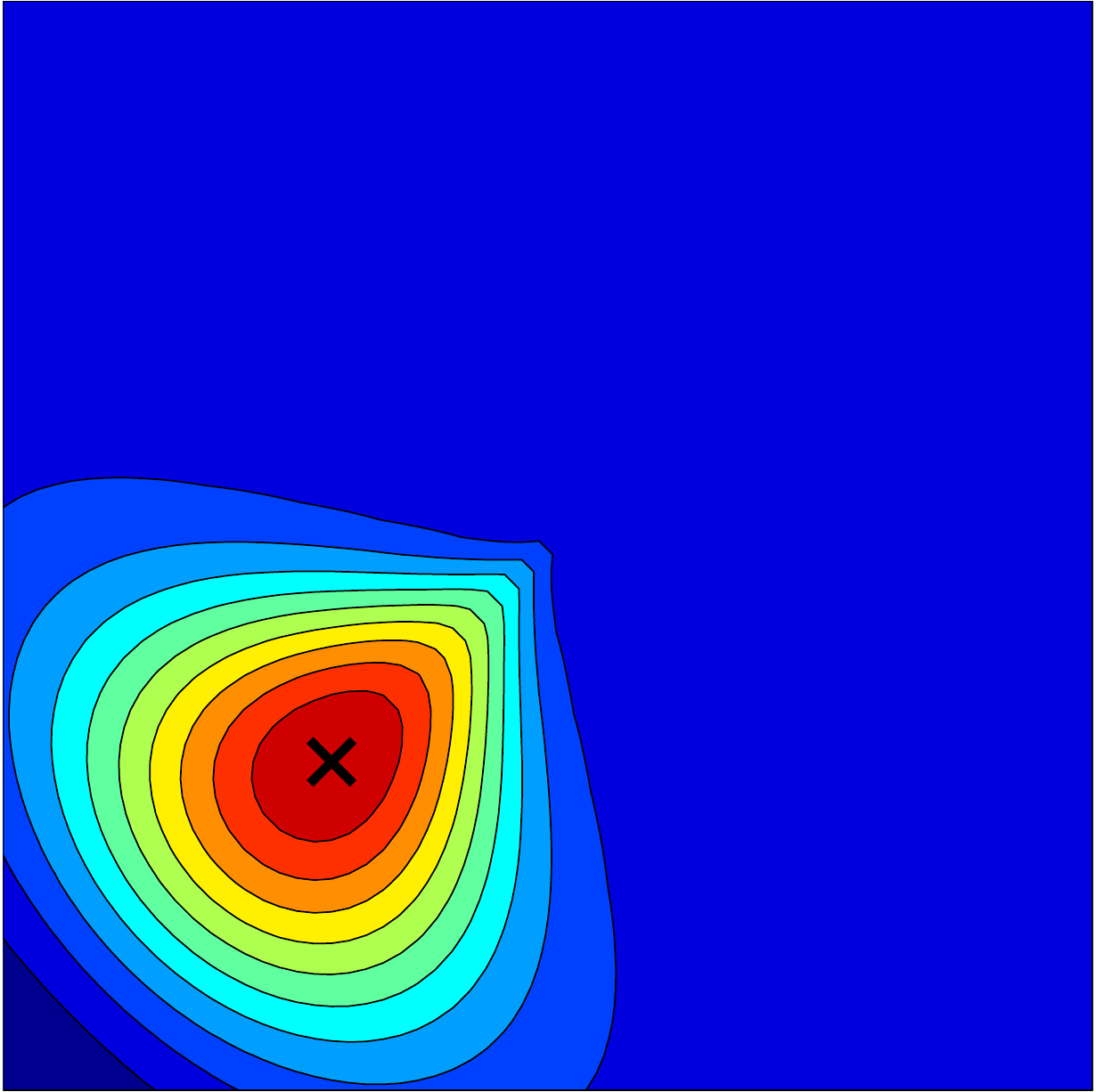}
           & \includegraphics[height=60pt]{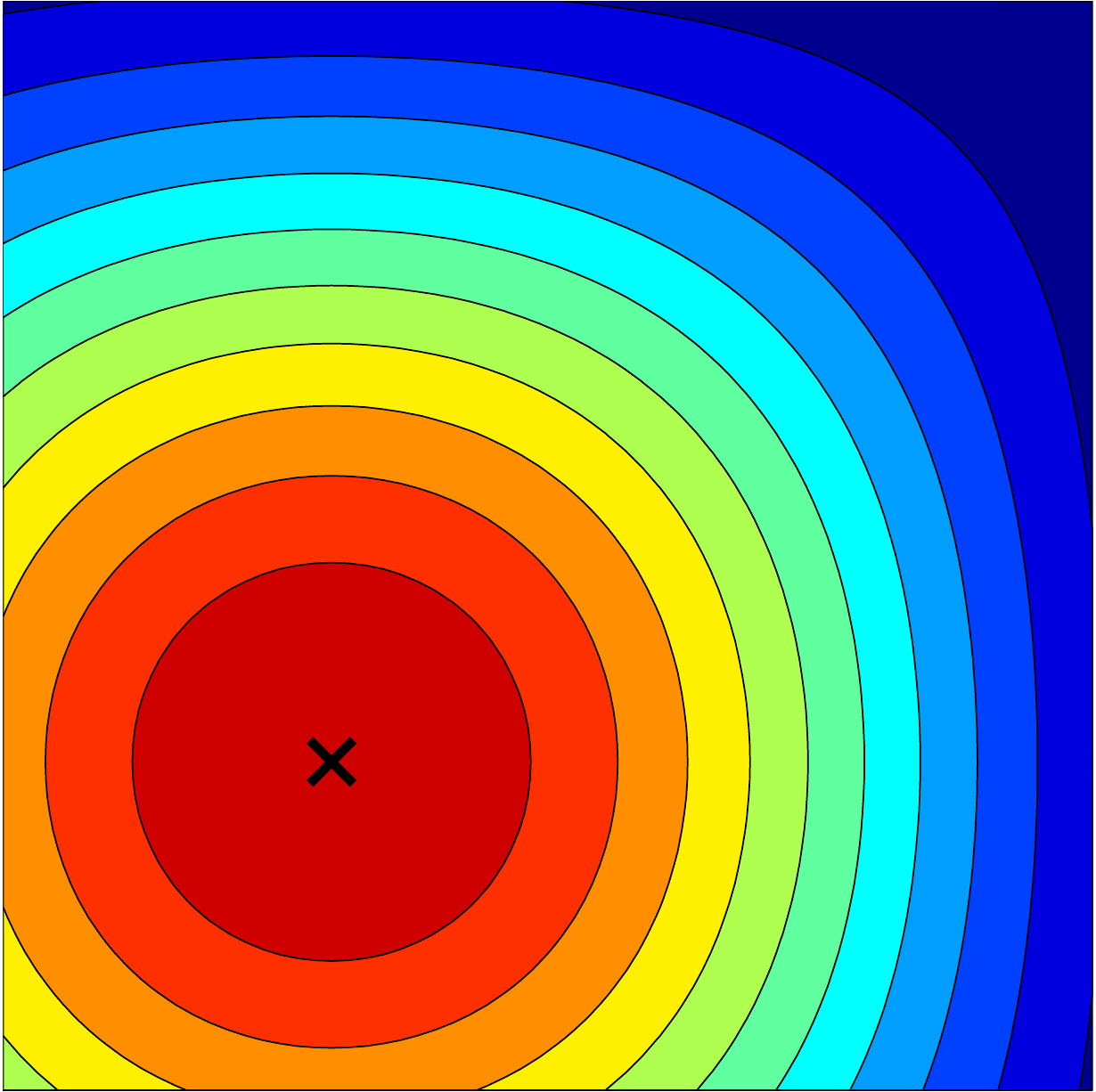}
             & \includegraphics[height=60pt]{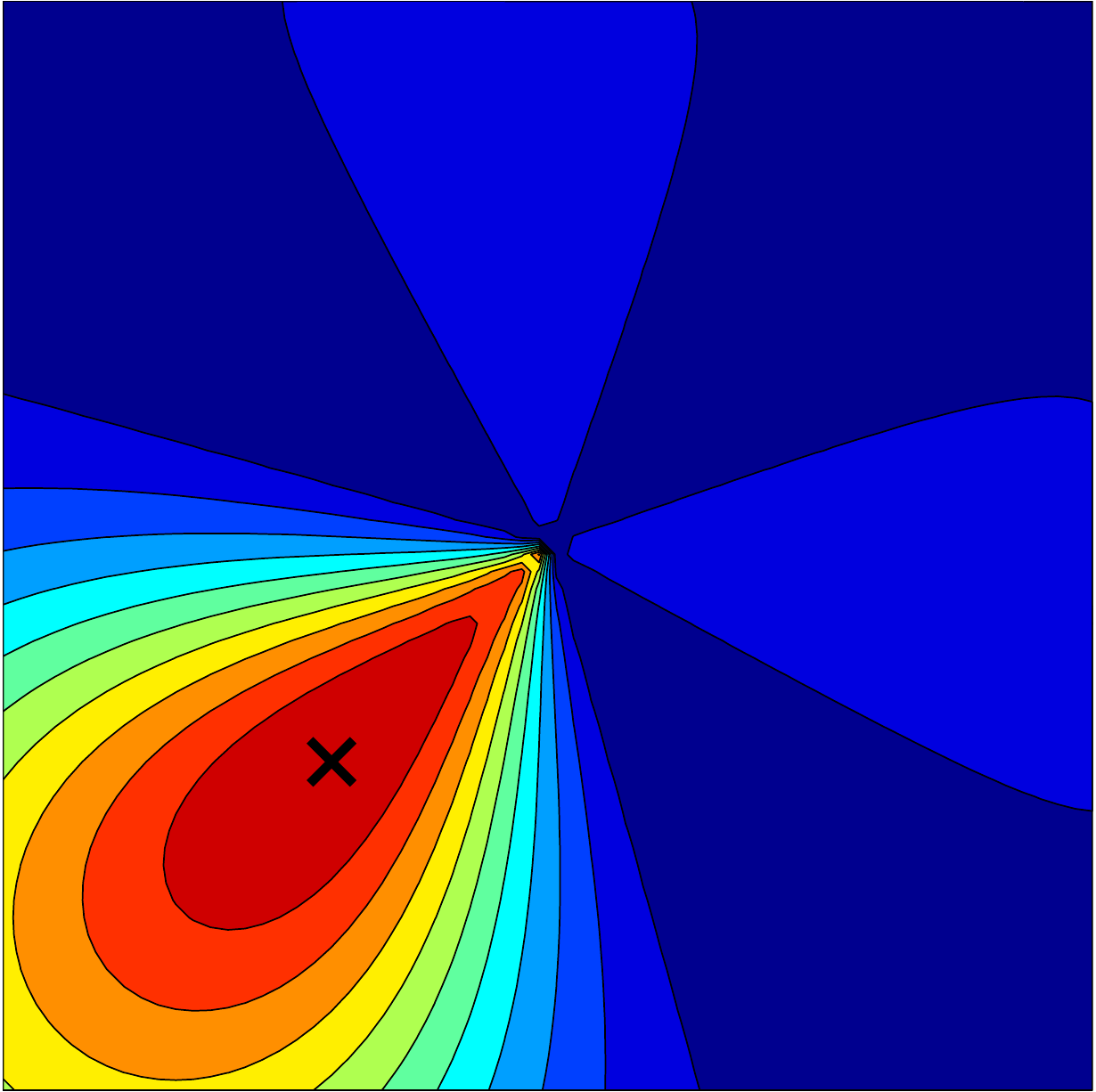}
\\
$\p_{\theta} = 0$ & $\q_{\theta} = -\pi/8, \forall \q \in \Q$ & & & & \\
  \input{figs/tex/pixel_arrow_1}
    & \input{figs/tex/patch_arrow_2}
        & \includegraphics[height=60pt]{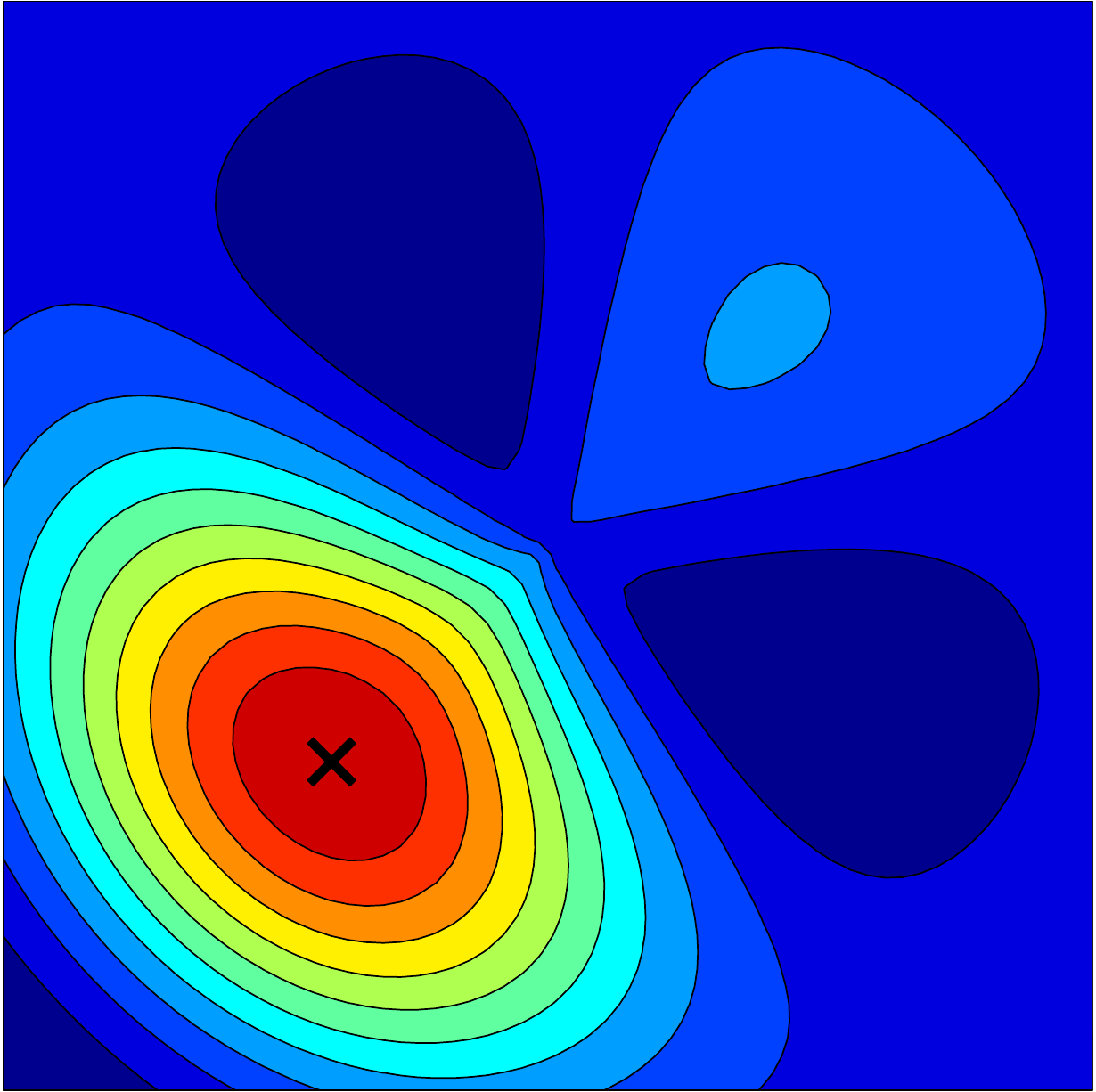}
          & \includegraphics[height=60pt]{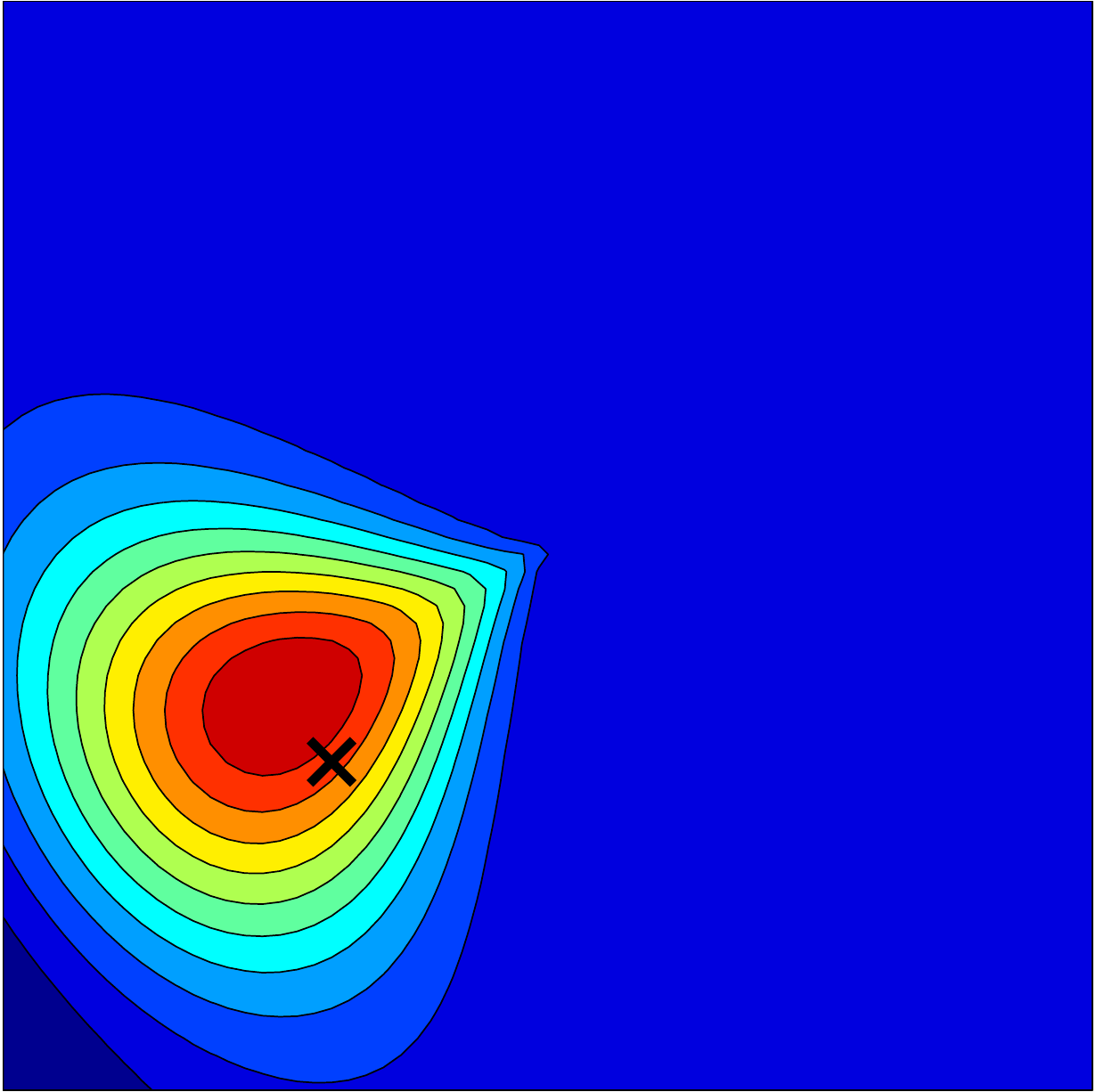}
            & \includegraphics[height=60pt]{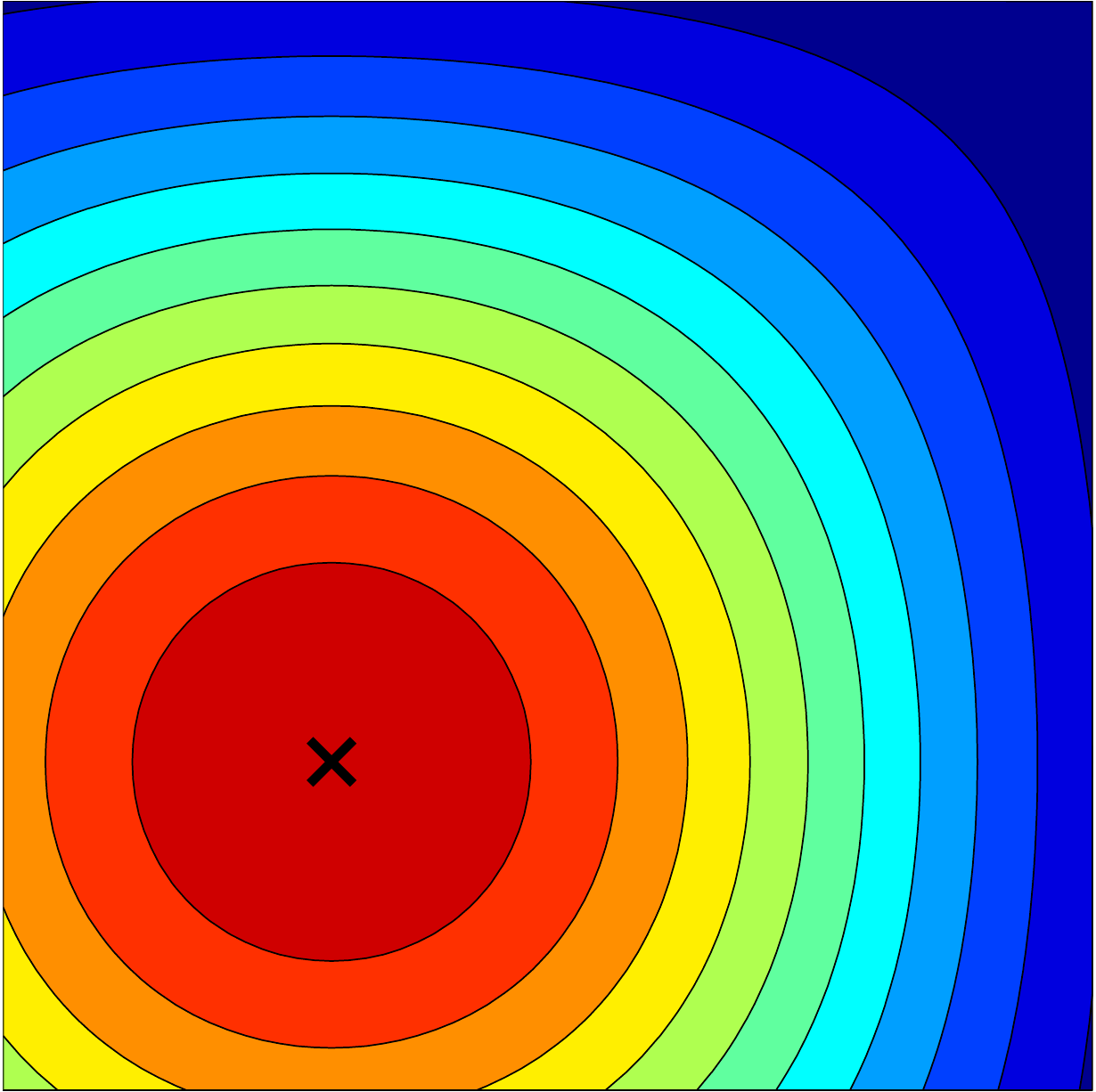}
              & \includegraphics[height=60pt]{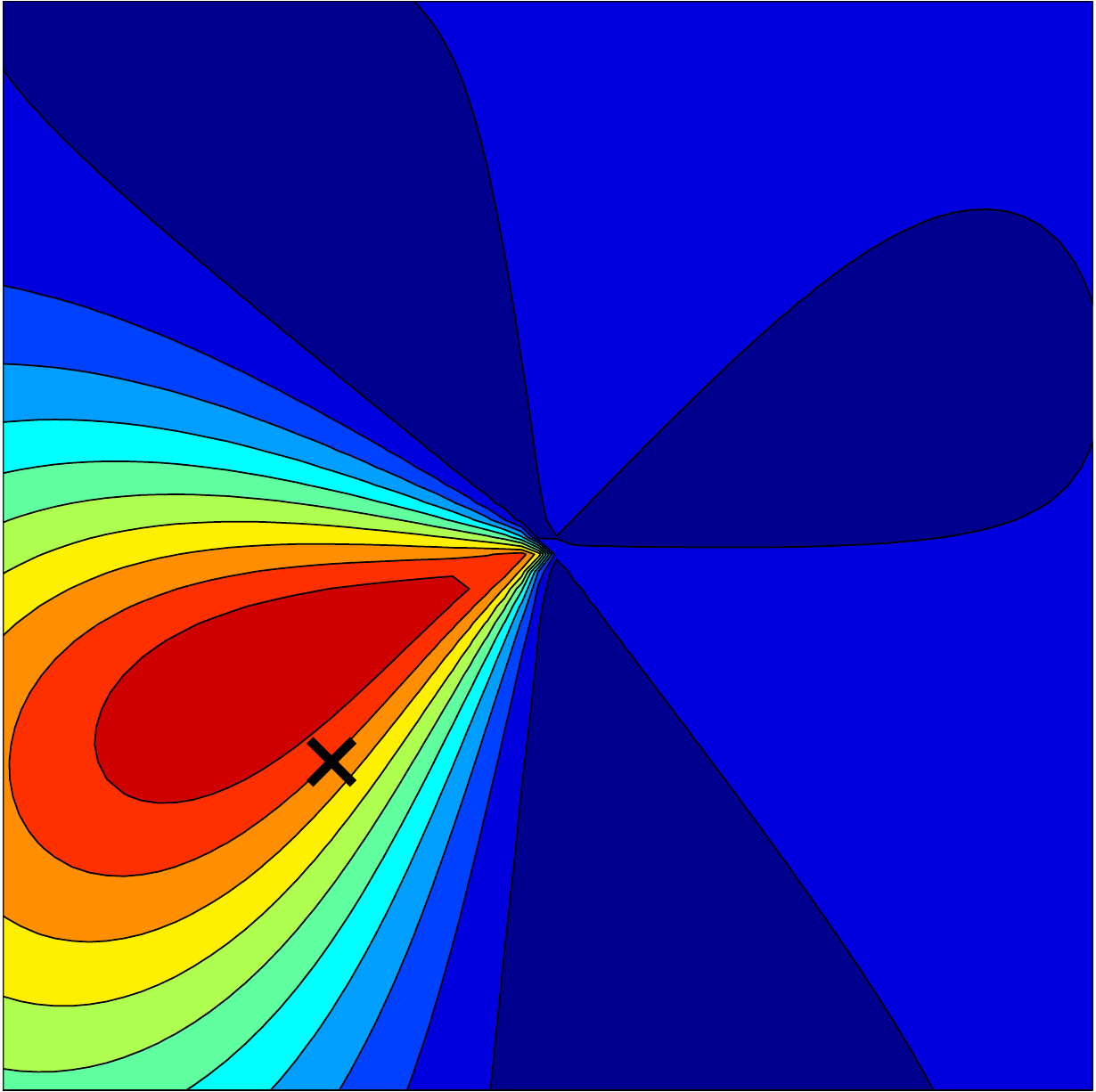}
\\
\end{tabular}
\vspace{10pt}
\caption{Patch maps for different parametrizations and kernels. We present two parametrizations in polar and two in cartesian coordinates, with absolute or relative gradient angle for each one.
The similarity between each pixel of patch \Q and a single pixel \p is shown over patch \Q.
All pixels in \Q have the same gradient angle, which is shown in red arrows.
The position of pixel \p is shown with ``$\times$'' on the patch maps.
We show examples for $\Delta\theta$ equal to $0$ (top) and $\pi/8$ (bottom).
At the top of each column the kernels that are used (patch similarity) are shown.
The similarity is shown in a relative manner and, therefore, the absolute scale is missing.
Ten isocontours are sampled uniformly and shown in different color. 
\label{fig:parametrizations}}
\end{figure*}

%% file: figs/tex/pixel_arrow_1.tex
\tikzset{fontscale/.style = {font=\relsize{#1}}}
\tikzset{>=latex}

\begin{tikzpicture}[ scale=0.0170 ]

% pixel
\draw[->,green,line width=1pt] (-25,-20) to (0,-20);

\draw (63,63) rectangle (-63,-63);

\end{tikzpicture}

%% file: figs/tex/patch_arrow_1.tex
\tikzset{fontscale/.style = {font=\relsize{#1}}}
\tikzset{>=latex}

\begin{tikzpicture}[ scale=0.0165 ]

% patch
\foreach \x in {-45,-15,15,45}
  \foreach \y in {-45,-15,15,45}
    \draw[->,red,line width=1pt] (\x-10,\y) to (\x+10,\y);

\draw (63,63) rectangle (-63,-63);

\end{tikzpicture}

%% file: figs/tex/patch_arrow_2.tex
\tikzset{fontscale/.style = {font=\relsize{#1}}}
\tikzset{>=latex}

\begin{tikzpicture}[ scale=0.0165 ]

% patch
\foreach \x in {-45,-15,15,45}
  \foreach \y in {-45,-15,15,45}
    \draw[->,red,line width=1pt] (\x-7,\y+2) to (\x+18.48-7,\y-7.65+2);

\draw (63,63) rectangle (-63,-63);

\end{tikzpicture}

%% file: figs/tex/rotins.tex
\begin{figure*}
\centering
\begin{tabular}{@{\sssp}c@{\ssp}c@{\ssp}c@{\ssp}c@{\ssp}c@{\sssp}}
  Pixel \p & Patch \Q &   \multirow{2}{*}{\kp\hspace{-2pt}\kr\hspace{-2pt}\ktt} & & \\
$\p_{\theta} = -\pi/4$ &  $\q_{\theta} = -\pi/2, \forall \q \in \Q$ & & & \\
\input{figs/tex/pixel_arrow_0}
  & \input{figs/tex/patch_arrow_0.tex}
  % & \includegraphics[height=80pt, width=80pt]{figs/patchsim/quiver/quiver_patch12.png}
    & \includegraphics[height=80pt, width=80pt]{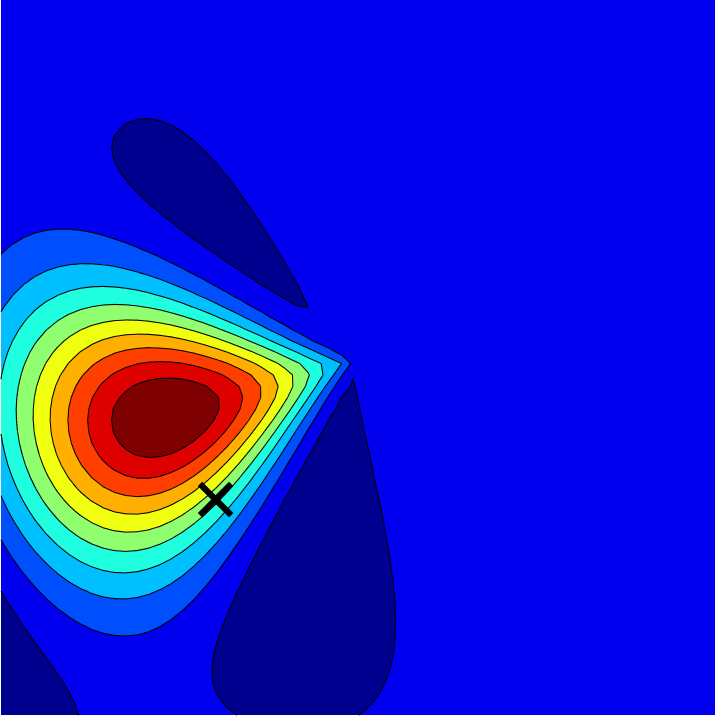}
      & \input{figs/tex/rot_diag}
        & \input{figs/tex/rotins_aa2r}
\\
\end{tabular}
\vspace{10pt}
\caption{Patch map with polar parametrization $\kp\hspace{-1pt}\kr\hspace{-1pt}\ktt$ for $\Delta\theta = \pi /4$ and the pair of toy pixel and patch on the left.
The example explains why the kernel undergoes shifting away from the position of pixel \p.
The diagram of the 4th column overlays pixel \p and 3 pixels of patch \Q with the same distance from the center as \p.
On the rightmost plot, we illustrate \ktt(\ptt, \qtt), \kp(\pp, \qp) for pixels \q with $\qr=\pr$ (on the black dashed circle).
\ktt is maximized at $\q^{(3)}$, \kp at $\q^{(1)}$, and their product at $\q^{(2)}$.
 \label{fig:rotins}}
\end{figure*}

%% file: figs/tex/pixel_arrow_0.tex
\tikzset{fontscale/.style = {font=\relsize{#1}}}
\tikzset{>=latex}

\begin{tikzpicture}[ scale=0.0225 ]

% pixel
\draw[->,green,line width=1pt] (-25,-25) to (-10,-40);

\draw (63,63) rectangle (-63,-63);

\end{tikzpicture}

%% file: figs/tex/patch_arrow_0.tex
\tikzset{fontscale/.style = {font=\relsize{#1}}}
\tikzset{>=latex}

\begin{tikzpicture}[ scale=0.0225 ]

% patch
\foreach \x in {-45,-15,15,45}
  \foreach \y in {-45,-15,15,45}
    \draw[->,red,line width=1pt] (\x,\y+10) to (\x,\y-10);

\draw (63,63) rectangle (-63,-63);

\end{tikzpicture}

%% file: figs/tex/rot_diag.tex
\tikzset{fontscale/.style = {font=\relsize{#1}}}
\tikzset{>=latex}

\begin{tikzpicture}[ scale=0.0225 ]
\node (origin) at (13,1) [fontscale=1] {\tiny$(0,0)$};
% outer circle
\draw[dashed,thick] (-43,0) arc (180:225:43);
% vectors
\foreach \x in {180,200,225}
	\draw[>=,blue,line width=.7pt] (0,0) to (\x:43);
%angles
\draw[>=](-35,0)arc(0:-90:7);
\draw (-28,-5) node [fontscale=1] {\miniscule$\nicefrac{\pi}{2}$};
\draw[>=](-25,-25)arc(45:-45:7);
\draw (-17,-27) node [fontscale=1] {\miniscule$\nicefrac{\pi}{2}$};
% labels
\draw (180:53) node [fontscale=1] {\tiny$q^{(3)}$};
\draw (200:55) node [fontscale=1] {\tiny$q^{(2)}$};
\draw (225:59) node [fontscale=1] {\tiny$q^{(1)}$};
% patch
\foreach \x/\y in {-43/0,-39.26/-14.7,-30.41/-30.41}
  \draw[->,red,line width=1pt] (\x,\y) to (\x,\y-15);
% pixel
\draw[->,green,line width=1pt] (-30.41,-30.41) to (-30.41+15,-30.41-15);

\draw (63,63) rectangle (-63,-63);

\end{tikzpicture}

%% file: figs/tex/rotins_aa2r.tex
\raisebox{-4pt}{
\begin{tikzpicture}
 \tikzstyle{every node}=[font=\footnotesize]
   \begin{axis}[%
      title={},
      height=112pt,
      width=140pt,
      xlabel={},
      ylabel={},
      legend pos=south east,
      legend cell align=left,
      legend style={font=\tiny, fill opacity=0.8, row sep=-3pt},
      ytick = {0, 1},
      yticklabels={{{\tiny $0$},{\tiny $1$}}},
      xtick={3.1416, 3.4212, 3.9270},
      xmin = 2.9452, xmax = 4.1233,
      xticklabels={{{\tiny $q^{(3)}$},{\tiny $q^{(2)}$},{\tiny $q^{(1)}$}}},
      y label style={at={(axis description cs:0.25,0.5)}},
      title style={at={(axis description cs:0.5,.8)}},
   ]
   \addplot[color=red  , style=solid , line width=1pt] table[x index=0,y index=1]{figs/rotins/approx_rel_angle_del.dat};
   \addlegendentry{$\ktt$}
   \addplot[color=green, style=solid , line width=1pt] table[x index=0,y index=1]{figs/rotins/approx_rel_angle_phi.dat};
   \addlegendentry{$\kp$}
   \addplot[color=blue , style=solid , line width=1pt] table[x index=0,y index=1]{figs/rotins/approx_rel_angle_res.dat};
   \addlegendentry{$\ktt\kp$}
   \addplot[color=red   , style=dotted, line width=1pt] coordinates {(3.1416,-.25) (3.1416,1.0   )};
   \addplot[color=blue  , style=dotted, line width=1pt] coordinates {(3.4212,-.25) (3.4212,0.6344)};
   \addplot[color=green , style=dotted, line width=1pt] coordinates {(3.9270,-.25) (3.9270,1.0   )};
   \end{axis}
\end{tikzpicture}
}

%% file: figs/tex/joint.tex
\begin{figure}
\centering
\begin{tabular}{ccc}
\kp\hspace{-2pt}\kr\hspace{-2pt}\ktt
& \kx\hspace{-2pt}\ky\hspace{-2pt}\kt
  & \kp\hspace{-2pt}\kr\hspace{-2pt}\ktt + \kx\hspace{-2pt}\ky\hspace{-2pt}\kt
\\
\includegraphics[height=65pt]{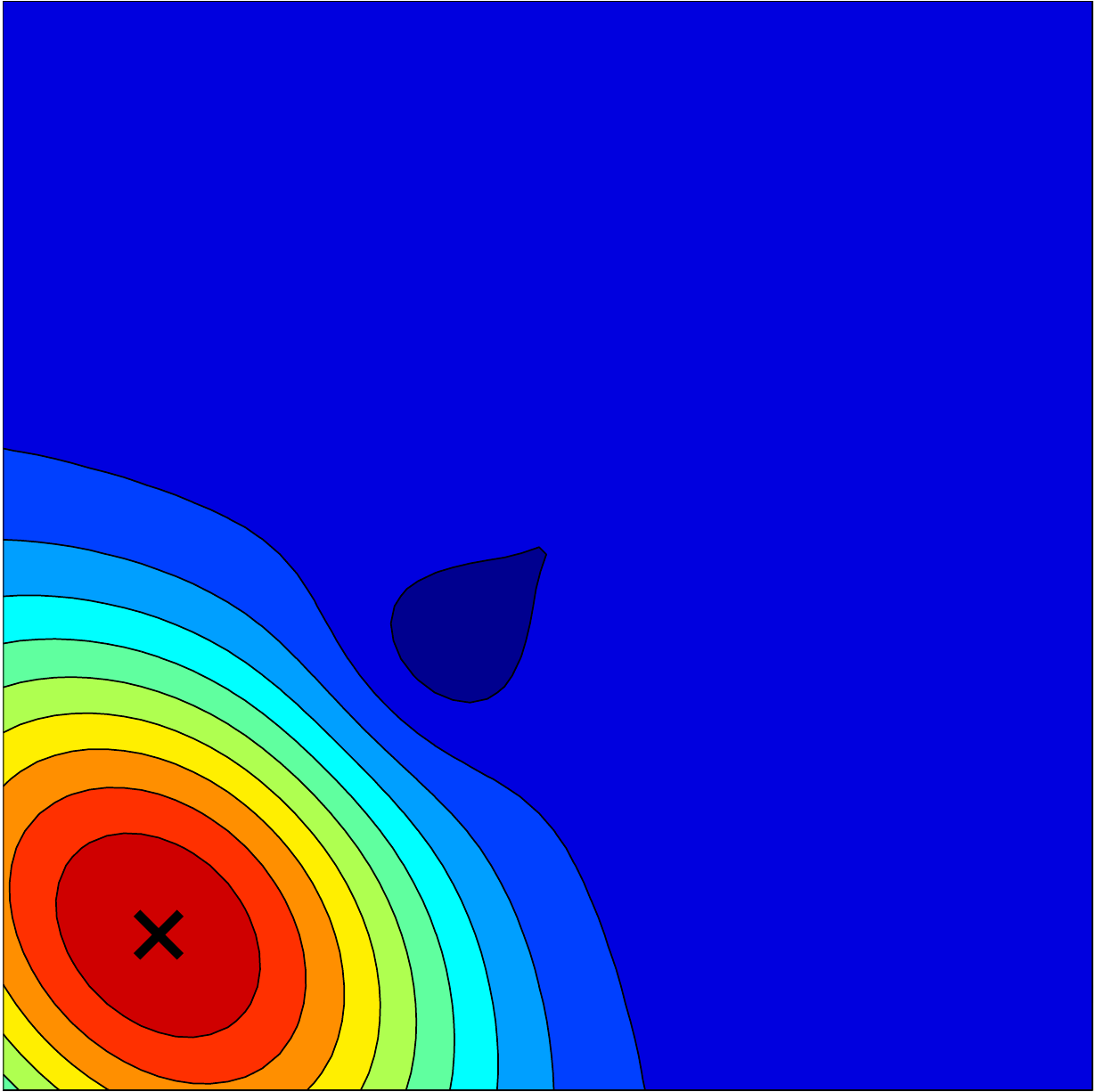}
& \includegraphics[height=65pt]{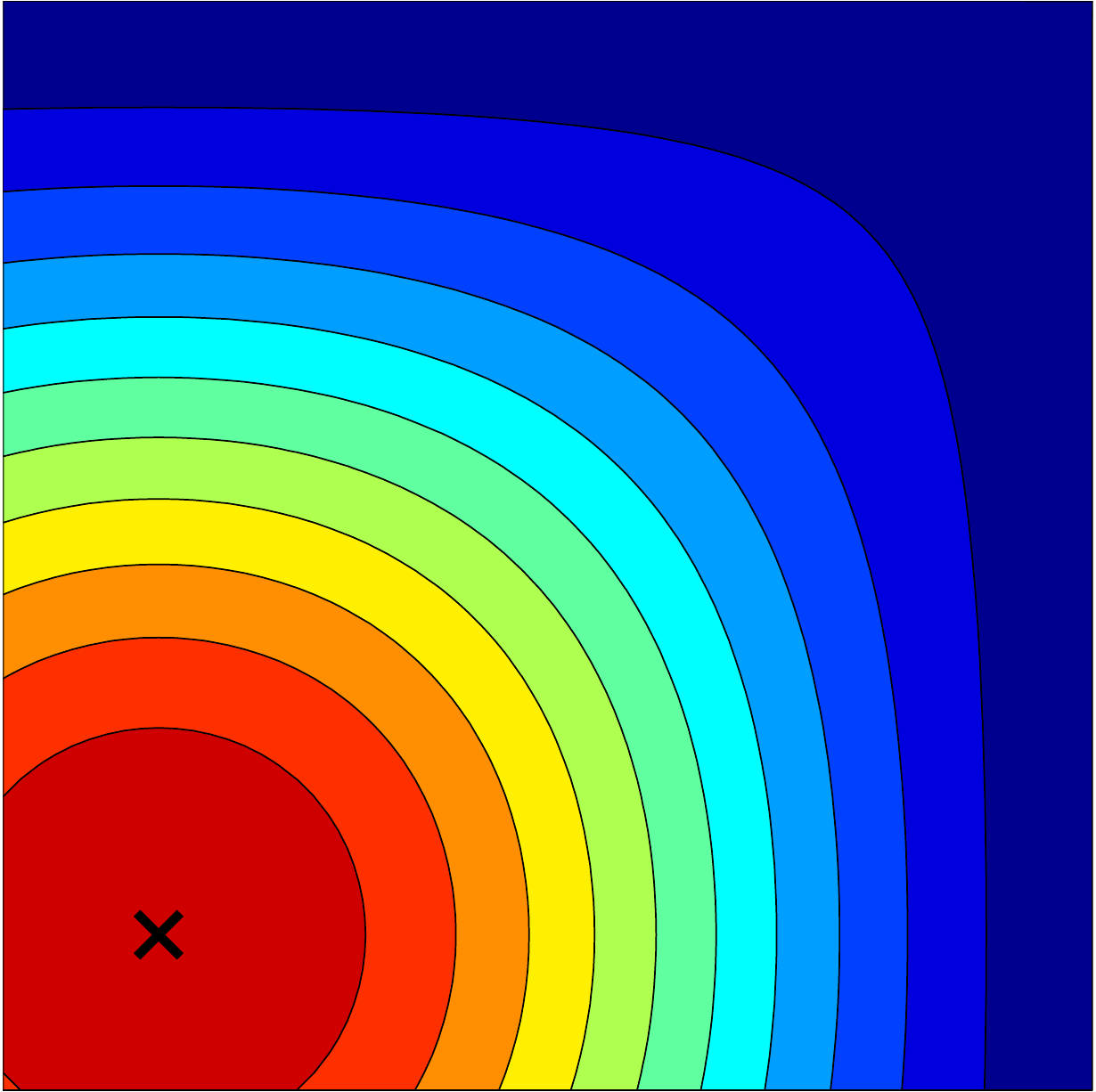}
  & \includegraphics[height=65pt]{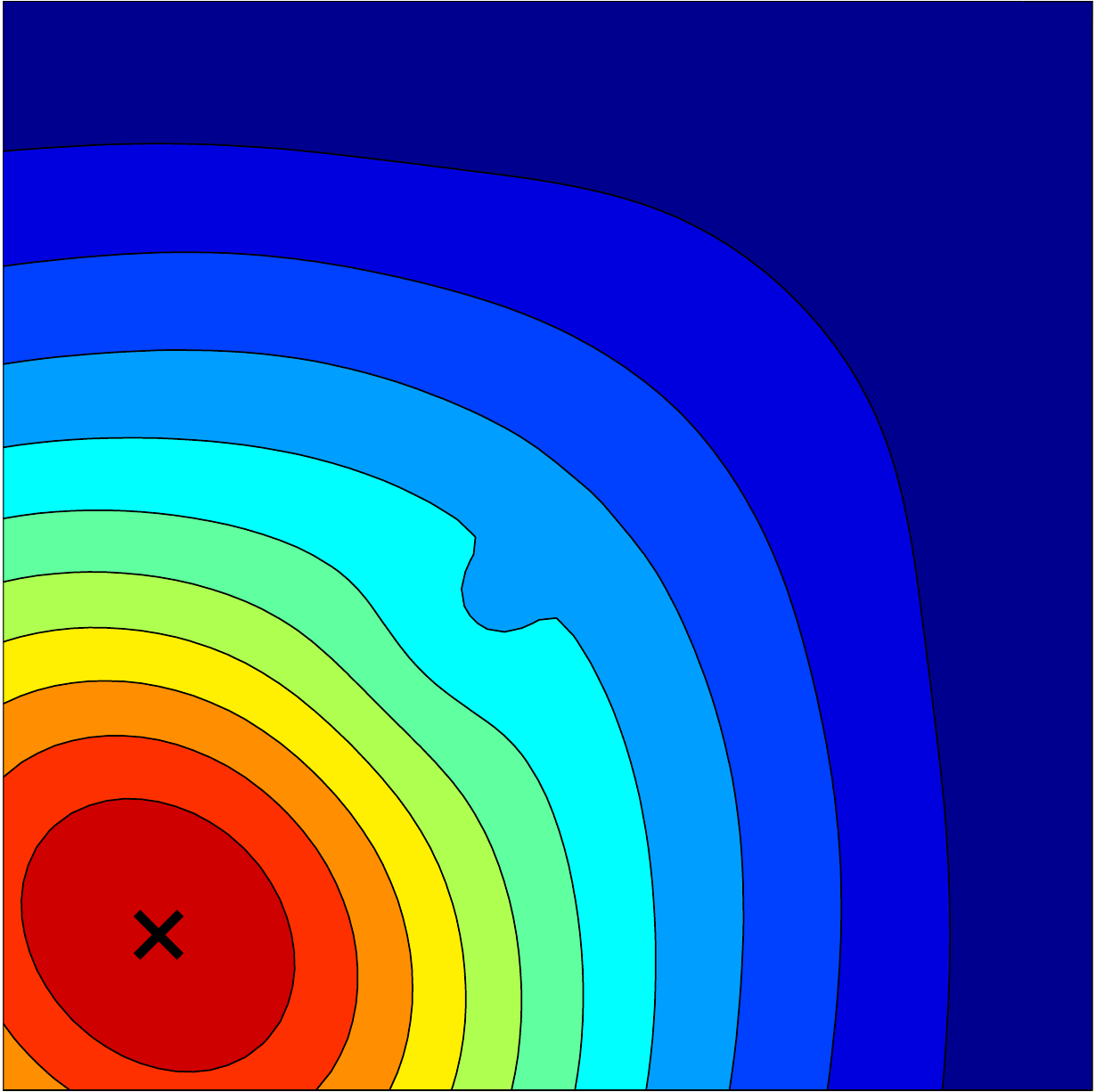}
\\
\includegraphics[height=65pt]{figs/patchsim/aa2r_rhofull_featuremap/p_20_20_d_0_0_.pdf}
& \includegraphics[height=65pt]{figs/patchsim/xya1_rhofull_smallkappa_featuremap/p_20_20_d_0_0_.pdf}
  & \includegraphics[height=65pt]{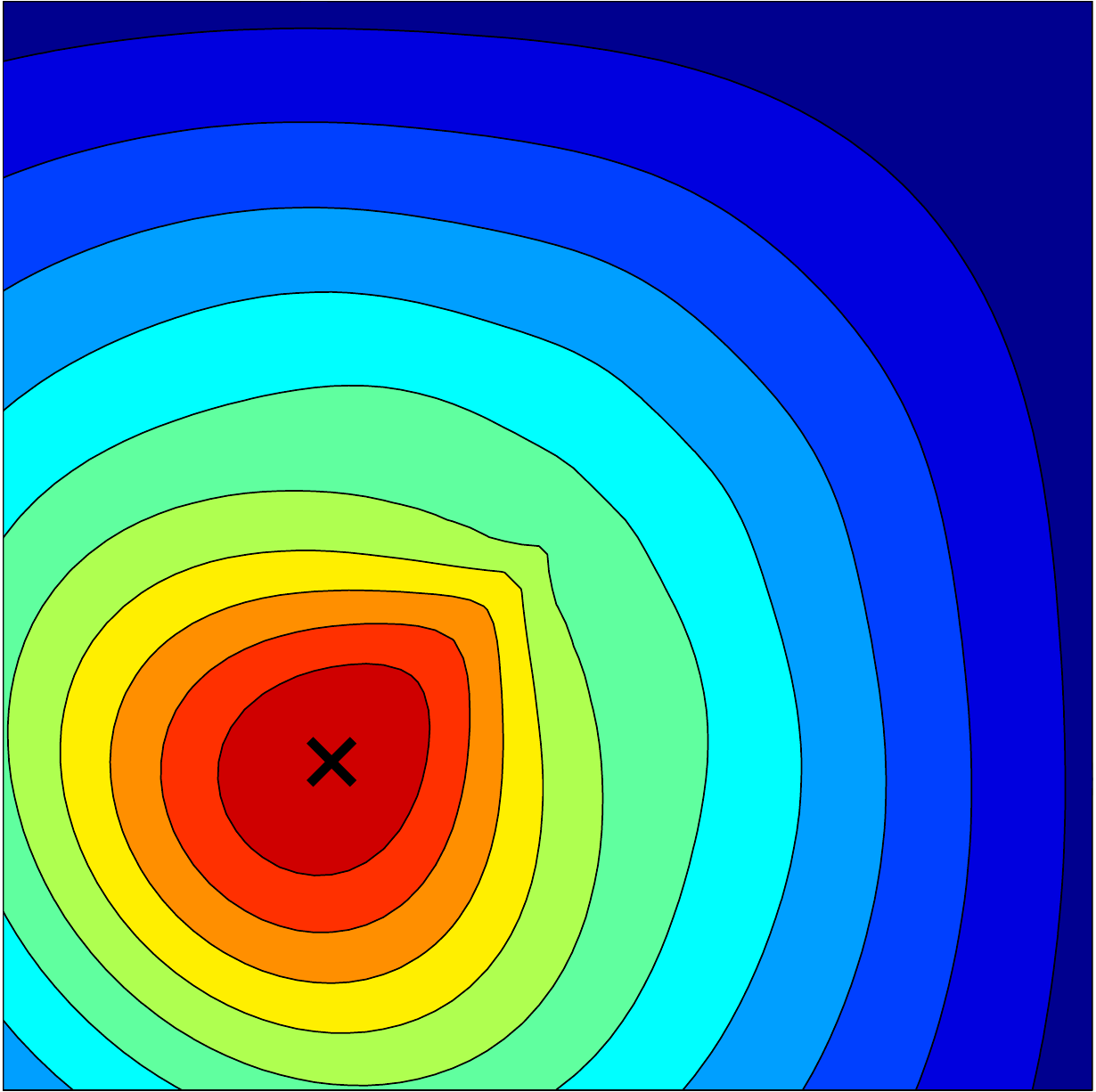}
\\
\includegraphics[height=65pt]{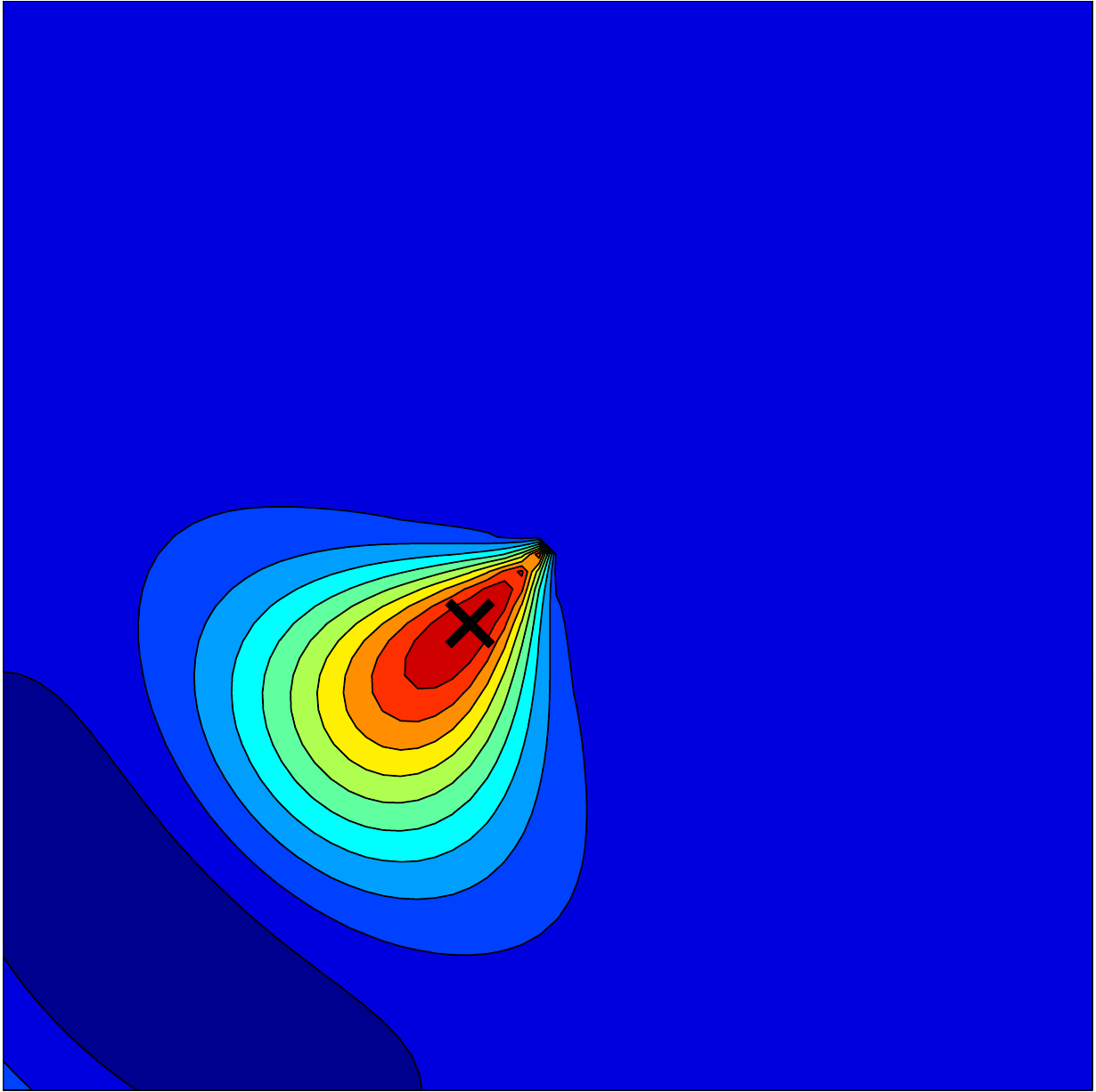}
& \includegraphics[height=65pt]{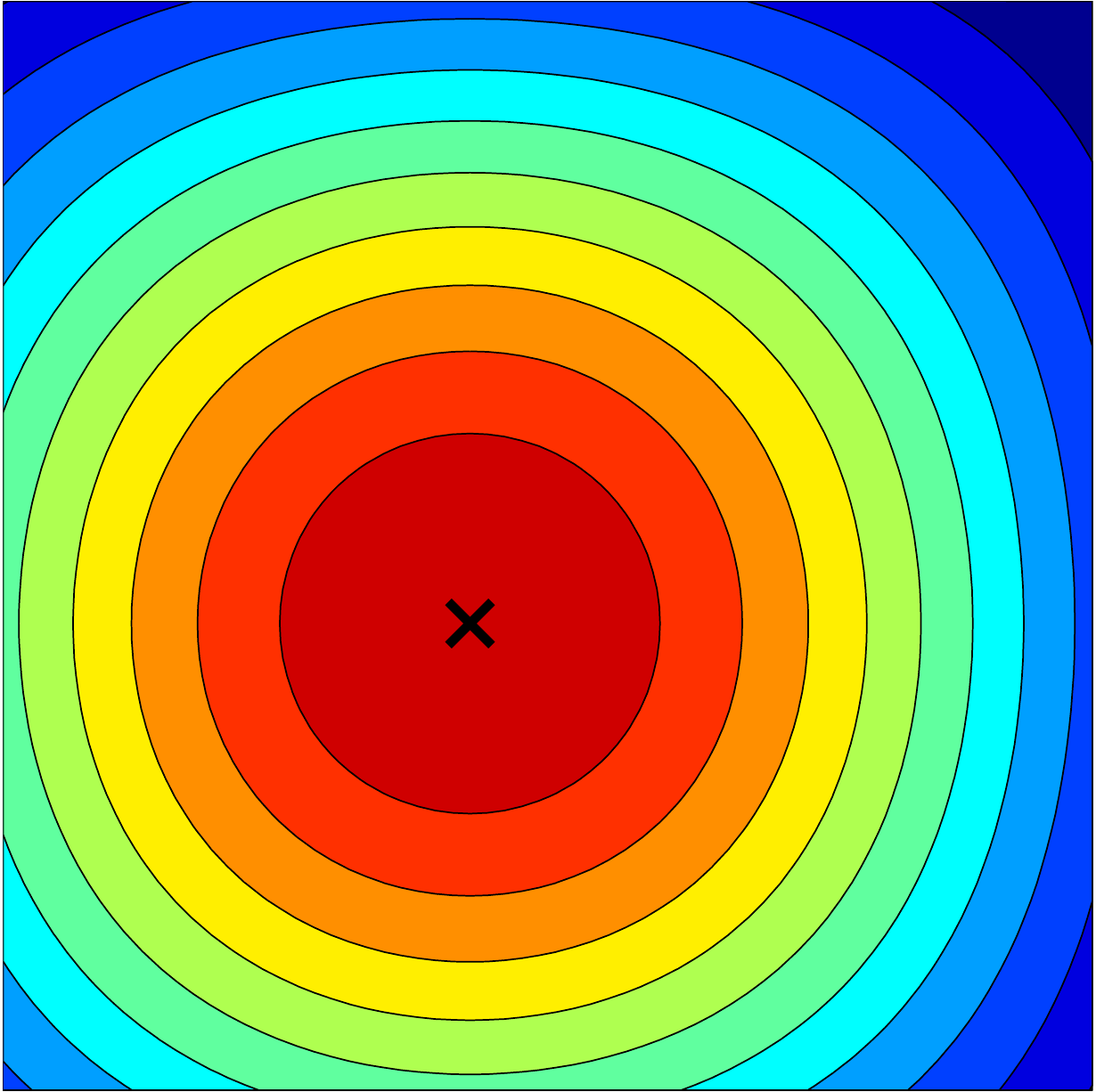}
  & \includegraphics[height=65pt]{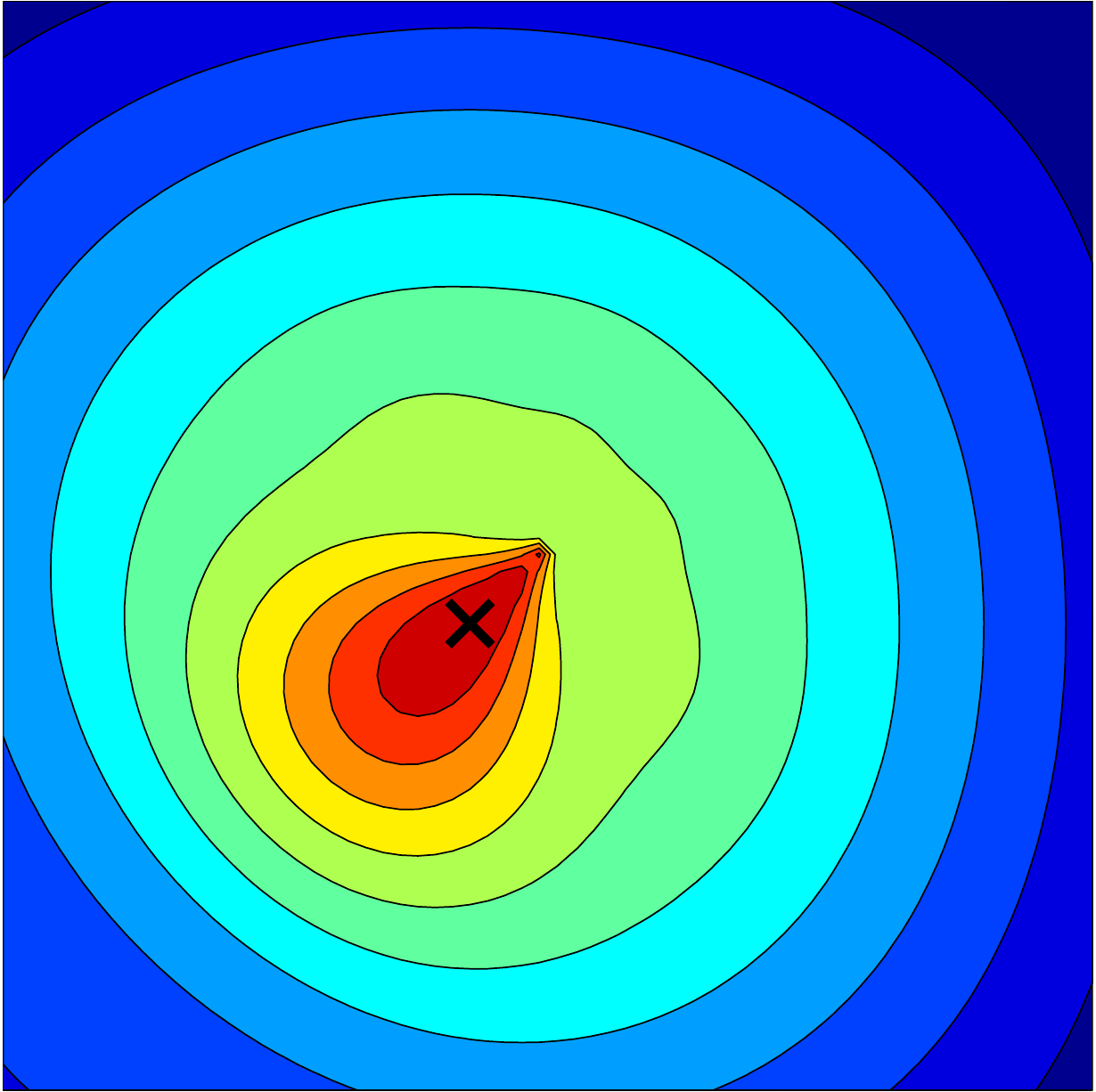}
\\
\end{tabular}
\vspace{10pt}
\caption{Patch maps for different pixels and parametrizations and their concatenation.
We present two parametrizations in polar and Cartesian coordinates, with relative and
absolute gradient angle, respectively. $\Delta\theta$ is fixed to be $0$ 
(individual values of \pt and \qt do not matter due to shift invariance)
and pixel \p is shown with ``$\times$''.
Note the behaviour around the centre in the concatenated case. 
Ten isocontours are sampled uniformly and shown in different color. 
\label{fig:joint}}
\end{figure}

%% file: figs/tex/learning.tex
\def\spl{\hspace{-1pt}+\hspace{-1pt}}
\def\cart{C}
\def\pol{P}

\newcommand{\tabintab}[2]{\begin{tabular}{c}#1\\#2\end{tabular}}

\begin{figure*}
\vspace{10pt}
\small
\centering
\begin{tabular}{c@{\msp}c@{\msp}c@{\msp}c@{\msp}c@{\msp}c@{\msp}c@{\msp}}
    &  \pol\cart   & \pol\spl\lw  & \cart\spl\lw  & \pol\cart\spl\lw &  \pol\cart\spl\pcawt & \pol\cart\spl\pcaws  \\
\raisebox{10pt}{\tabintab{$\pt = 0$}{$\qt = 0, \forall \q \in \Q$}}
& \includegraphics[height=60pt]{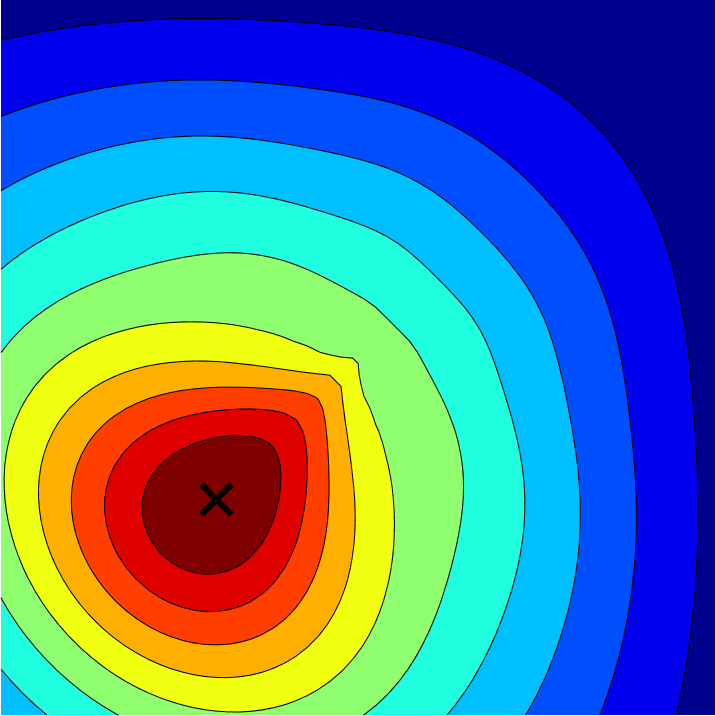}
& \includegraphics[height=60pt]{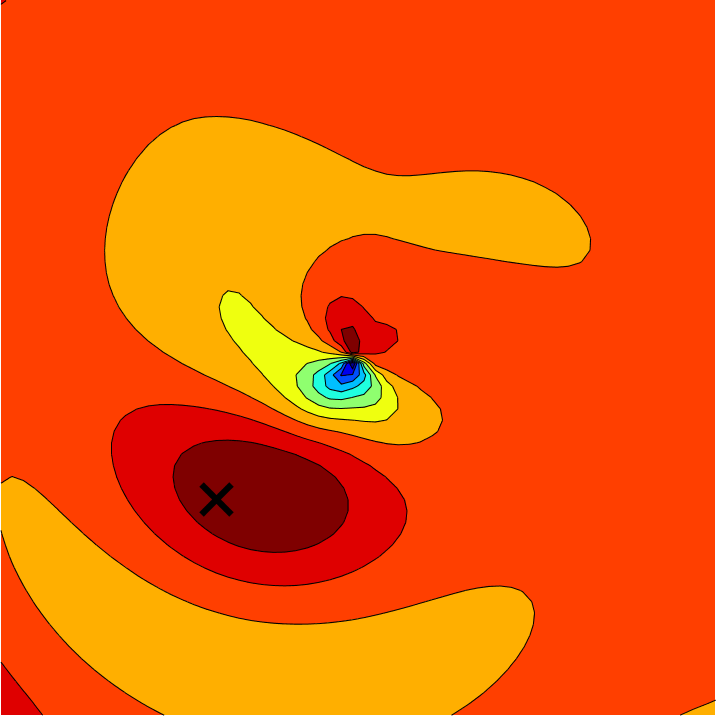}
& \includegraphics[height=60pt]{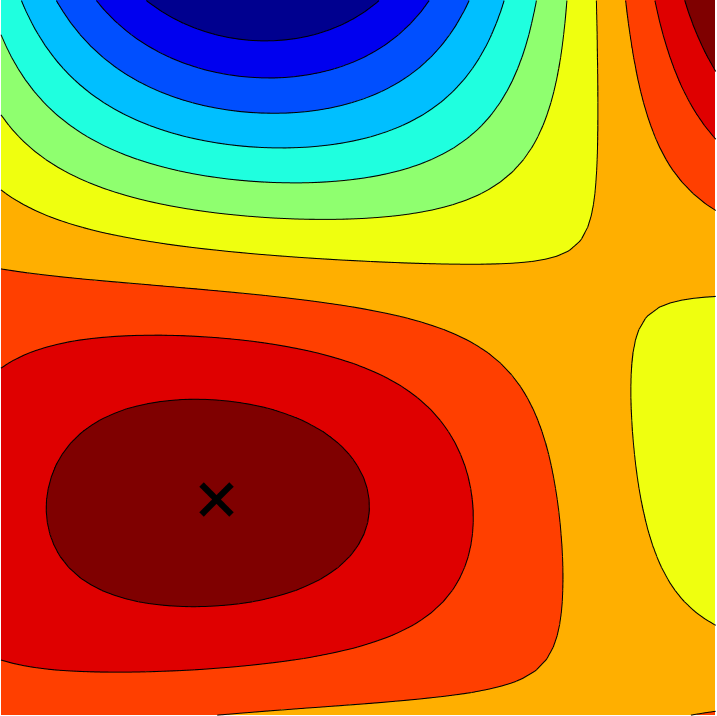}
& \includegraphics[height=60pt]{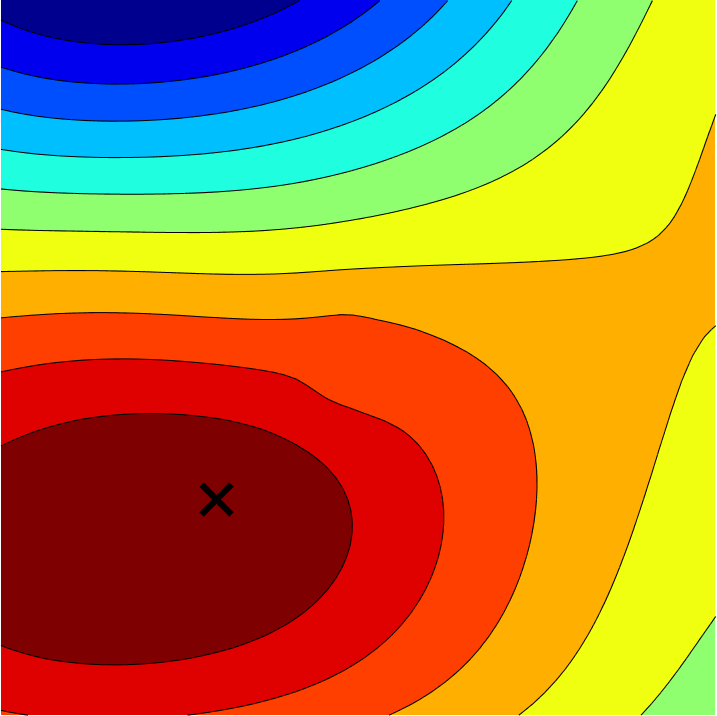}
& \includegraphics[height=60pt]{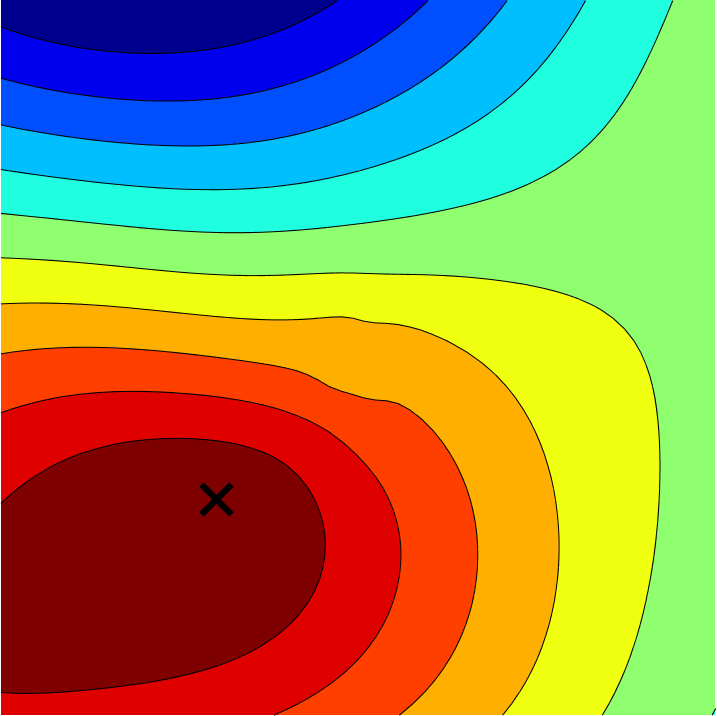}
& \includegraphics[height=60pt]{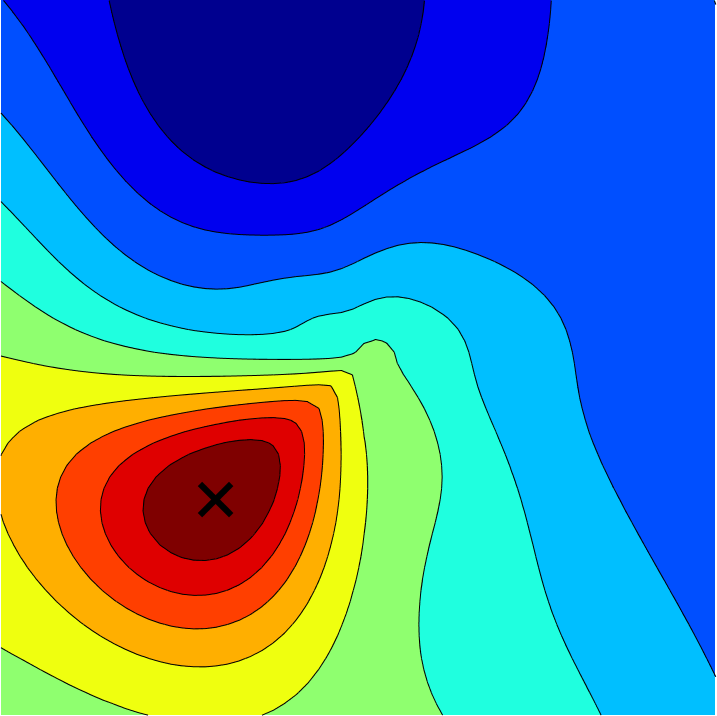} \\
\raisebox{10pt}{\tabintab{$\pt = \pi/4$}{$\qt = \pi/4, \forall \q \in \Q$}}
& \includegraphics[height=60pt]{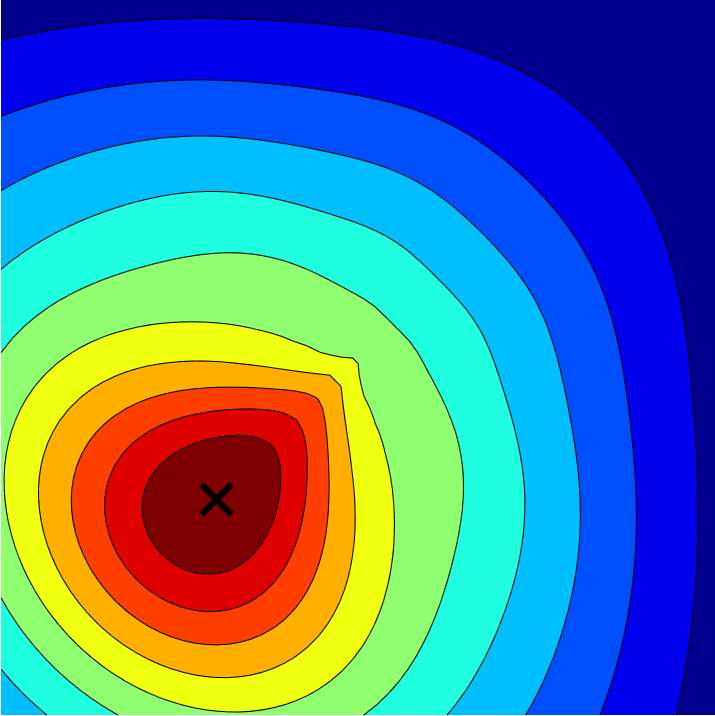}
& \includegraphics[height=60pt]{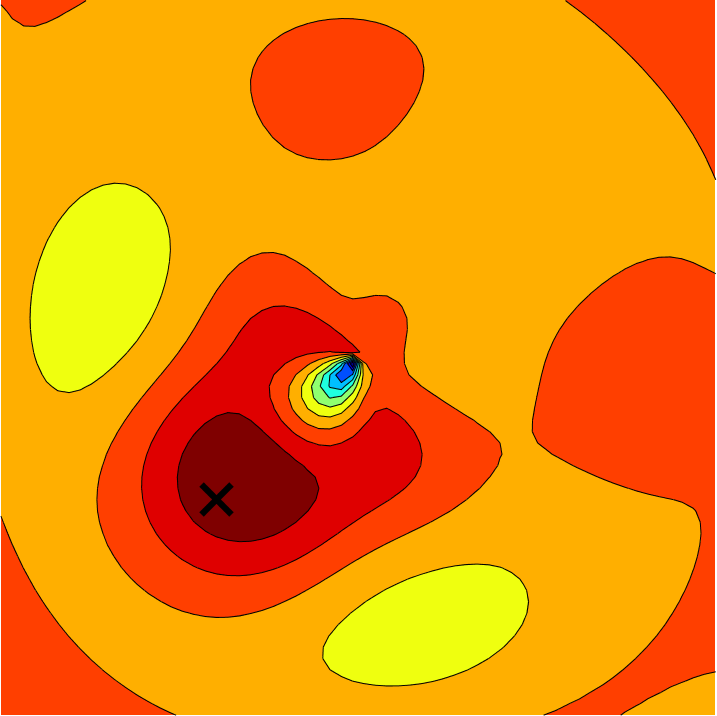}
& \includegraphics[height=60pt]{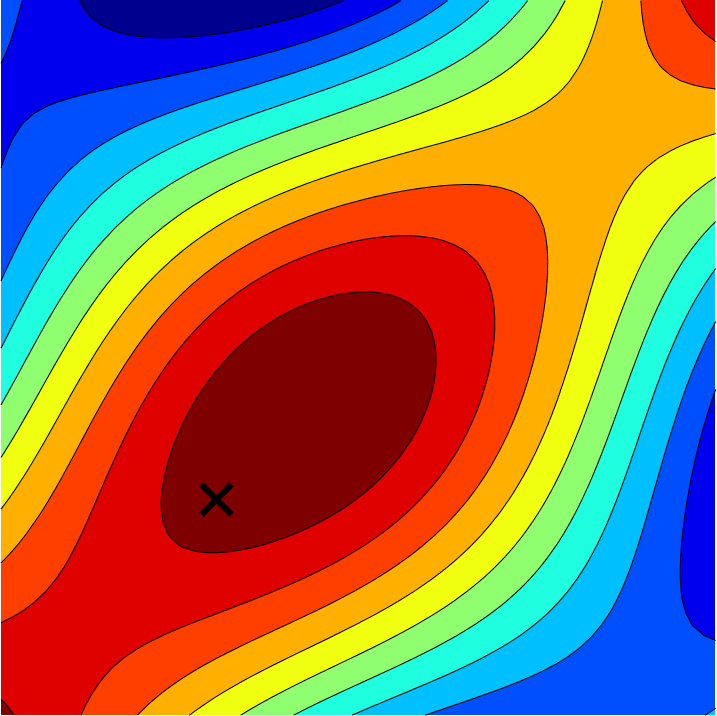}
& \includegraphics[height=60pt]{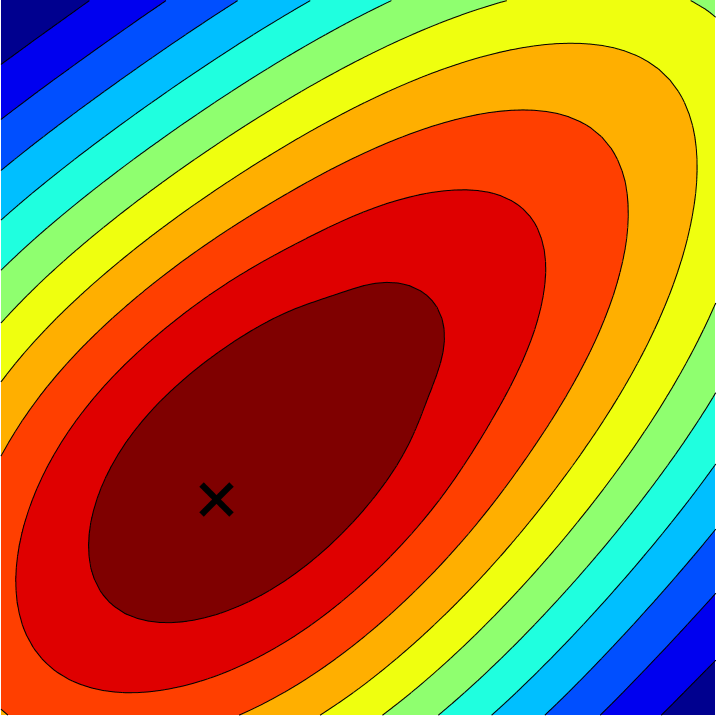}
& \includegraphics[height=60pt]{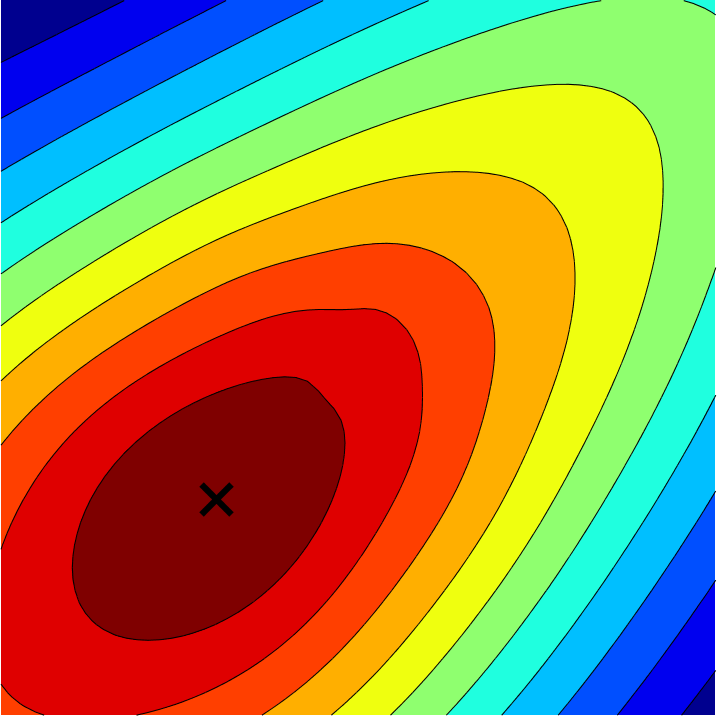}
& \includegraphics[height=60pt]{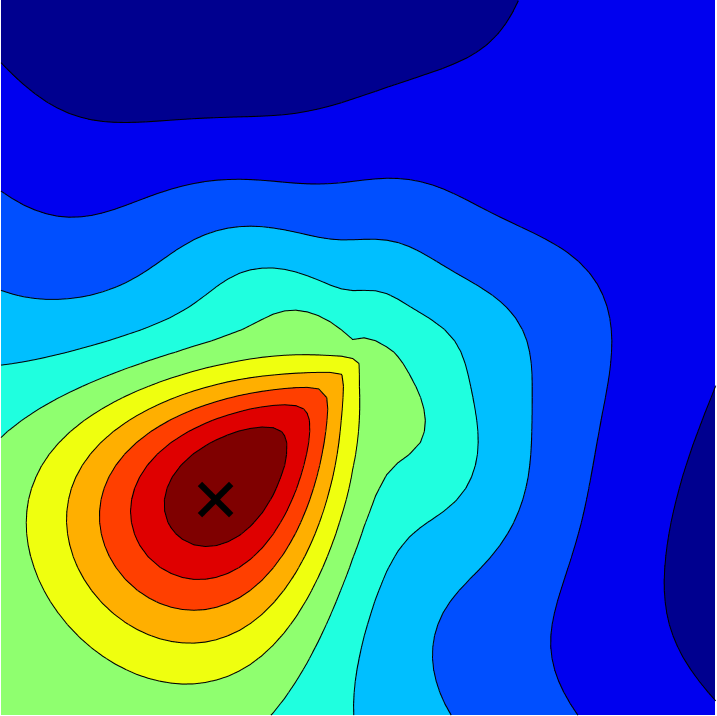} \\
\raisebox{10pt}{\tabintab{$\pt = \pi/2$}{$\qt = \pi/2, \forall \q \in \Q$}}
& \includegraphics[height=60pt]{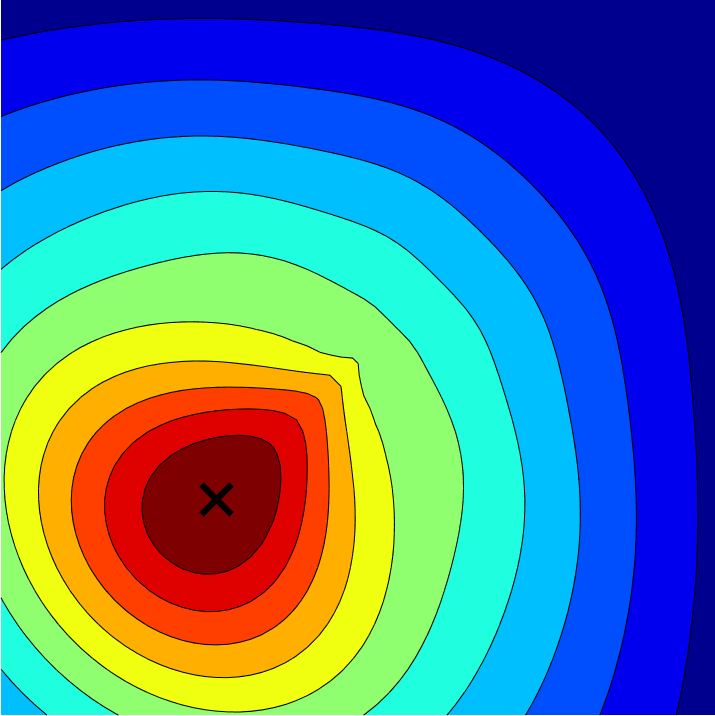}
& \includegraphics[height=60pt]{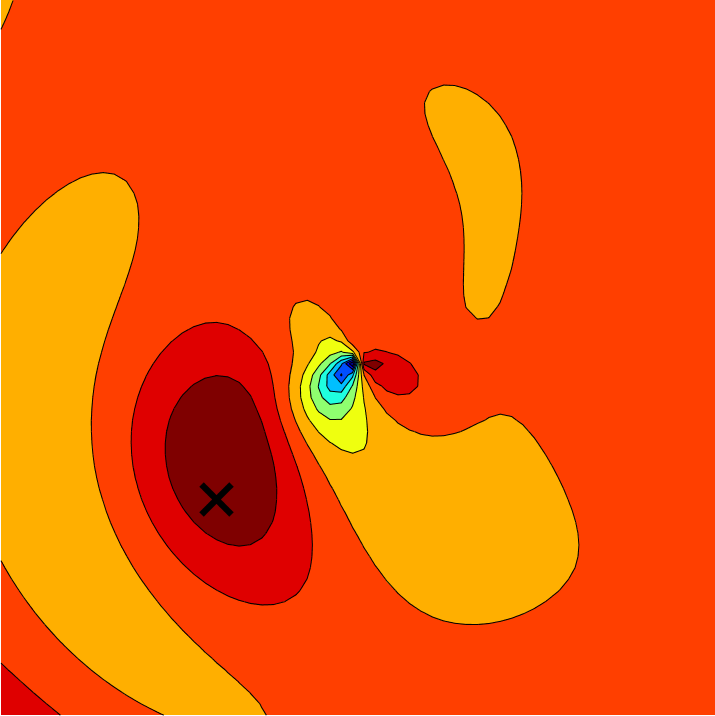}
& \includegraphics[height=60pt]{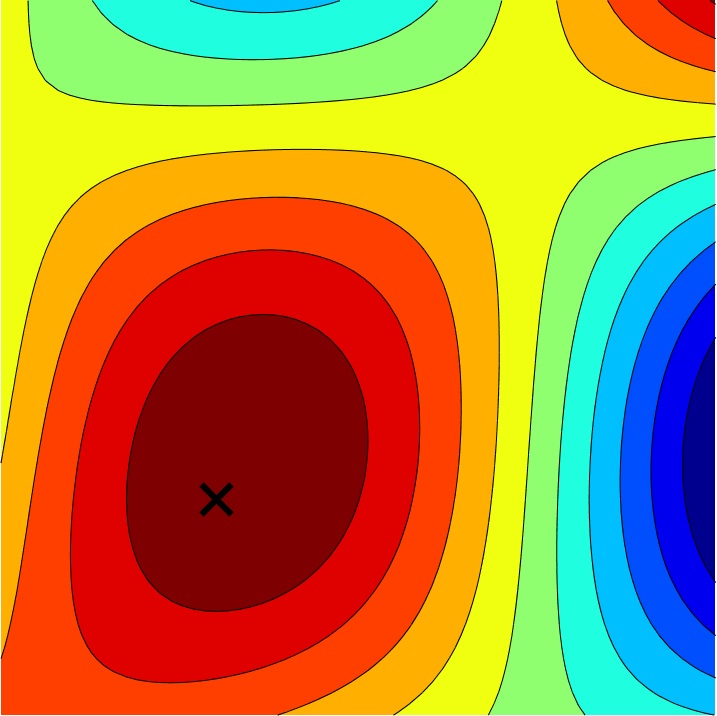}
& \includegraphics[height=60pt]{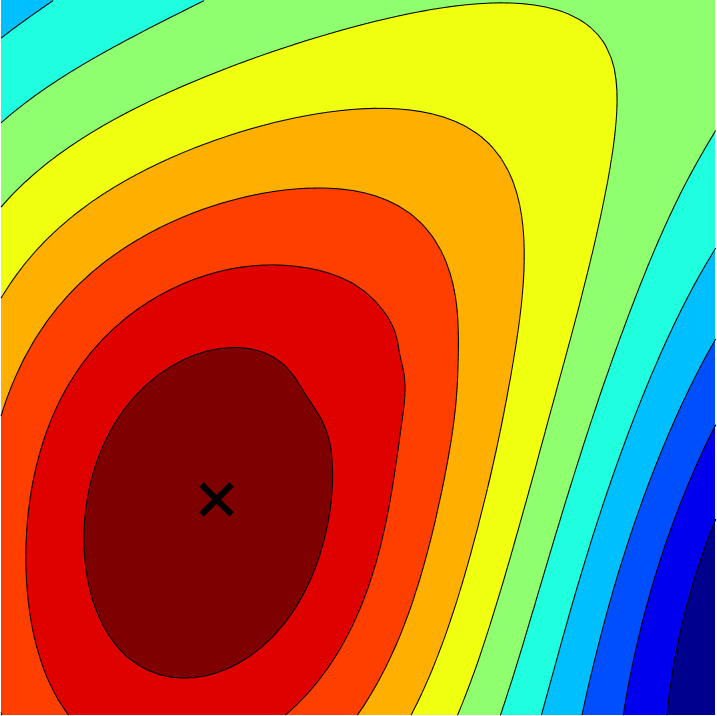}
& \includegraphics[height=60pt]{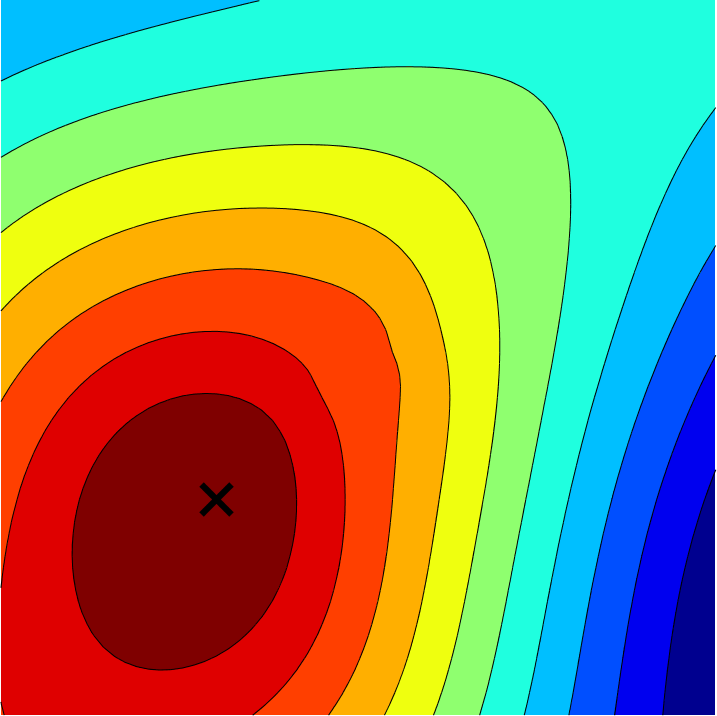}
& \includegraphics[height=60pt]{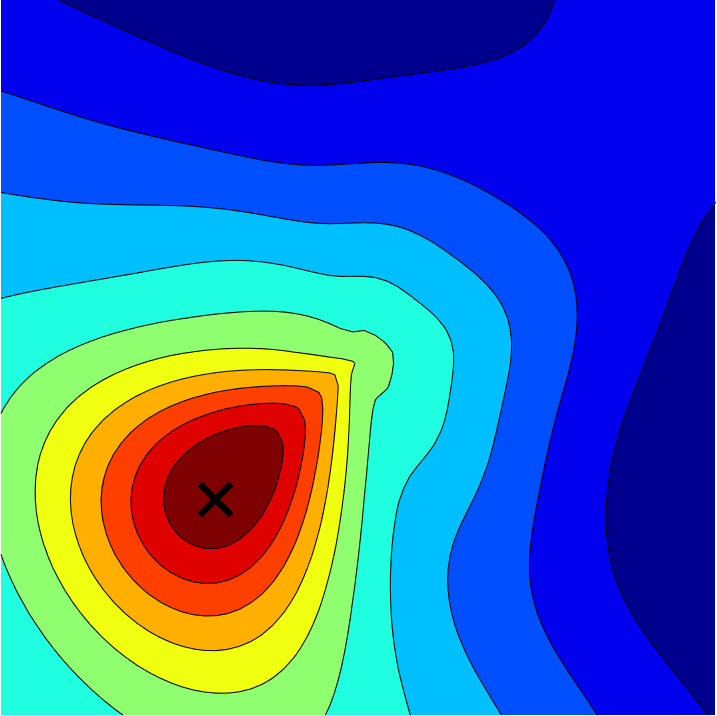} \\
\raisebox{10pt}{\tabintab{$\pt = 3\pi/4$}{$\qt = 3\pi/4, \forall \q \in \Q$}}
& \includegraphics[height=60pt]{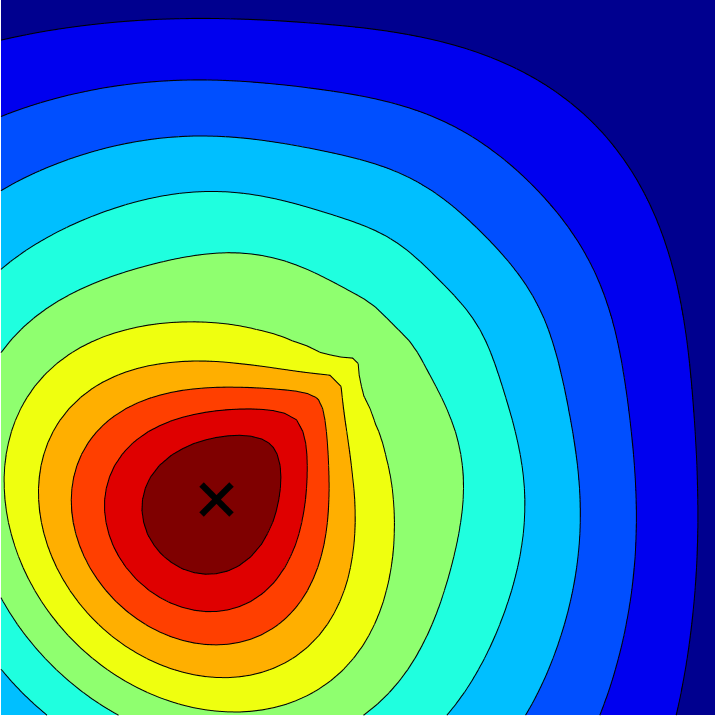}
& \includegraphics[height=60pt]{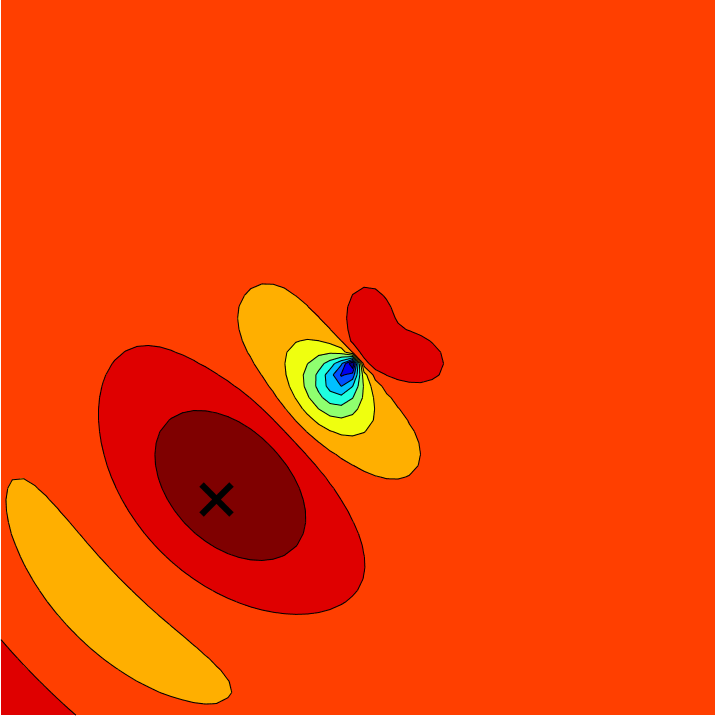}
& \includegraphics[height=60pt]{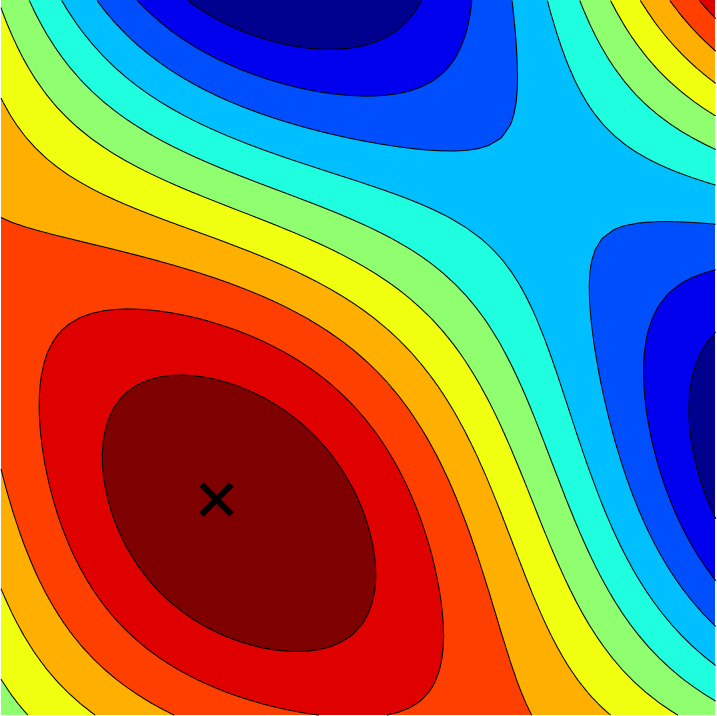}
& \includegraphics[height=60pt]{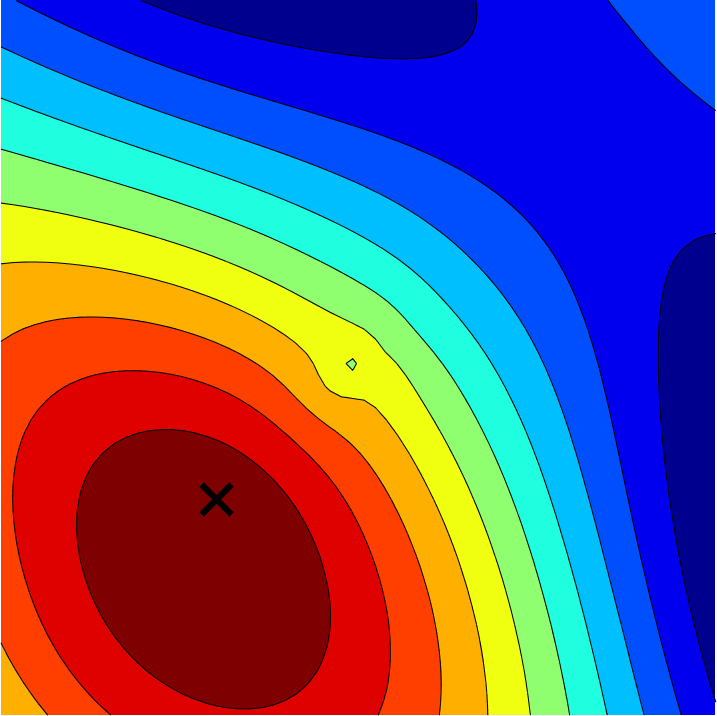}
& \includegraphics[height=60pt]{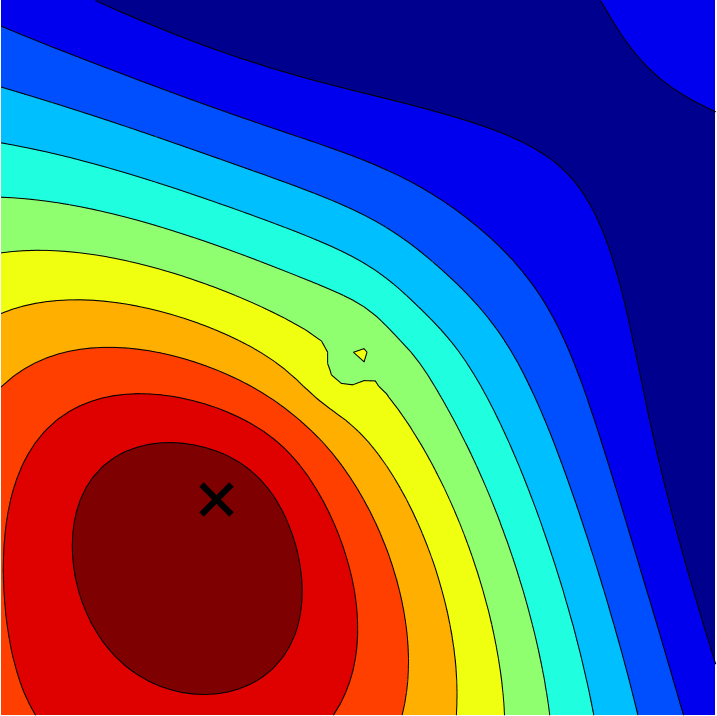}
& \includegraphics[height=60pt]{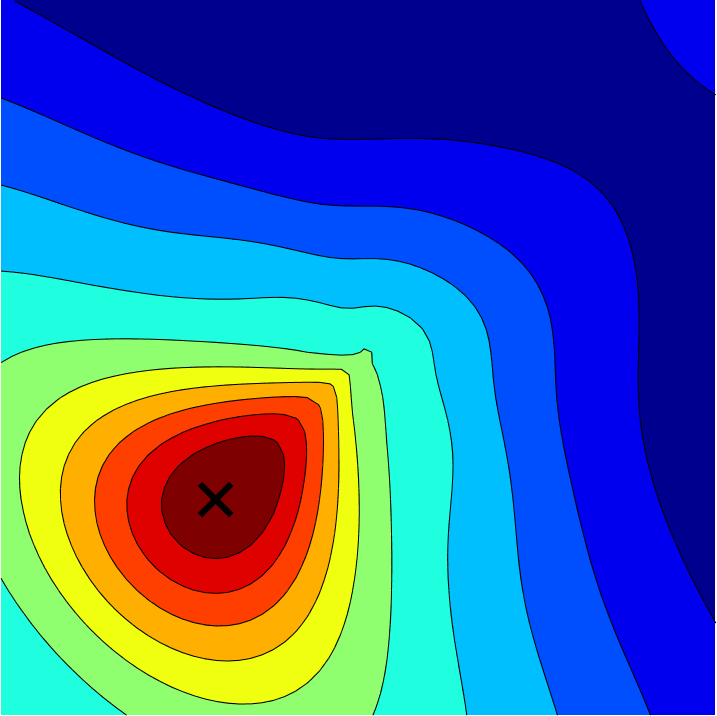} \\
\raisebox{10pt}{\tabintab{$\pt = \pi$}{$\qt = \pi, \forall \q \in \Q$}}
& \includegraphics[height=60pt]{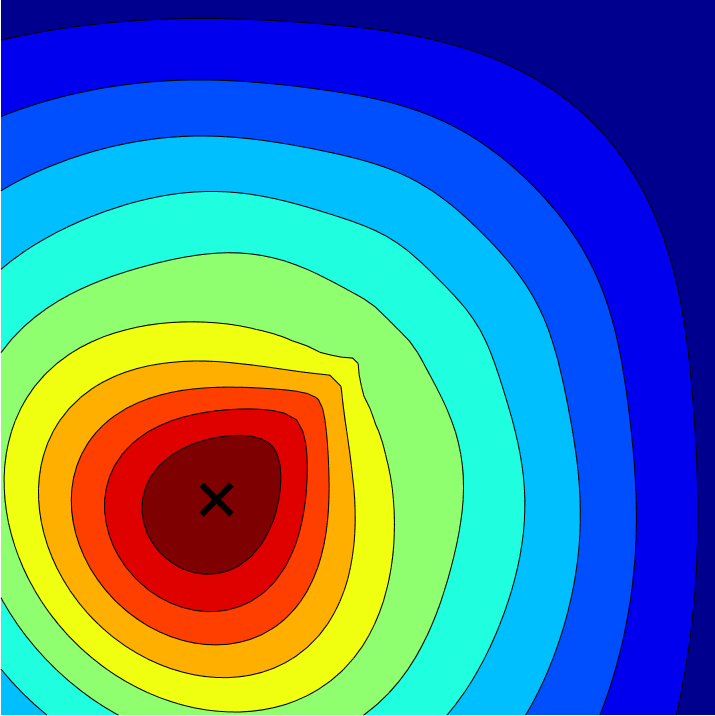}
& \includegraphics[height=60pt]{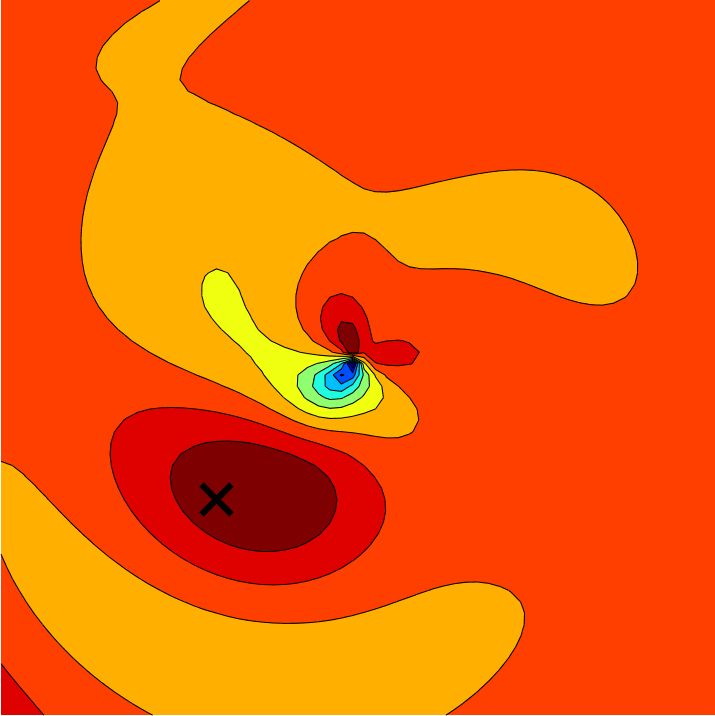}
& \includegraphics[height=60pt]{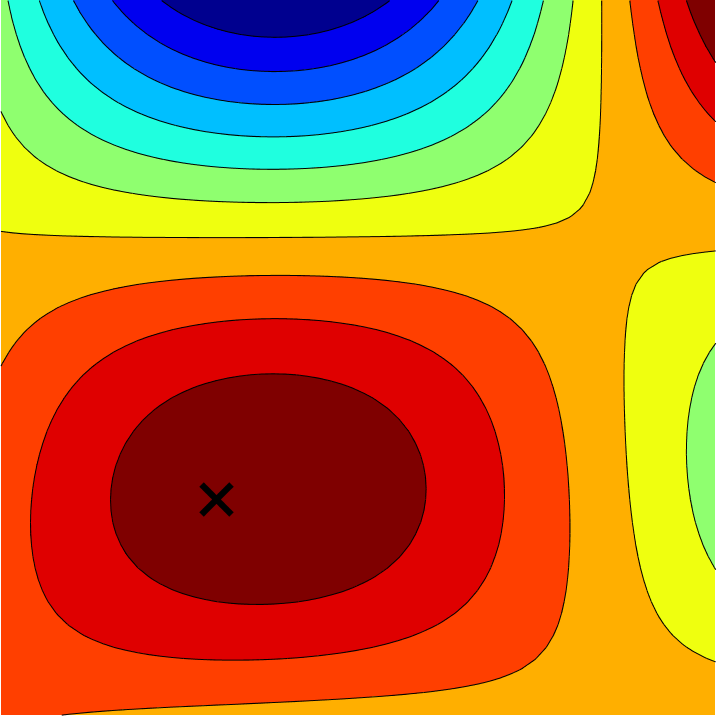}
& \includegraphics[height=60pt]{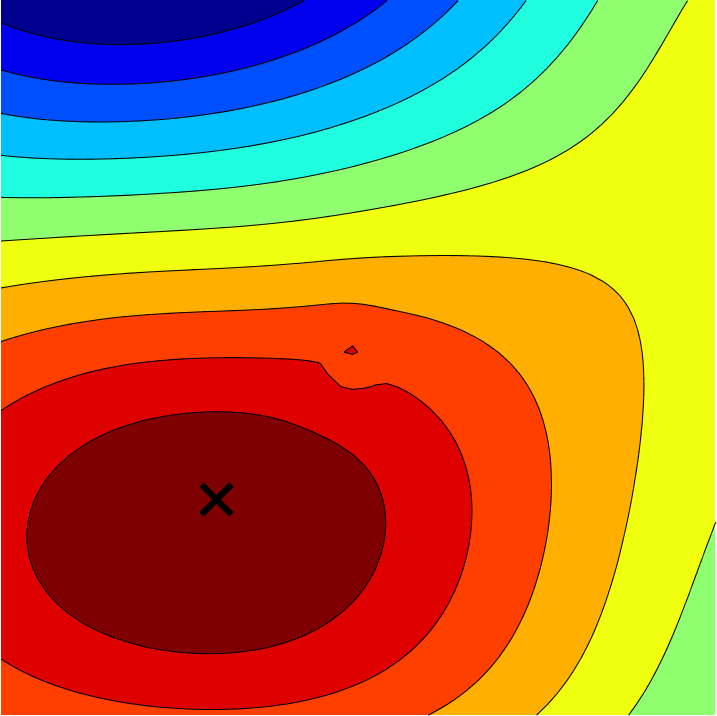}
& \includegraphics[height=60pt]{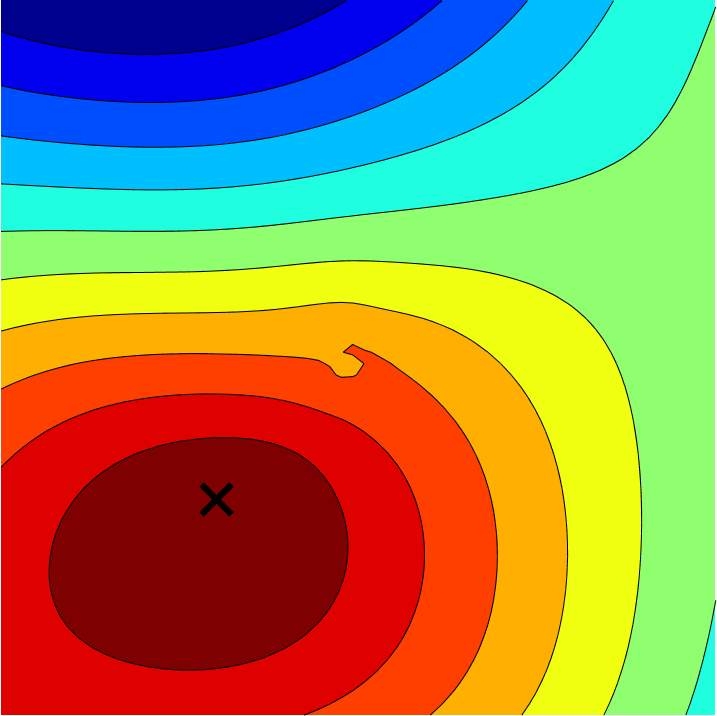}
& \includegraphics[height=60pt]{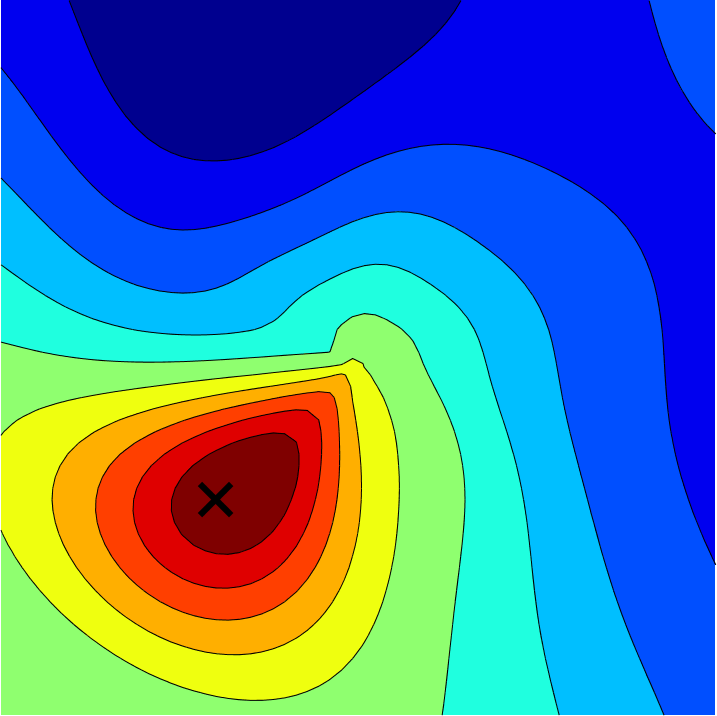} \\
\end{tabular}
\vspace{10pt}
\caption{Patch maps for different parametrizations, their concatenation, different post-processing methods,
and varying \pt and \qt, while $\Delta\theta$ is always 0.
Pixel \p is shown with ``$\times$''.
P: polar parametrization, C: cartesian parametrization, \pcawt \ is shown for $t=0.7$ and \pcaws \ for $\beta = \lambda_{40}$.
Whitening is learned on Liberty dataset. 
Observe that the similarity is no more shift invariant after the whitening and how the shape follows the angle of the gradients.
Ten isocontours are sampled uniformly and shown in different color. 
\label{fig:learning}}
\end{figure*}

%% file: figs/tex/sim1D.tex
\def\spl{\hspace{-1pt}+\hspace{-1pt}}
\def\cart{C}
\def\pol{P}

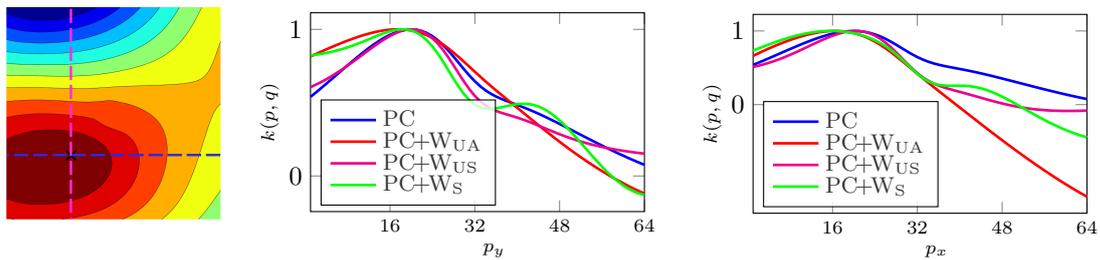
\begin{figure*}
\vspace{20pt}
\centering
\begin{tabular}{ccc}
\raisebox{17pt}{\includegraphics[height=80pt,width=80pt]{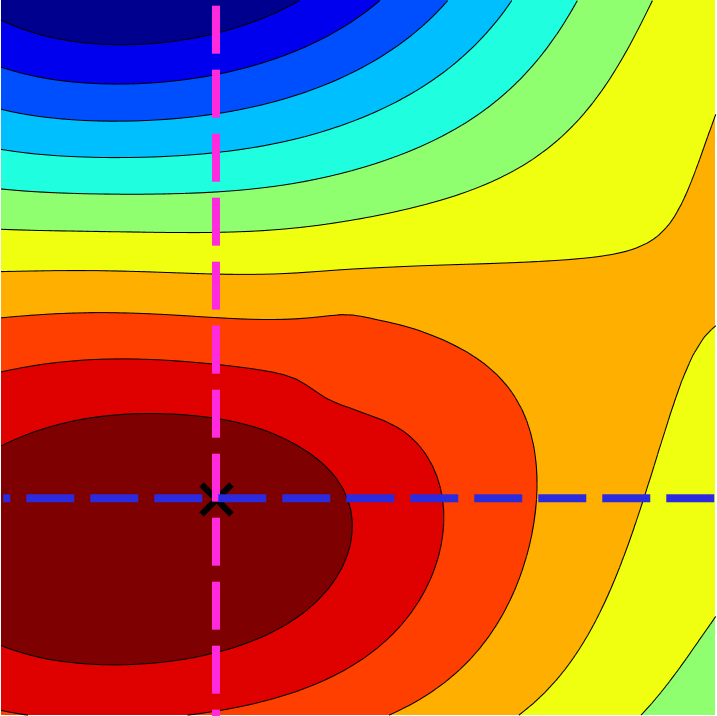}}%
&
\begin{tikzpicture}
 \tikzstyle{every node}=[font=\footnotesize]
   \begin{axis}[%
      title={},
      height=120pt,
      width=170pt,
      xlabel={\py},
      ylabel={$\k(\p, \q)$},
      legend pos=south west,
      legend cell align=left,
      legend style={font=\scriptsize, fill opacity=0.8, row sep=-3pt},
      ytick = {0, 1},
      yticklabels={{{\normalsize $0$},{\normalsize $1$}}},
      xmin = 1, xmax = 64,
      xtick = {16, 32, 48, 64},
      y label style={at={(axis description cs:-0.07,0.5)}},
      x label style={at={(axis description cs:0.55,-0.12)}},
      title style={at={(axis description cs:0.5,.8)}},
   ]
   \addplot[color=blue,style=solid,line width=1pt] table[x index=0,y index=1]{figs/sim1D/sx.dat};\addlegendentry{\pol\cart}
   \addplot[color=red,style=solid,line width=1pt] table[x index=0,y index=4]{figs/sim1D/sx.dat};\addlegendentry{\pol\cart\spl\pcawt}
   \addplot[color=magenta,style=solid,line width=1pt] table[x index=0,y index=5]{figs/sim1D/sx.dat};\addlegendentry{\pol\cart\spl\pcaws}
   \addplot[color=green,style=solid,line width=1pt] table[x index=0,y index=6]{figs/sim1D/sx.dat};\addlegendentry{\pol\cart\spl\lw}
   \end{axis}
\end{tikzpicture}%
&
\begin{tikzpicture}
 \tikzstyle{every node}=[font=\footnotesize]
   \begin{axis}[%
      title={},
      height=120pt,
      width=170pt,
      xlabel={\px},
      ylabel={$\k(\p, \q)$},
      legend pos=south west,
      legend cell align=left,
      legend style={font=\scriptsize, fill opacity=0.8, row sep=-3pt},
      ytick = {0, 1},
      yticklabels={{{\normalsize $0$},{\normalsize $1$}}},
      xmin = 1, xmax = 64,
      xtick = {16, 32, 48, 64},
      y label style={at={(axis description cs:-0.07,0.5)}},
      x label style={at={(axis description cs:0.55,-0.12)}},
      title style={at={(axis description cs:0.5,.8)}},
   ]
   \addplot[color=blue  , style=solid , line width=1pt] table[x index=0,y index=1]{figs/sim1D/sy.dat};\addlegendentry{\pol\cart}
   \addplot[color=red  , style=solid , line width=1pt] table[x index=0,y index=4]{figs/sim1D/sy.dat};\addlegendentry{\pol\cart\spl\pcawt}
   \addplot[color=magenta  , style=solid , line width=1pt] table[x index=0,y index=5]{figs/sim1D/sy.dat};\addlegendentry{\pol\cart\spl\pcaws}
   \addplot[color=green  , style=solid , line width=1pt] table[x index=0,y index=6]{figs/sim1D/sy.dat};\addlegendentry{\pol\cart\spl\lw}
   \end{axis}
\end{tikzpicture}%
\\
\end{tabular}
\caption{Visualizing 1D slices of a patch map. Showing similarity $\k(\p,\q)$ for all pixels \q with $\qx = \px$ (middle figure) and $\qy = \py$ (right figure). It corresponds to 1D similarity across the dashed lines (magenta and blue, respectively) of the patch map on the left. The particular patch map on the left is only chosen as an illustrative example. We show similarity for the first row and for columns 1, 4, 5 and 6 of patch maps from Figure~\ref{fig:learning}. Pixel $\p$ has $\px = 20$ and $\py = 20$ with the origin considered at the bottom left corner.
 \label{fig:sim1D}}
\end{figure*}

%% file: figs/response/sample_resp.tex
\begin{figure*}
\centering
\begin{tabular}{@{\sssp}c@{\sssp}c@{\sssp}c@{\sssp}c@{\sssp}c@{\msp}c@{\sssp}c@{\sssp}c@{\sssp}c@{\sssp}c@{\sssp}}
   \Q & P & C & PC & PC+\lw & \P & P & C & PC & PC+\lw\\
     \includegraphics[height=47pt]{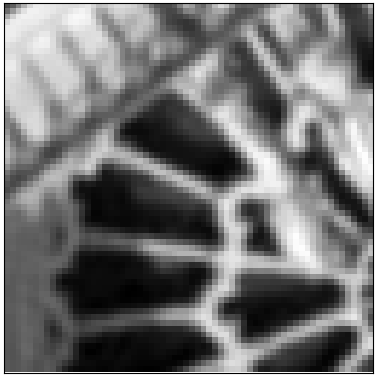}
   & \includegraphics[height=47pt]{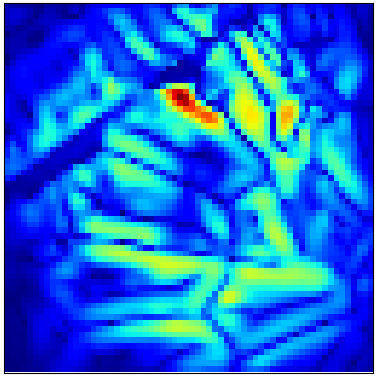}
   & \includegraphics[height=47pt]{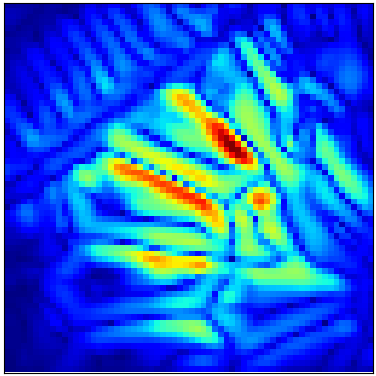}
   & \includegraphics[height=47pt]{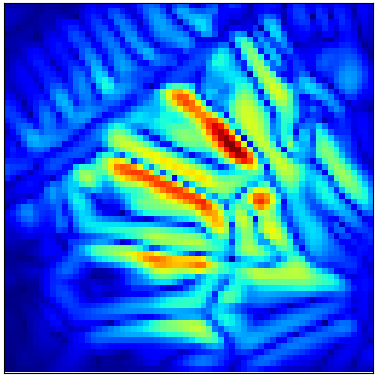}
   & \includegraphics[height=47pt]{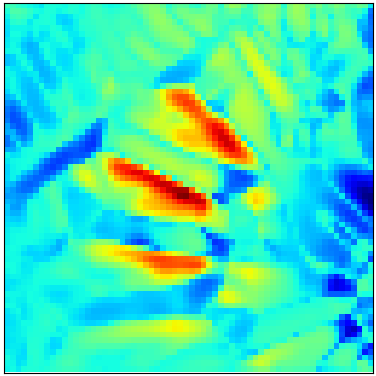}
   &
     \includegraphics[height=47pt]{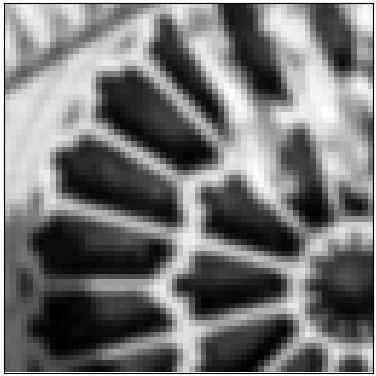}
   & \includegraphics[height=47pt]{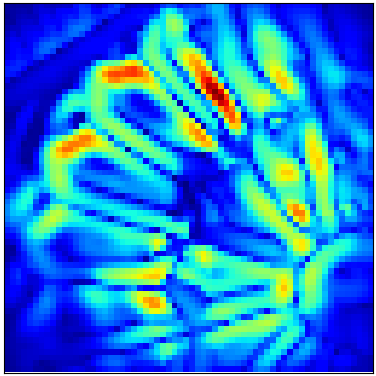}
   & \includegraphics[height=47pt]{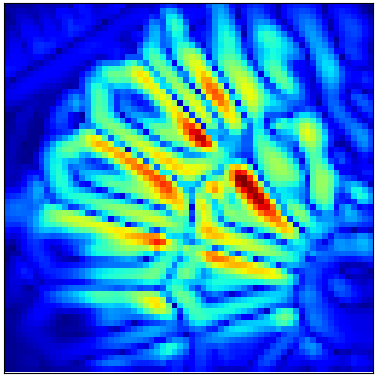}
   & \includegraphics[height=47pt]{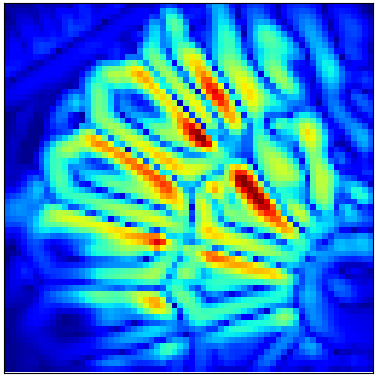}
   & \includegraphics[height=47pt]{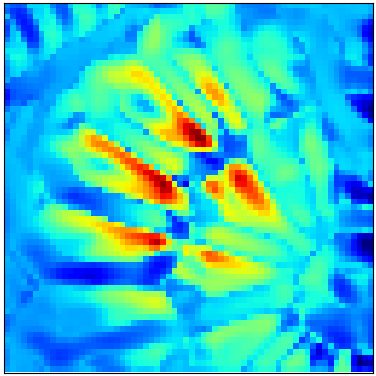}
   \\[1pt]
     \includegraphics[height=47pt]{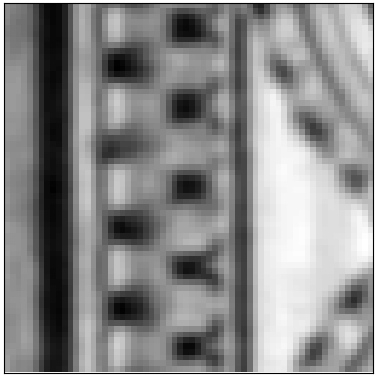}
   & \includegraphics[height=47pt]{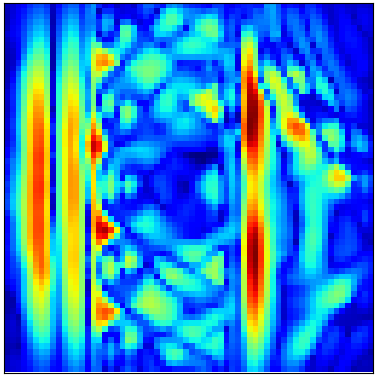}
   & \includegraphics[height=47pt]{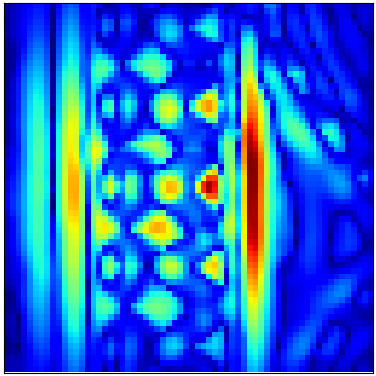}
   & \includegraphics[height=47pt]{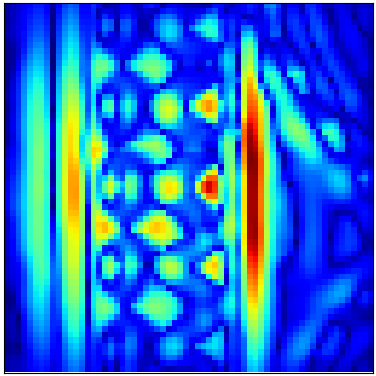}
   & \includegraphics[height=47pt]{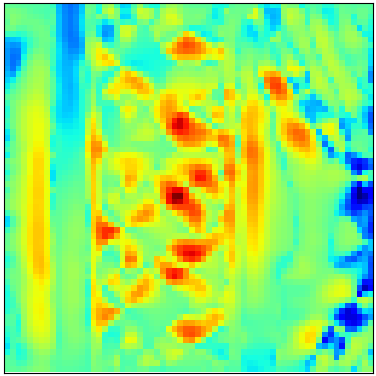}
   &
     \includegraphics[height=47pt]{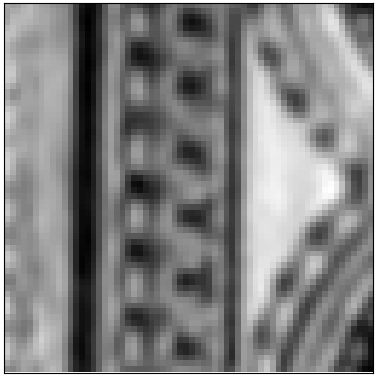}
   & \includegraphics[height=47pt]{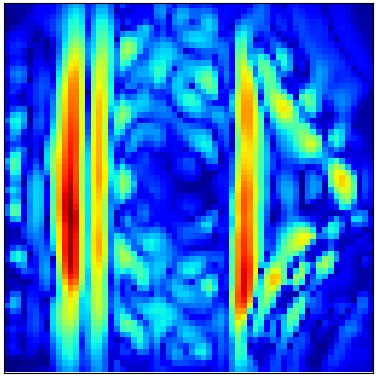}
   & \includegraphics[height=47pt]{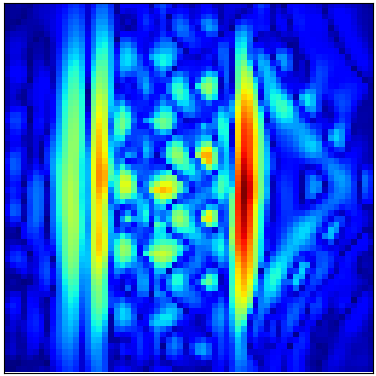}
   & \includegraphics[height=47pt]{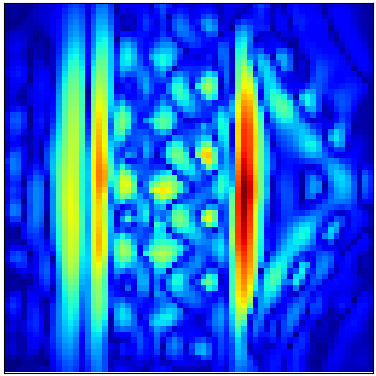}
   & \includegraphics[height=47pt]{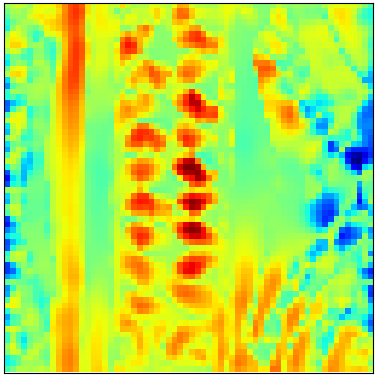}

\end{tabular}
\vspace{-3pt}

\caption{Positive patch pairs (patches \Q and \P) and the corresponding heat maps for polar (P), Cartesian (C), combined (PC), and whitened combined (PC+\lw) parametrization. Red (blue) corresponds to maximum (minimum) value. Heat maps on the left side correspond to $\sum_{\p \in \P}\M(\p,\q)$, while the ones on the right side to $\sum_{\q \in \Q}\M(\p,\q)$. In the case of PC+\lw, $\hat{\M}(\p,\q)$ is used instead of $\M(\p,\q)$. 
\label{fig:heatmaps}}
\end{figure*}

%% file: experiments.tex
\input{figs/tex/whitening_scores}
\section{Experiments}
\label{sec:exp}
We evaluate our descriptor on two benchmarks, namely the widely used
\emph{Phototourism} (PT) dataset~\cite{WB07}, and the recently released
\emph{HPatches} (HP) dataset~\cite{BLVM17}.
We first show the impact of the shrinkage parameters in
unsupervised whitening, and then compare with the baseline method
of Bursuc \etal~\cite{BTJ15} on top of which we build our descriptor.
We examine the generalization properties of whitening when learned on
PT but tested on HP, and finally compare against state-of-the-art descriptors
on both benchmarks.
In all our experiments with descriptor post-processing the dimensionality is
reduced to 128, while the combined descriptor original has 238 dimensions,
 except for the cases where the input descriptor is already of
lower dimension.
Our experiments are conducted with a Matlab implementation of the descriptor,
which takes 5.6 ms per patch for extraction on a single CPU on a 3.5GHz desktop machine.
A GPU implementation reduces time to 0.1 ms per patch on an Nvidia Titan X.

\subsection{Datasets and protocols.} The Phototourism dataset contains three sets
of patches, namely, Liberty (Li), Notredame (No) and Yosemite (Yo). Additionally,
labels are provided to indicate the 3D point that the patch corresponds to,
thereby providing supervision. It has been widely used for training and
evaluating local descriptors. Performance is measured by the false positive
rate at 95\% of recall (FPR95). The protocol is to train on one of the three
sets and test on the other two. An average over all six combinations is reported.

The HPatches dataset contains local patches of higher diversity, is more
realistic, and during evaluation the performance is measured on three tasks:
\emph{verification}, \emph{retrieval}, and \emph{matching}.
We follow the standard evaluation protocol~\cite{BLVM17} and report
mean Average Precision (mAP).
We follow the common practice and use models learned on Liberty of PT
to compare descriptors that have not used HP during learning.
We evaluate on all 3 train/test splits and report the average performance.
All reported results on HP (our and other descriptors) are produced
by our own evaluation by using the provided framework,
and descriptors\footnote{L2Net and HardNet descriptors were provided by the
authors of HardNet~\cite{MMR+17}.}.

\input{figs/tex/eigval}

\subsection{Impact of the shrinkage parameter.}
We evaluate the impact of the shrinkage parameter involved
in the unsupervised whitening. It is $t$ for \pcawt and $\beta=\lambda_i$ for \pcaws.
Results are presented in Figure~\ref{fig:paramshrink}
for evaluation on PT and HP dataset, while the whitening is learned on the same
or different dataset.
The performance is stable for a range of values, which makes it easy to tune in a robust
way across cases and datasets.
In the rest of our experiments we set $t = 0.7$ and $\beta= \lambda_{40}$.
In Figure~\ref{fig:eval} we show the eigenvalues used by \pcaw, \pcawt, and \pcaws.
The contrast between the larger and smaller eigenvalues is decreased.

\input{tables/ptscores}

\input{figs/response/hist}
\input{tables/hp_generalization}

\subsection{Comparison with the baseline.}

We compare the combined descriptor against the different parametrizations when used alone.
The experimental evaluation is shown in Table~\ref{tab:ptscores} for the PT dataset.
The baseline is followed by PCA and square-rooting, as originally proposed in ~\cite{BTJ15}.
We did not consider the square-rooting variant in our analysis
in Section~\ref{sec:method} because such non-linearity does not allow to visualize the
underlined patch similarity.
Supervised whitening on top of the combined descriptor performs the best.
Unsupervised whitening significantly improves too, while it does not require
any labeling of the patches.

Polar parametrization with the relative gradient direction (\baseline)
significantly outperforms the Cartesian parametrization with the absolute
gradient direction (\xya). After the descriptor post-processing
(\baseline + \lw vs.\ \xya + \lw), the gap is reduced. The performance of the
combined descriptor (\baseline + \xya) without descriptor post-processing is
worse than the baseline descriptor. That is caused by the fact, that the two
descriptors are combined with an equal weight, which is clearly suboptimal.
No attempt is made to estimate the mixing parameter explicitly.
It is implicitly included in the post-processing stage (see Appendix~\ref{sec:proof}).

Figure~\ref{fig:histograms} presents patch similarity histograms for matching and non-matching pairs, showing how their separation is improved by the final descriptor.

We perform an experiment with synthetic patch transformations to test the robustness of different parametrizations.
The whole patch is synthetically rotated or translated by appropriately transforming pixel position and gradient angle in the case of rotation.
A fixed amount of rotation/translation is performed for one patch of each pair of the PT dataset and results are presented in Figure~\ref{fig:rottra}.
It is indeed verified that the Cartesian parametrization is more robust to translations, while the polar one to rotations.
The joint one finally partially enjoys the benefits of both.

\input{figs/response/syn_rot}

\subsection{Generalization of whitening.}
We learn the whitening on PT or HP (supervised and unsupervised)
and evaluate the performance on HP.
We present results in Table~\ref{tab:hpgeneral}.
Whitening always improves the performance of the raw descriptor.
The unsupervised variant is superior when learning it on an independent dataset.
It generalizes better, implying over-fitting of the supervised one (recall the observations
of Figure~\ref{fig:sim1D}).
Learning on HP (the corresponding training part per split) with supervision significantly helps.
Note that PT contains only patches detected by DoG, while HP uses a combination of detectors.

\subsection{Comparison with the State of the Art.}

We compare the performance of the proposed descriptor with previously published results on Phototourism dataset.
Results are shown  in Table~\ref{tab:ptbestscores}.
Our method obtains the best performance among the unsupervised/hand-crafted approaches by a large margin.
Overall, it comes right after the two very recent CNN-based  descriptors, namely L2Net~\cite{TW17} and HardNet~\cite{MMR+17}.
The advantage of our approach is the low cost of the learning. It takes less than a minute; about 45 seconds to extract descriptors of Liberty and about 10 seconds to compute the projection matrix. CNN-based competitors require several hours or days of training.

The comparison on the HPatches dataset is reported in Figure~\ref{fig:hp_bars}.
We use the provided descriptors and framework to evaluate all approaches by ourselves.
For the descriptors that require learning, the model that is learned on Liberty-PT
is used. Our unsupervised descriptor is the top performing hand-crafted variant by a large margin.
Overall, it is always outperformed by HardNet, L2Net, while on verification is it additionally
outperformed by DDesc and TF-M. Verification is closer to the learning task (loss) involved in the learning of these
CNN-based methods.

Finally, we learn supervised whitening \lw for all other descriptors,
 post-process them, and present results in Figure~\ref{fig:hp_bars_proj}.
The projection matrix is learned on HP, in particular the training part of each split.
Supervised whitening \lw consistently boosts the performance of all descriptors, while
this comes at a minimal extra cost compared to the initial training
of a CNN descriptor.
Our descriptor comes 3rd at 2 out of 3 tasks.
Note that it uses the whitening learned on HP (similarly to
 all other descriptors of this comparison), but does not use the PT dataset at all.
All CNN-based descriptors train their parameters on Liberty-PT which is costly,
while the overall learning of our descriptor is again in the order of a single minute.

\input{tables/ptbestscores}
\input{tables/hpbest_bars}

%% file: figs/tex/whitening_scores.tex
\begin{figure*}[t]
\centering
\begin{tabular}{@{\sssp}c@{\msp}c@{\msp}c@{\msp}c@{\sssp}}
\input{figs/tex/pt_w_test.tex} &
\input{figs/tex/hp_w_test.tex} \\
\input{figs/tex/pt_ws_test.tex} &
\input{figs/tex/hp_ws_test.tex}\\
\end{tabular}
\vspace{10pt}
\caption{Impact of the shrinkage parameter for unsupervised whitening when trained on the same or different dataset. We evaluate performance on Photo Tourism and HPatches datasets versus shrinkage parameter $t$ for the attenuated whitening \pcawt (top row), and versus shrinkage parameter $\beta=\lambda_i$  for whitening with shrinkage \pcaws (bottom row). 
\label{fig:paramshrink}}
\end{figure*}
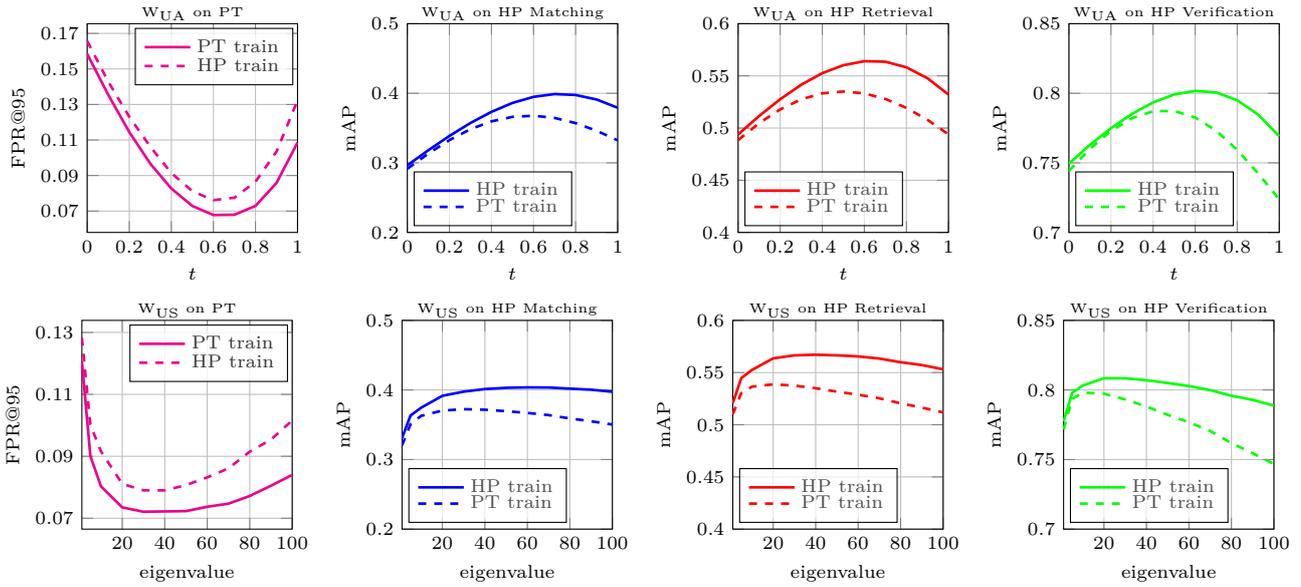

%% file: figs/tex/pt_w_test.tex
\begin{tikzpicture}

\begin{axis}[%
width=0.25\textwidth,
height=0.25\textwidth,
ylabel = {FPR@95},
xlabel = {$t$},
title = {\tiny \pcawt~on PT},
title style={at={(axis description cs:0.5,0.9)}},
xmin=0,
xmax=1,
ytick = {0.05, 0.07, ..., 0.21},
y tick label style={/pgf/number format/.cd,fixed,fixed zerofill,precision=2,/tikz/.cd},
grid=both,
legend style={font=\scriptsize, fill opacity=0.8, row sep=-3pt, legend cell align=left, align=left}
]
\addplot [color=magenta,style=solid,line width=1pt]
  table[row sep=crcr]{%
0	0.15848\\
0.1	0.135483333333333\\
0.2	0.114533333333333\\
0.3	0.0966866666666667\\
0.4	0.08269\\
0.5	0.0728666666666667\\
0.6	0.0677666666666667\\
0.7	0.0679466666666667\\
0.8	0.0729\\
0.9	0.0858266666666667\\
1	0.1086\\
};
\addlegendentry{PT train}

\addplot [color=magenta,style=dashed,line width=1pt]
  table[row sep=crcr]{%
0	0.165706666666667\\
0.1	0.14292\\
0.2	0.123113333333333\\
0.3	0.106266666666667\\
0.4	0.0914133333333333\\
0.5	0.0812066666666667\\
0.6	0.0761466666666667\\
0.7	0.0775333333333333\\
0.8	0.0865466666666667\\
0.9	0.103493333333333\\
1	0.1323\\
};
\addlegendentry{HP train}

\end{axis}
\end{tikzpicture}%

%% file: figs/tex/hp_w_test.tex
\begin{tikzpicture}
\begin{axis}[%
width=0.25\textwidth,
height=0.25\textwidth,
xmin=0,
xmax=1,
ymin=0.2,
ymax=0.5,
ylabel = {mAP},
xlabel = {$t$},
title = {\tiny \pcawt~on HP Matching},
title style={at={(axis description cs:0.5,0.9)}},
grid=both,
legend pos=south west,
legend style={font=\scriptsize, fill opacity=0.7, row sep=-4pt, legend cell align=left, align=left}
]
\addplot [color=blue,style=solid,line width=1pt]
  table[row sep=crcr]{%
0	0.29646363802565\\
0.1	0.318162397323536\\
0.2	0.338657261463504\\
0.3	0.357309059752853\\
0.4	0.3732810854779\\
0.5	0.386131787535922\\
0.6	0.394817391194786\\
0.7	0.398842086861292\\
0.8	0.397539515027142\\
0.9	0.391003572949066\\
1	0.379137284378302\\
};
\addlegendentry{HP train}

\addplot [color=blue,style=dashed,line width=1pt]
  table[row sep=crcr]{%
0	0.29147834437917\\
0.1	0.313356026648935\\
0.2	0.332665356148897\\
0.3	0.348400974871897\\
0.4	0.359595513353177\\
0.5	0.366119929865132\\
0.6	0.367702741388424\\
0.7	0.364589835098694\\
0.8	0.357095695033489\\
0.9	0.346091871777097\\
1	0.332300663437739\\
};
\addlegendentry{PT train}
\end{axis}%
\end{tikzpicture}%

&  

\begin{tikzpicture}
\begin{axis}[%
width=0.25\textwidth,
height=0.25\textwidth,
xmin=0,
xmax=1,
ymin=0.4,
ymax=0.6,
ylabel = {mAP},
xlabel = {$t$},
title = {\tiny \pcawt~on HP Retrieval},
title style={at={(axis description cs:0.5,0.9)}},
grid=both,
legend pos=south west,
legend style={font=\scriptsize, fill opacity=0.7, row sep=-4pt, legend cell align=left, align=left}
]
\addplot [color=red,style=solid,line width=1pt]
  table[row sep=crcr]{%
0	0.49386676399589\\
0.1	0.511595809160802\\
0.2	0.527610609951659\\
0.3	0.541494361508906\\
0.4	0.552593750907569\\
0.5	0.560178942749209\\
0.6	0.564057803901565\\
0.7	0.563567187820765\\
0.8	0.558205783724899\\
0.9	0.547747766652748\\
1	0.532097752058546\\
};
\addlegendentry{HP train}

\addplot [color=red,style=dashed,line width=1pt]
  table[row sep=crcr]{%
0	0.488435975492071\\
0.1	0.50474958697828\\
0.2	0.51785722032738\\
0.3	0.527541920023478\\
0.4	0.533270627000347\\
0.5	0.53498366872512\\
0.6	0.53325469340495\\
0.7	0.52784344763427\\
0.8	0.519338590377238\\
0.9	0.507893084395045\\
1	0.493974682655357\\
};
\addlegendentry{PT train}

\end{axis}%
\end{tikzpicture}%

&

\begin{tikzpicture}
\begin{axis}[%
width=0.25\textwidth,
height=0.25\textwidth,
xmin=0,
xmax=1,
ymin=0.7,
ymax=0.85,
ylabel = {mAP},
xlabel = {$t$},
title = {\tiny \pcawt~on HP Verification},
title style={at={(axis description cs:0.5,0.9)}},
grid=both,
legend pos=south west,
legend style={font=\scriptsize, fill opacity=0.7, row sep=-4pt, legend cell align=left, align=left}
]
\addplot [color=green,style=solid,line width=1pt]
  table[row sep=crcr]{%
0	  0.7495\\
0.1	0.7629\\
0.2	0.7749\\
0.3	0.7852\\
0.4	0.7934\\
0.5	0.7991\\
0.6	0.8017\\
0.7	0.8006\\
0.8	0.7950\\
0.9	0.7846\\
1	  0.7691\\
};
\addlegendentry{HP train}

\addplot [color=green,style=dashed,line width=1pt]
  table[row sep=crcr]{%
0	  0.7442\\
0.1	0.7601\\
0.2	0.7729\\
0.3	0.7821\\
0.4	0.7871\\
0.5	0.7873\\
0.6	0.7826\\
0.7	0.7731\\
0.8	0.7593\\
0.9	0.7423\\
1	  0.7237\\
};
\addlegendentry{PT train}

\end{axis}%
\end{tikzpicture}%

%% file: figs/tex/pt_ws_test.tex
\begin{tikzpicture}

\begin{axis}[%
width=0.25\textwidth,
height=0.25\textwidth,
ylabel = {FPR@95},
xlabel = {eigenvalue},
title = {\tiny \pcaws~on PT},
title style={at={(axis description cs:0.5,0.9)}},
xmin=1,
xmax=100,
ytick = {0.05, 0.07, ..., 0.21},
y tick label style={/pgf/number format/.cd,fixed,fixed zerofill,precision=2,/tikz/.cd},
grid=both,
legend style={font=\scriptsize, fill opacity=0.8, row sep=-3pt, legend cell align=left, align=left}
]
\addplot [color=magenta,style=solid,line width=1pt]
  table[row sep=crcr]{%
    1.0000    0.1207\\
    5.0000    0.0901\\
   10.0000    0.0803\\
   20.0000    0.0735\\
   30.0000    0.0721\\
   40.0000    0.0722\\
   50.0000    0.0723\\
   60.0000    0.0737\\
   70.0000    0.0747\\
   80.0000    0.0772\\
   90.0000    0.0805\\
  100.0000    0.0840\\};
\addlegendentry{PT train}

\addplot [color=magenta,style=dashed,line width=1pt]
  table[row sep=crcr]{%
    1.0000    0.1283\\
    5.0000    0.1002\\
   10.0000    0.0916\\
   20.0000    0.0811\\
   30.0000    0.0790\\
   40.0000    0.0790\\
   50.0000    0.0807\\
   60.0000    0.0832\\
   70.0000    0.0861\\
   80.0000    0.0915\\
   90.0000    0.0954\\
  100.0000    0.1016\\
};
\addlegendentry{HP train}

\end{axis}
\end{tikzpicture}%

%% file: figs/tex/hp_ws_test.tex
\begin{tikzpicture}
\begin{axis}[%
width=0.25\textwidth,
height=0.25\textwidth,
xmin=1,
xmax=100,
ymin=0.2,
ymax=0.5,
ylabel = {mAP},
xlabel = {eigenvalue},
title = {\tiny \pcaws~on HP Matching},
title style={at={(axis description cs:0.5,0.9)}},
grid=both,
legend pos=south west,
legend style={font=\scriptsize, fill opacity=0.7, row sep=-4pt, legend cell align=left, align=left}
]
\addplot [color=blue,style=solid,line width=1pt]
  table[row sep=crcr]{%
    1.0000    0.3310\\
    5.0000    0.3633\\
   10.0000    0.3745\\
   20.0000    0.3916\\
   30.0000    0.3976\\
   40.0000    0.4013\\
   50.0000    0.4031\\
   60.0000    0.4037\\
   70.0000    0.4035\\
   80.0000    0.4019\\
   90.0000    0.4004\\
  100.0000    0.3974\\
};
\addlegendentry{HP train}

\addplot [color=blue,style=dashed,line width=1pt]
  table[row sep=crcr]{%
  1.0000    0.3203\\
  5.0000    0.3508\\
 10.0000    0.3628\\
 20.0000    0.3704\\
 30.0000    0.3724\\
 40.0000    0.3716\\
 50.0000    0.3695\\
 60.0000    0.3670\\
 70.0000    0.3639\\
 80.0000    0.3591\\
 90.0000    0.3550\\
100.0000    0.3503\\
};
\addlegendentry{PT train}

\end{axis}
\end{tikzpicture}%

&

\begin{tikzpicture}
\begin{axis}[%
width=0.25\textwidth,
height=0.25\textwidth,
xmin=1,
xmax=100,
ymin=0.4,
ymax=0.6,
ylabel = {mAP},
xlabel = {eigenvalue},
title = {\tiny \pcaws~on HP Retrieval},
title style={at={(axis description cs:0.5,0.9)}},
grid=both,
legend pos=south west,
legend style={font=\scriptsize, fill opacity=0.7, row sep=-4pt, legend cell align=left, align=left}
]

\addplot [color=red,style=solid,line width=1pt]
  table[row sep=crcr]{%
   1.0000     0.5207\\
    5.0000    0.5447\\
   10.0000    0.5525\\
   20.0000    0.5636\\
   30.0000    0.5665\\
   40.0000    0.5671\\
   50.0000    0.5665\\
   60.0000    0.5655\\
   70.0000    0.5634\\
   80.0000    0.5599\\
   90.0000    0.5572\\
  100.0000    0.5531\\
};
\addlegendentry{HP train}

\addplot [color=red,style=dashed,line width=1pt]
  table[row sep=crcr]{%
   1.0000     0.5101\\
   5.0000     0.5303\\
   10.0000    0.5365\\
   20.0000    0.5387\\
   30.0000    0.5375\\
   40.0000    0.5350\\
   50.0000    0.5317\\
   60.0000    0.5288\\
   70.0000    0.5253\\
   80.0000    0.5205\\
   90.0000    0.5164\\
  100.0000    0.5118\\
};
\addlegendentry{PT train}

\end{axis}
\end{tikzpicture}%

&

\begin{tikzpicture}
\begin{axis}[%
width=0.25\textwidth,
height=0.25\textwidth,
xmin=1,
xmax=100,
ymin=0.7,
ymax=0.85,
ylabel = {mAP},
xlabel = {eigenvalue},
title = {\tiny \pcaws~on HP Verification},
title style={at={(axis description cs:0.5,0.9)}},
grid=both,
legend pos=south west,
legend style={font=\scriptsize, fill opacity=0.7, row sep=-4pt, legend cell align=left, align=left}
]
\addplot [color=green,style=solid,line width=1pt]
  table[row sep=crcr]{%
    1.0000    0.7775\\
    5.0000    0.7979\\
   10.0000    0.8032\\
   20.0000    0.8084\\
   30.0000    0.8084\\
   40.0000    0.8070\\
   50.0000    0.8050\\
   60.0000    0.8028\\
   70.0000    0.7999\\
   80.0000    0.7958\\
   90.0000    0.7929\\
  100.0000    0.7887\\
};
\addlegendentry{HP train}
        
\addplot [color=green,style=dashed,line width=1pt]
  table[row sep=crcr]{%
    1.0000    0.7716\\
    5.0000    0.7934\\
   10.0000    0.7981\\
   20.0000    0.7973\\
   30.0000    0.7932\\
   40.0000    0.7881\\
   50.0000    0.7821\\
   60.0000    0.7768\\
   70.0000    0.7705\\
   80.0000    0.7618\\
   90.0000    0.7545\\
  100.0000    0.7468\\
};
\addlegendentry{PT train}
\end{axis}
\end{tikzpicture}%

%% file: figs/tex/eigval.tex
\begin{figure}
\centering
\begin{tabular}{c}

\begin{tikzpicture}

\begin{axis}[%
width=0.45\textwidth,
height=0.3\textwidth,
ylabel = {$\lambda_i$},
xlabel = {$i$},
ymin=0,
ymax=1,
xmin = 1,
xmax = 120,
grid=both,
legend style={font=\scriptsize, fill opacity=0.8, row sep=1pt, legend cell align=left, align=left}
]
\addplot [color=blue,style=solid,line width=1pt]
  table[row sep=crcr]{%
1	1\\
2	0.709754699227506\\
3	0.357337314106219\\
4	0.270629115551343\\
5	0.220512378843122\\
6	0.163816953357251\\
7	0.136943674611685\\
8	0.128284303246623\\
9	0.118090822190159\\
10	0.102102510889261\\
11	0.0921471860495204\\
12	0.0850478205155922\\
13	0.0783569365422453\\
14	0.0579480040045282\\
15	0.0548883796878453\\
16	0.0537391405475421\\
17	0.0519084734728179\\
18	0.0504368800201321\\
19	0.0479304236833874\\
20	0.0449936775652901\\
21	0.0430435564334036\\
22	0.0397486227040373\\
23	0.0380837452062915\\
24	0.0353955415373363\\
25	0.0347822768261472\\
26	0.0329666990488744\\
27	0.0311676688799948\\
28	0.0266179197695989\\
29	0.0263590740709373\\
30	0.024626882951974\\
31	0.0224214376669482\\
32	0.0215966255781161\\
33	0.0207100855811783\\
34	0.0199864937681926\\
35	0.0195281173941568\\
36	0.0192243852169435\\
37	0.0175859279830685\\
38	0.0169592239960769\\
39	0.016354571233263\\
40	0.0153262660206233\\
41	0.0144732618660348\\
42	0.0141267765843994\\
43	0.0132878445924316\\
44	0.0127101934001257\\
45	0.0124649932113559\\
46	0.0116019030644735\\
47	0.0112283396675253\\
48	0.0108521411082902\\
49	0.0107266120549431\\
50	0.00998101420546585\\
51	0.00972046646367394\\
52	0.00936534898229191\\
53	0.00890122174732463\\
54	0.00853031898925371\\
55	0.00817449946607666\\
56	0.00786894715608809\\
57	0.00771749472430721\\
58	0.00762313405379903\\
59	0.00729353320810423\\
60	0.00718533701365901\\
61	0.00662425278736294\\
62	0.00633345156109058\\
63	0.00583890435822903\\
64	0.00577264203128653\\
65	0.00556268950498155\\
66	0.00547689574220041\\
67	0.00540477386094521\\
68	0.00519334246066692\\
69	0.00511724690308132\\
70	0.00499962155881158\\
71	0.00483349921684396\\
72	0.00471225593528092\\
73	0.00462309266963181\\
74	0.00435437222071419\\
75	0.00407763039095627\\
76	0.00396689815313579\\
77	0.00380950977020003\\
78	0.00367364227200932\\
79	0.0033495154347604\\
80	0.00308181918598368\\
81	0.00293786140408898\\
82	0.00276676684926289\\
83	0.00271809155674291\\
84	0.00262620776681419\\
85	0.00249224769235192\\
86	0.00233407853877714\\
87	0.00224283129935827\\
88	0.00221536884352948\\
89	0.00204846177816474\\
90	0.00201832245653023\\
91	0.00175569823127302\\
92	0.00169707013912564\\
93	0.00166182364746242\\
94	0.00164337476576358\\
95	0.00139564872220358\\
96	0.001371210645292\\
97	0.00127066491852721\\
98	0.00125083779610917\\
99	0.0012379244192853\\
100	0.0012162214513051\\
101	0.00112702180510842\\
102	0.00111156885220148\\
103	0.00105278968169136\\
104	0.00100043913795096\\
105	0.00094066420407423\\
106	0.000939594187341017\\
107	0.000912432704640536\\
108	0.000852729018195186\\
109	0.000797786913544271\\
110	0.000735974888368824\\
111	0.000698909743916139\\
112	0.000607556373445585\\
113	0.00058360557934877\\
114	0.000564799508644586\\
115	0.00055970841224458\\
116	0.00053635578907573\\
117	0.000520317759536547\\
118	0.000456228653082616\\
119	0.000448477008118862\\
120	0.000436926709262459\\
121	0.000406603484443955\\
122	0.000404022936996326\\
123	0.000371771721748194\\
124	0.000339964919592303\\
125	0.000332325140845213\\
126	0.000306705319569569\\
127	0.000303089877493625\\
128	0.000270261483500187\\
129	0.000245894886089116\\
130	0.000238750403017643\\
131	0.00022375395726397\\
132	0.000207929369297207\\
133	0.000198785448333422\\
134	0.000187985384511533\\
135	0.000181800053665706\\
136	0.000169637842767637\\
137	0.000165997872374456\\
138	0.000157350293788441\\
139	0.000155288730802707\\
140	0.000147067376370138\\
141	0.000142860499222927\\
142	0.000131225127961261\\
143	0.000116816988027596\\
144	0.000106484467243005\\
145	0.000106099240268542\\
146	0.000100881168866464\\
147	9.5043895657927e-05\\
148	9.28432948438206e-05\\
149	8.65280984098289e-05\\
150	8.47618560377757e-05\\
151	8.03617711490817e-05\\
152	7.81363590158364e-05\\
153	7.38762390148352e-05\\
154	7.12091770197657e-05\\
155	6.87915431973124e-05\\
156	6.72537303592643e-05\\
157	5.9480264600959e-05\\
158	5.82416828817431e-05\\
159	5.75685630242634e-05\\
160	5.63056468464044e-05\\
161	5.45354843851079e-05\\
162	4.62735971408559e-05\\
163	4.29733283442453e-05\\
164	4.24240808341664e-05\\
165	4.01807621500121e-05\\
166	3.97966904572185e-05\\
167	3.76317552421803e-05\\
168	3.638940078373e-05\\
169	3.51368887212928e-05\\
170	2.89377235372202e-05\\
171	2.8218214804502e-05\\
172	2.70880148860159e-05\\
173	2.58138819295928e-05\\
174	2.57559595587173e-05\\
175	2.52439446661751e-05\\
176	2.20200278139389e-05\\
177	2.09773390784496e-05\\
178	2.05499938701463e-05\\
179	1.69660974274311e-05\\
180	1.65185444535051e-05\\
181	1.58553239808916e-05\\
182	1.45483537295709e-05\\
183	1.38894106330639e-05\\
184	1.32931094678691e-05\\
185	1.19753374298662e-05\\
186	1.18119383339922e-05\\
187	1.15587980815388e-05\\
188	1.02175203667383e-05\\
189	9.37233931428891e-06\\
190	9.16458407361488e-06\\
191	9.13999182238038e-06\\
192	8.36540147809453e-06\\
193	7.499808724907e-06\\
194	6.28559095156651e-06\\
195	6.26074835848811e-06\\
196	6.0099930941167e-06\\
197	5.59673358155202e-06\\
198	5.28460199005361e-06\\
199	4.717200121954e-06\\
200	4.44726083304575e-06\\
201	4.27095918057021e-06\\
202	4.13831963516171e-06\\
203	4.07967863500669e-06\\
204	3.71673950212057e-06\\
205	3.50345002751268e-06\\
206	3.26644700940814e-06\\
207	2.73679924917443e-06\\
208	2.65670475565162e-06\\
209	2.57507859550117e-06\\
210	2.51578553988846e-06\\
211	2.40865936210039e-06\\
212	2.38153521200469e-06\\
213	2.3589168082848e-06\\
214	2.15012666653514e-06\\
215	2.07983799918302e-06\\
216	2.07226039681809e-06\\
217	1.96264395003734e-06\\
218	1.81858715043821e-06\\
219	1.71992291793962e-06\\
220	1.43608628221765e-06\\
221	1.40999988868793e-06\\
222	1.33842809845569e-06\\
223	1.07428074956597e-06\\
224	1.06755398704792e-06\\
225	9.63380179216756e-07\\
226	9.18784343024027e-07\\
227	8.84427367790031e-07\\
228	8.69238914059276e-07\\
229	7.48785809296011e-07\\
230	6.61262632471129e-07\\
231	6.16680060813332e-07\\
232	5.7301096061453e-07\\
233	4.80509247371944e-07\\
234	4.67401559267711e-07\\
235	4.38587452127982e-07\\
236	3.84382029943411e-07\\
237	2.78527844193783e-07\\
238	1.64009095380869e-07\\
};
\addlegendentry{\pcaw,~$\lambda$}

\addplot [color=cyan,style=solid,line width=1pt]
  table[row sep=crcr]{%
1	1\\
2	0.786639578854834\\
3	0.486580676783091\\
4	0.400555005059059\\
5	0.347058816966357\\
6	0.281871516025837\\
7	0.248644801648514\\
8	0.237531638108567\\
9	0.22415648221753\\
10	0.202453596220128\\
11	0.188424535886926\\
12	0.178141141971397\\
13	0.16821093074983\\
14	0.13618426681629\\
15	0.131110141886\\
16	0.129182445436708\\
17	0.126085974711854\\
18	0.123573045335824\\
19	0.11924162546898\\
20	0.114079088957782\\
21	0.110595053518494\\
22	0.104598500689547\\
23	0.101512086779076\\
24	0.0964415080668454\\
25	0.0952687802933829\\
26	0.0917598773569169\\
27	0.0882252758547857\\
28	0.0789990517205726\\
29	0.0784605061330356\\
30	0.0748146437686271\\
31	0.0700590782862947\\
32	0.0682448886148279\\
33	0.066271580096161\\
34	0.0646421256029553\\
35	0.0636007539574223\\
36	0.0629066737663701\\
37	0.0591038407002587\\
38	0.0576214500887169\\
39	0.056175560444047\\
40	0.0536791279185968\\
41	0.0515699243039817\\
42	0.0507025910696579\\
43	0.0485756005009154\\
44	0.0470875951968417\\
45	0.0464498624657502\\
46	0.0441743800014184\\
47	0.0431738600470011\\
48	0.0421561365436869\\
49	0.0418142009243088\\
50	0.0397577933086447\\
51	0.0390284203976927\\
52	0.0380247857433853\\
53	0.0366956626188095\\
54	0.0356185052633145\\
55	0.03457186582863\\
56	0.0336621336668055\\
57	0.0332072903060918\\
58	0.0329225518166166\\
59	0.0319195364184598\\
60	0.0315873359824027\\
61	0.0298397945059913\\
62	0.0289166725668552\\
63	0.0273169339811098\\
64	0.0270995599683844\\
65	0.0264058026764618\\
66	0.0261200576814841\\
67	0.025878808046641\\
68	0.0251659235715454\\
69	0.0249072316702156\\
70	0.0245050725414128\\
71	0.0239322293843875\\
72	0.0235104103353763\\
73	0.0231981215286174\\
74	0.02224579267924\\
75	0.0212464071739641\\
76	0.0208408647567665\\
77	0.0202585503072293\\
78	0.0197500320458453\\
79	0.0185134513156327\\
80	0.0174648559603157\\
81	0.0168896991078738\\
82	0.0161949941100482\\
83	0.0159950225632252\\
84	0.01561458179328\\
85	0.0150526804358749\\
86	0.0143774155655151\\
87	0.0139816242256882\\
88	0.0138615638735368\\
89	0.013121984048097\\
90	0.0129865378799924\\
91	0.0117791759793258\\
92	0.0115024367628615\\
93	0.0113346849736562\\
94	0.011246454481132\\
95	0.0100309781776304\\
96	0.00990770172107441\\
97	0.00939337645674475\\
98	0.00929053442683979\\
99	0.0092232905720519\\
100	0.0091098003974319\\
101	0.00863679523082968\\
102	0.00855372843493637\\
103	0.00823453602489296\\
104	0.00794572392414816\\
105	0.00761034302548744\\
106	0.00760428219194457\\
107	0.00744973098605925\\
108	0.00710505966906892\\
109	0.00678142301208414\\
110	0.00640920403531983\\
111	0.0061815127120486\\
112	0.00560416104236395\\
113	0.00544858342127985\\
114	0.00532507853454299\\
115	0.00529143278698165\\
116	0.00513590616888813\\
117	0.00502791647738087\\
118	0.0045859338941946\\
119	0.00453125108818592\\
120	0.00444924187180397\\
121	0.004230773420262\\
122	0.00421195977517278\\
123	0.00397368509112249\\
124	0.00373253486995449\\
125	0.00367362006306842\\
126	0.0034730010610021\\
127	0.00344429231392495\\
128	0.00317869537072896\\
129	0.00297525811034666\\
130	0.00291447865742103\\
131	0.00278509175387688\\
132	0.00264570297316096\\
133	0.00256371186147843\\
134	0.00246539697930855\\
135	0.00240832894277883\\
136	0.00229438301365596\\
137	0.00225980916235796\\
138	0.00217674367243153\\
139	0.00215674082059247\\
140	0.0020761629728151\\
141	0.00203441015214781\\
142	0.00191695493555995\\
143	0.00176707182780354\\
144	0.00165615272562179\\
145	0.00165195644205647\\
146	0.00159465622997549\\
147	0.00152949112113609\\
148	0.00150461497903931\\
149	0.00143222091546404\\
150	0.00141169319281065\\
151	0.00135998666469387\\
152	0.00133351289339444\\
153	0.00128219283122465\\
154	0.00124961196983997\\
155	0.00121976041891927\\
156	0.00120060864129003\\
157	0.00110169410167912\\
158	0.00108558476723074\\
159	0.00107678692235976\\
160	0.00106019650068288\\
161	0.00103675329051381\\
162	0.000924130455236987\\
163	0.000877484177568561\\
164	0.000869618377656876\\
165	0.000837168192142183\\
166	0.000831558620238969\\
167	0.000799628257207636\\
168	0.000781056467587664\\
169	0.000762139221907387\\
170	0.000665311536596899\\
171	0.000653688243905679\\
172	0.000635249051980557\\
173	0.000614182338282532\\
174	0.000613217321946061\\
175	0.000604658348696049\\
176	0.000549506082798859\\
177	0.000531159897897768\\
178	0.000523562083941796\\
179	0.000457833291779314\\
180	0.000449345341403177\\
181	0.000436639061080244\\
182	0.000411121033451225\\
183	0.000397995968078886\\
184	0.000385956719868781\\
185	0.000358757865625093\\
186	0.000355324226826153\\
187	0.000349976492291633\\
188	0.00032102722038138\\
189	0.000302199403192036\\
190	0.000297494493670994\\
191	0.00029693546063235\\
192	0.000279087735678033\\
193	0.000258544397374499\\
194	0.000228476890995133\\
195	0.000227844407835638\\
196	0.000221417417503853\\
197	0.000210646498173308\\
198	0.000202352511393299\\
199	0.000186886964794202\\
200	0.000179334902571093\\
201	0.000174328259630579\\
202	0.000170520588234325\\
203	0.000168825547921966\\
204	0.000158166129253275\\
205	0.000151756446635573\\
206	0.000144495037485237\\
207	0.000127664473769538\\
208	0.000125037506738693\\
209	0.00012233573687971\\
210	0.000120357049207171\\
211	0.000116746204384293\\
212	0.000115824358891878\\
213	0.000115053235803926\\
214	0.000107826332014181\\
215	0.000105346632131208\\
216	0.000105077813940582\\
217	0.000101155398175718\\
218	9.58989567628439e-05\\
219	9.22266191324863e-05\\
220	8.12879878846138e-05\\
221	8.0251536255774e-05\\
222	7.73778274547581e-05\\
223	6.63410674926365e-05\\
224	6.60500107762659e-05\\
225	6.14693180140122e-05\\
226	5.94633688329636e-05\\
227	5.78979945525093e-05\\
228	5.72001822316089e-05\\
229	5.15288349903549e-05\\
230	4.7234768063587e-05\\
231	4.4982328681848e-05\\
232	4.27281461542394e-05\\
233	3.77738477070018e-05\\
234	3.70495647043688e-05\\
235	3.54355633677627e-05\\
236	3.23097913310436e-05\\
237	2.57874282378589e-05\\
238	1.77994897694553e-05\\
};
\addlegendentry{\pcawt,~$\lambda^t$}

\addplot [color=orange,style=solid,line width=1pt]
  table[row sep=crcr]{%
1	1\\
2	0.714285496976427\\
3	0.367369430536057\\
4	0.282014767455416\\
5	0.232680364853059\\
6	0.176869968359842\\
7	0.1504161878474\\
8	0.14189199131612\\
9	0.131857632908331\\
10	0.116118902925919\\
11	0.106318983060018\\
12	0.0993304403001169\\
13	0.0927440026070854\\
14	0.0726536583302417\\
15	0.0696417954723307\\
16	0.0685104962266476\\
17	0.0667084062976658\\
18	0.0652597847666097\\
19	0.0627924548057204\\
20	0.0599015519886709\\
21	0.0579818727085937\\
22	0.0547383736735502\\
23	0.0530994853070633\\
24	0.050453245132778\\
25	0.0498495536286849\\
26	0.0480623174496546\\
27	0.0462913705669392\\
28	0.0418126441155585\\
29	0.041557839059435\\
30	0.0398526878532683\\
31	0.0376816700904321\\
32	0.0368697335132634\\
33	0.035997032615225\\
34	0.0352847362414719\\
35	0.0348335152309789\\
36	0.0345345243848436\\
37	0.0329216438556636\\
38	0.0323047228660398\\
39	0.0317095088757723\\
40	0.0306972557505573\\
41	0.0298575671924424\\
42	0.0295164906279504\\
43	0.0286906545626715\\
44	0.0281220206420141\\
45	0.0278806480861477\\
46	0.0270310309804625\\
47	0.026663298996438\\
48	0.0262929729856328\\
49	0.0261694034704792\\
50	0.0254354445796656\\
51	0.0251789640497074\\
52	0.0248293900361471\\
53	0.0243725079370212\\
54	0.0240073950586722\\
55	0.0236571299623553\\
56	0.0233563473961475\\
57	0.0232072591726248\\
58	0.0231143714944869\\
59	0.0227899157958995\\
60	0.0226834085696234\\
61	0.0221310830007205\\
62	0.0218448212503255\\
63	0.0213579940461944\\
64	0.0212927660898375\\
65	0.0210860909720936\\
66	0.0210016364702282\\
67	0.0209306404285864\\
68	0.0207225095224415\\
69	0.0206476018344962\\
70	0.020531812649653\\
71	0.020368283516709\\
72	0.0202489328714289\\
73	0.0201611614654316\\
74	0.0198966358062377\\
75	0.0196242139818544\\
76	0.0195152103004396\\
77	0.0193602787873626\\
78	0.019226532212933\\
79	0.0189074650722258\\
80	0.0186439476251665\\
81	0.0185022370582738\\
82	0.0183338133298893\\
83	0.0182858978701907\\
84	0.0181954484079409\\
85	0.0180635794819045\\
86	0.0179078793862027\\
87	0.0178180565377588\\
88	0.0177910227773566\\
89	0.0176267211707333\\
90	0.0175970523310333\\
91	0.0173385277320076\\
92	0.0172808148382363\\
93	0.017246118552639\\
94	0.0172279576623441\\
95	0.0169840986808779\\
96	0.0169600420881232\\
97	0.0168610659039187\\
98	0.0168415482875681\\
99	0.0168288364916087\\
100	0.0168074723120885\\
101	0.0167196650934526\\
102	0.0167044533647931\\
103	0.016646591751028\\
104	0.0165950584116492\\
105	0.016536216578619\\
106	0.0165351632650999\\
107	0.0165084257795644\\
108	0.0164496540817763\\
109	0.0163955696363654\\
110	0.0163347225115132\\
111	0.0162982359627263\\
112	0.0162083086399615\\
113	0.0161847317234217\\
114	0.0161662192199263\\
115	0.0161612075967445\\
116	0.016138219513543\\
117	0.0161224318414358\\
118	0.016059343181082\\
119	0.0160517125411284\\
120	0.0160403425451672\\
121	0.0160104926730555\\
122	0.0160079524085634\\
123	0.0159762046424084\\
124	0.0159448943519522\\
125	0.0159373738319571\\
126	0.0159121539421472\\
127	0.0159085949379766\\
128	0.0158762790029729\\
129	0.0158522927739069\\
130	0.0158452598179048\\
131	0.015830497470214\\
132	0.0158149199078058\\
133	0.0158059187256096\\
134	0.0157952872533367\\
135	0.0157891984769674\\
136	0.0157772261210561\\
137	0.0157736429714611\\
138	0.0157651303836208\\
139	0.0157631010021206\\
140	0.015755007984974\\
141	0.0157508667781734\\
142	0.0157394130378073\\
143	0.0157252298123436\\
144	0.0157150585846507\\
145	0.0157146793711604\\
146	0.0157095427550862\\
147	0.0157037966030922\\
148	0.015701630354177\\
149	0.015695413739452\\
150	0.0156936750685408\\
151	0.0156893436700173\\
152	0.0156871529970935\\
153	0.0156829593785726\\
154	0.0156803339500457\\
155	0.0156779540560586\\
156	0.0156764402488425\\
157	0.015668788128722\\
158	0.0156675688815562\\
159	0.0156669062692588\\
160	0.015665663067501\\
161	0.0156639205376941\\
162	0.0156557876204626\\
163	0.015652538869642\\
164	0.0156519981960153\\
165	0.0156497898960661\\
166	0.0156494118198232\\
167	0.0156472806797586\\
168	0.0156460577187469\\
169	0.0156448247586938\\
170	0.0156387223639435\\
171	0.0156380140869122\\
172	0.0156369015296815\\
173	0.0156356472862417\\
174	0.0156355902680528\\
175	0.0156350862458337\\
176	0.0156319126550856\\
177	0.0156308862429677\\
178	0.0156304655687191\\
179	0.015626937617747\\
180	0.0156264970511807\\
181	0.0156258441837365\\
182	0.0156245576155997\\
183	0.0156239089587607\\
184	0.0156233219659976\\
185	0.0156220247646926\\
186	0.0156218639162952\\
187	0.0156216147276219\\
188	0.0156202943875693\\
189	0.0156194623999929\\
190	0.0156192578878606\\
191	0.0156192336795002\\
192	0.0156184711806943\\
193	0.0156176191000486\\
194	0.0156164238365017\\
195	0.0156163993817074\\
196	0.0156161525407919\\
197	0.015615745732358\\
198	0.0156154384732145\\
199	0.0156148799286236\\
200	0.0156146142031508\\
201	0.0156144406536088\\
202	0.0156143100845981\\
203	0.0156142523589978\\
204	0.0156138950854305\\
205	0.0156136851254551\\
206	0.0156134518221101\\
207	0.0156129304422766\\
208	0.015612851598077\\
209	0.0156127712461205\\
210	0.0156127128786435\\
211	0.0156126074247307\\
212	0.015612580723995\\
213	0.0156125584586699\\
214	0.0156123529277916\\
215	0.015612283736347\\
216	0.0156122762770328\\
217	0.0156121683717246\\
218	0.0156120265636857\\
219	0.0156119294396252\\
220	0.0156116500337465\\
221	0.0156116243545677\\
222	0.01561155390003\\
223	0.0156112938760837\\
224	0.0156112872543275\\
225	0.0156111847066975\\
226	0.0156111408070128\\
227	0.0156111069863581\\
228	0.0156110920349997\\
229	0.0156109734621964\\
230	0.015610887305277\\
231	0.0156108434186499\\
232	0.0156108004312346\\
233	0.015610709373495\\
234	0.015610696470421\\
235	0.0156106681061089\\
236	0.0156106147468462\\
237	0.0156105105450693\\
238	0.0156103978139852\\
};
\addlegendentry{\pcaws,~$(1-\beta)\lambda +\beta$}

\end{axis}
\end{tikzpicture}%

\end{tabular}
\vspace{-3pt}
\caption{Eigenvalues for standard PCA whitening, attenuated whitening ($t=0.7$) and whitening with shrinkage ($\beta=\lambda_{40}$). We normalize so that the maximum eigenvalue is 1. First 120 eigenvalues (out of 238) are shown.
\label{fig:eval}}
\end{figure}
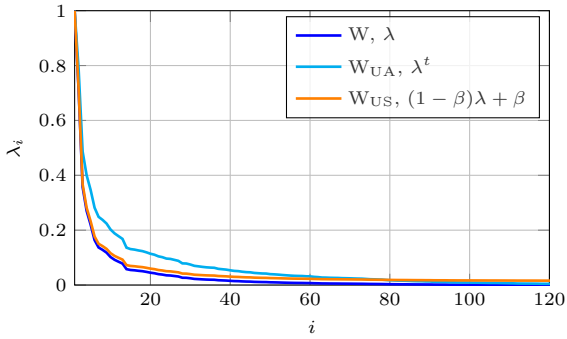

%% file: tables/ptscores.tex
\begin{table*}[t]
\small
\centering
\setlength\extrarowheight{0pt}
\begin{tabular}{LRRRRRRRR}
\toprule
Test & & & \multicolumn{2}{c}{Liberty} & \multicolumn{2}{c}{Notredame} & \multicolumn{2}{c}{Yosemite} \\ \cmidrule(lr){4-5}\cmidrule(lr){6-7}\cmidrule(lr){8-9}
Train & D & Mean & No& Yo& Li & Yo & Li & No  \\\midrule
\baseline \text{~\cite{BTJ15}}          & 175 &      22.42   &      24.34   &      24.34   &      16.06   &      16.06   &      26.85   &      26.85   \\
\xya                                    & 63  &      35.87   &      34.06   &      34.06   &      34.10   &      34.10   &      39.47   &      39.47   \\
\baseline + \xya                        & 238 &      25.37   &      26.16   &      26.16   &      20.04   &      20.04   &      29.91   &      29.91   \\
\baseline+PCA+SQRT \text{~\cite{BTJ15}}      & 128 &       8.30   &      12.09   &      13.13   &       5.16   &       5.41   &       7.52   &       6.49   \\
\baseline \text{~\cite{BTJ15}}+\lw      & 128 &       7.06   &       8.55   &      10.48   &       4.40   &       3.94   &       8.86   &       6.12   \\
\xya + \lw                              & 63  &      15.13   &      17.31   &      20.34   &      10.90   &      11.85   &      16.84   &      13.55   \\
\baseline + \xya + \lw                  & 128 &   \bf{5.94}  &   \bf{7.46}  &   \bf{9.85}  &   \bf{3.45}  &   \bf{3.55}  &       6.47   &   \bf{4.89}  \\
\baseline + \xya + \pcawt						    & 128 &       6.79   &      10.59   &      11.17   &       3.80   &       4.36   &   \bf{5.58}  &       5.16  \\
\baseline + \xya + \pcaws							  & 128 &       7.22   &      10.61   &      11.14   &       4.27   &       4.46   &       6.75   &       6.09   \\
\midrule[\heavyrulewidth]
\bottomrule
\end{tabular}
\vspace{1.5em}
\caption{Performance comparison on Phototourism dataset between the baseline approach and our combined descriptor.
We further show the benefit of learned whitening (\lw) over the standard PCA followed by square-rooting,
  as well as the other variants that do additional regularization (\pcawt, \pcaws) without supervision.
  FPR95 is reported for all methods.}\label{tab:ptscores}
\end{table*}

%% file: figs/response/hist.tex
\begin{figure}
\centering
\begin{tabular}{cc}
\begin{tikzpicture}
\begin{axis}[
      height=4.5cm,
      width=4.5cm,
      title={polar},
      xlabel={$\V_p^\top \V_q$},
      xmin = 0.1, xmax = 1.1,
      legend pos=north west,
      legend cell align=left,
      legend style={font=\tiny, opacity = 0.7},
      ytick=\empty,
      ybar,
      ]

\addplot[ybar,fill=blue!50, draw=none,bar width=0.2cm, opacity=0.5]
  table [x, y, col sep=comma] {figs/response/data/histn_aa2r_xya1_polar_notredame.txt};\leg{Negatives}
\addplot[ybar,fill=red!50, draw=none,bar width=0.2cm, opacity=0.5]
  table [x, y, col sep=comma] {figs/response/data/histp_aa2r_xya1_polar_notredame.txt};\leg{Positives}
\end{axis}
\end{tikzpicture}
&
\begin{tikzpicture}
\begin{axis}[ height=4.5cm,
      width=4.5cm,
      title={cartes},
      xlabel={$\V_p^\top \V_q$},
      xmin = 0.1, xmax = 1.1,
      legend pos=north west,
      legend cell align=left,
      legend style={font=\tiny, opacity = 0.7},
      ytick=\empty,
            ybar,
]
\addplot[ybar,fill=blue!50,draw=none,bar width=0.2cm, opacity=0.5]
  table [x, y, col sep=comma,] {figs/response/data/histn_aa2r_xya1_cart_notredame.txt};\leg{Negatives}
\addplot[ybar,fill=red!50,draw=none,bar width=0.2cm, opacity=0.5]
  table [x, y, col sep=comma] {figs/response/data/histp_aa2r_xya1_cart_notredame.txt};\leg{Positives}
\end{axis}
\end{tikzpicture}
\\
\begin{tikzpicture}
\begin{axis}[ height=4.5cm,
      width=4.5cm,
      title={polar+cartes},
      xlabel={$\V_p^\top \V_q$},
      xmin = 0.1, xmax = 1.1,
      legend pos=north west,
      legend cell align=left,
      legend style={font=\tiny, opacity = 0.7},
      ytick=\empty,
            ybar,
]
\addplot[ybar,fill=blue!50,draw=none,bar width=0.2cm, opacity=0.5]
  table [x, y, col sep=comma,] {figs/response/data/histn_aa2r_xya1_concat_notredame.txt};\leg{Negatives}
\addplot[ybar,fill=red!50,draw=none,bar width=0.2cm, opacity=0.5]
  table [x, y, col sep=comma] {figs/response/data/histp_aa2r_xya1_concat_notredame.txt};\leg{Positives}
\end{axis}
\end{tikzpicture}
&
\begin{tikzpicture}
\begin{axis}[ height=4.5cm,
      width=4.5cm,
      title={polar+cartes+\lw},
      xlabel={$\V_p^\top \V_q$},
      xmin = -0.5, xmax = 1.1,
      legend pos=north west,
      legend cell align=left,
      legend style={font=\tiny, opacity = 0.7},
      ytick=\empty,
            ybar,
]
\addplot[ybar,fill=blue!50,draw=none,bar width=0.2cm, opacity=0.5]
  table [x, y, col sep=comma,] {figs/response/data/histn_aa2r_xya1_lda_notredame.txt};\leg{Negatives}
\addplot[ybar,fill=red!50,draw=none,bar width=0.2cm, opacity=0.5]
  table [x, y, col sep=comma] {figs/response/data/histp_aa2r_xya1_lda_notredame.txt};\leg{Positives}
\end{axis}
\end{tikzpicture}
\end{tabular}
\caption{Histograms of patch similairity for positive and negative patch pairs.
Histograms are constructed from 50K matching and 50K non-matching pairs from Notredame dataset.
\label{fig:histograms}}

\end{figure}
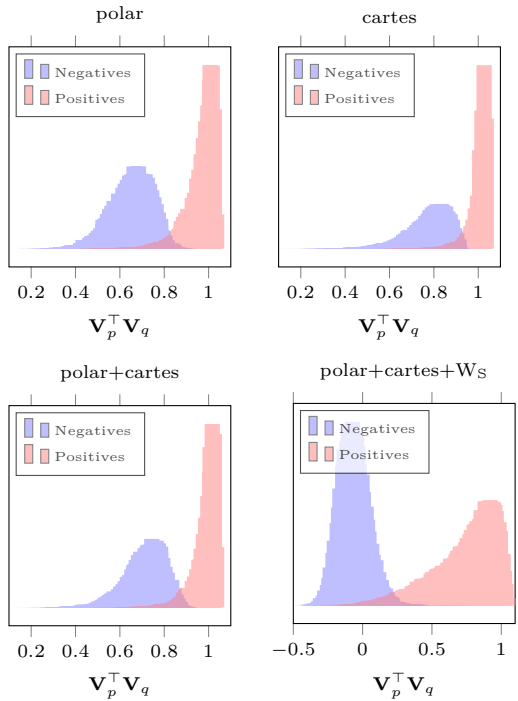

%% file: tables/hp_generalization.tex
\begin{table}
\footnotesize
\centering
\setlength\extrarowheight{0pt}
\def\dd{\hspace{-2pt}-\hspace{-2pt}}
\footnotesize
   \begin{tabular}{@{\sssp}L@{\msp}L@{\msp}L@{\msp}C@{\msp}C@{\msp}C@{\sssp}}
      \toprule
      Name                         &  Train      & Sup.        & R           &                     M       &       V      \\
      \midrule
      \baseline + \xya             & N/A         & N/A         &  45.23      &      29.68          &       77.78             \\
      \midrule
      % \baseline + \xya + \pcawt  & PT          & No          &  53.33      &      36.77          &       84.85             \\
      \baseline + \xya + \pcawt    & PT          & No          &  52.78      &      36.46          &       77.31             \\
      % \baseline + \xya + \pcaws  & PT          & No          &  53.17      &      36.95          &       84.77             \\
      \baseline + \xya + \pcaws    & PT          & No          &  \textbf{53.50} &  \textbf{37.16} &       \textbf{78.81}    \\
      \baseline + \xya + \lw       & PT          & Yes         &  49.66      &      32.58          &       75.82    \\
      \midrule
      % \baseline + \xya + \pcawt  & HP          & No          &  56.41      &      39.48          &       86.16             \\ % t = 0.6
      \baseline + \xya + \pcawt    & HP          & No          &  56.36      &      39.88          &       80.06            \\  % t= 0.7
      % \baseline + \xya + \pcaws  & HP          & No          &  56.65      &      40.31          &       86.32             \\ % ib = 50-th eigenvalue
      \baseline + \xya + \pcaws    & HP          & No          &  56.71      &      40.13          &       80.70             \\ % ib = 40-th eigenvalue
      \baseline + \xya + \lw       & HP          & Yes         &  \textbf{61.79} &  \textbf{44.40} &       \textbf{83.50}    \\
      \midrule[\heavyrulewidth]
     \bottomrule
   \end{tabular}
   \vspace{1.5em}
   \caption{Generalization of different whitening approaches. Mean Average Precision(mAP) for 3 tasks of HP, namely Retrieval (R), Matching (M), and Verification (V).
   The whitening is learned on PT or HP. We denote supervised by \emph{Sup.}\label{tab:hpgeneral}}
\end{table}

%% file: figs/response/syn_rot.tex
\begin{figure}
\begin{tikzpicture}
 % \tikzstyle{every node}=[font=\normalsize]
   \begin{axis}[%
      title style={yshift=-1.3ex,},
      title={Synthetic patch rotation},
      height=0.25\textwidth,
      width=0.45\textwidth,
      ylabel={FPR@95},
      xlabel={Rotation (degrees)},      
      legend style={font=\scriptsize, opacity = 0.7,  legend cell align=left, legend pos=north west},
      xmin = 0, xmax = 30,
      ymin = 0, ymax = 25,
   ]
   \addplot[color=green, style=solid, line width=2pt] table[x index=0,y index=1]{figs/response/data/syn_rot.txt};
   \addlegendentry{PC+\lw}
   \addplot[color=blue, style=solid, line width=2pt] table[x index=0,y index=2]{figs/response/data/syn_rot.txt};
   \addlegendentry{P+\lw}
   \addplot[color=red, style=solid, line width=2pt] table[x index=0,y index=3]{figs/response/data/syn_rot.txt};
   \addlegendentry{C+\lw}
   \end{axis}
\end{tikzpicture}
\\
\begin{tikzpicture}
 % \tikzstyle{every node}=[font=\normalsize]
   \begin{axis}[%
      title style={yshift=-1.3ex,},
      title={Synthetic patch translation},
      height=0.25\textwidth,
      width=0.45\textwidth,
      ylabel={FPR@95},
      xlabel={Translation (pixels)},      
      legend style={font=\scriptsize, opacity = 0.7,  legend cell align=left, legend pos=north west},
      xmin = 0, xmax = 12,
      ymin = 0, ymax = 25,
   ]
   \addplot[color=green, style=solid, line width=2pt] table[x index=0,y index=1]{figs/response/data/syn_trans.txt};
   \addlegendentry{PC+\lw}
   \addplot[color=blue, style=solid, line width=2pt] table[x index=0,y index=2]{figs/response/data/syn_trans.txt};
   \addlegendentry{P+\lw}
   \addplot[color=red, style=solid, line width=2pt] table[x index=0,y index=3]{figs/response/data/syn_trans.txt};
   \addlegendentry{C+\lw}
   \end{axis}
\end{tikzpicture}
\caption{Performance on PT (training on Liberty , testing on Notredame) when one patch of each pair undergoes synthetic rotation or translation.\label{fig:rottra}}
\end{figure}
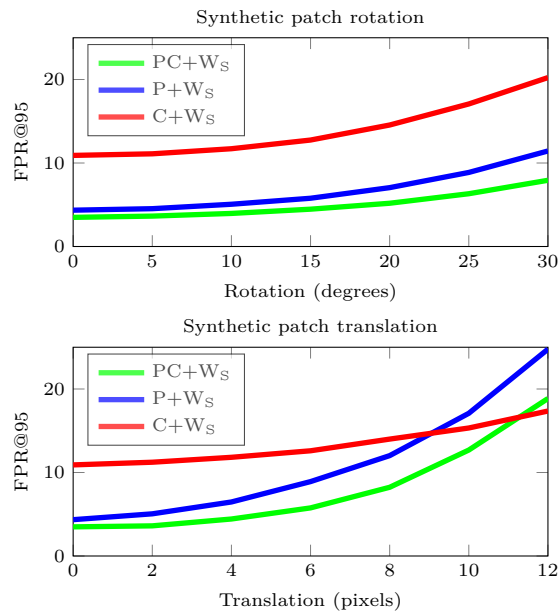

%% file: tables/ptbestscores.tex
\begin{table*}
\small
\centering
\def\dd{\hspace{-2pt}-\hspace{-2pt}}

\begin{tabular}{c@{\hspace{50pt}}c}

\begin{tabular}{LRR}
\multicolumn{3}{c}{Supervised}\\
\toprule
  Name & D  & FPR@95 \\
\midrule[\heavyrulewidth]
  Brown~\etal \text{~\cite{BHW11}} & 29\dd36  & 15.36 \\
  Trzcinski~\etal \text{~\cite{TCL+12}}    & 128    & 17.08 \\                
  Simonyan~\etal \text{~\cite{SVZ13}}         & 73\dd77  & 10.38 \\
  DC\dd S2S \text{~\cite{ZK15}}    &  512   & 9.67 \\
  DDESC \text{~\cite{STFK+15}}     &  128   & 9.85 \\
  Matchnet \text{~\cite{HLJS+15}}  & 4096   & 7.75 \\
  TF\dd M \text{~\cite{BRPM16}}    &  128   & 6.47 \\
  L2Net\hspace{-2pt}+ \text{~\cite{TW17}}       &  128   & 2.22 \\
  HardNet\hspace{-2pt}+ \text{~\cite{MMR+17}}   &  128   & {\bf 1.51} \\
  \baseline + \xya + \lw ~(our)          &  128   & 5.98 \\

\midrule[\heavyrulewidth]
\bottomrule
\end{tabular}

&

\begin{tabular}{LRR}
\multicolumn{3}{c}{Unsupervised}\\
\toprule
  Name & D   & FPR@95 \\
\midrule[\heavyrulewidth]
  RootSIFT                                   &  128 &  26.14 \\
  RootSIFT + PCA + SQRT \text{\cite{BTJ15}}         &  80  &  17.51 \\
  \baseline + PCA + SQRT \text{\cite{BTJ15}}        &  128 &  8.30  \\
  \baseline + \xya + \pcawt  ~(our)                &  128 & {\bf 6.79} \\
  \baseline + \xya + \pcaws  ~(our)               &  128 & 7.21 \\
\midrule[\heavyrulewidth]
\bottomrule

\end{tabular}

\end{tabular}

\caption{Performance comparison with the state of the art on Phototourism dataset. We report FPR@95 averaged over 6 dataset combinations for supervised (left) and unsupervised (right) approaches. The whitening for our descriptor is learned on the corresponding training part of PT for each combination.
}\label{tab:ptbestscores}
\end{table*}

%% file: tables/hpbest_bars.tex
\newcommand{\ttb}[1]{{\scshape #1}\xspace}
\newcommand{\meanstd}{{\ttb{MStd}}\xspace}
\newcommand{\resize}{{\ttb{Resz}}\xspace}
\newcommand{\sift}{{\ttb{SIFT}}\xspace}
\newcommand{\rootsift}{{\ttb{RSIFT}}\xspace}
\newcommand{\brief}{{\ttb{BRIEF}}\xspace}
\newcommand{\binboost}{{\ttb{BBoost}}\xspace}
\newcommand{\orb}{{\ttb{ORB}}\xspace}
\newcommand{\dcsiam}{{\ttb{DC-S}}\xspace}
\newcommand{\dcsiamts}{{\ttb{DC-S2S}}\xspace}
\newcommand{\deepdesc}{{\ttb{DDesc}}\xspace}
\newcommand{\tfmargin}{{\ttb{TF-M}}\xspace}
\newcommand{\tfratio}{{\ttb{TF-R}}\xspace}
\newcommand{\psift}{{\ttb{+SIFT}}\xspace}
\newcommand{\prootsift}{{\ttb{+RSIFT}}\xspace}
\newcommand{\pdcsiam}{{\ttb{+DC-S}}\xspace}
\newcommand{\pdcsiamts}{{\ttb{+DC-S2S}}\xspace}
\newcommand{\pdeepdesc}{{\ttb{+DDesc}}\xspace}
\newcommand{\ptfmargin}{{\ttb{+TF-M}}\xspace}
\newcommand{\ptfratio}{{\ttb{+TF-R}}\xspace}
\newcommand{\hardnet}{{\ttb{HardNet+}}\xspace}
\newcommand{\lnet}{{\ttb{L2Net+}}\xspace}
\newcommand{\chance}{{\ttb{Chance}}\xspace}

\newcommand{\easy}{\textsc{Easy}\xspace}
\newcommand{\hard}{\textsc{Hard}\xspace}
\newcommand{\tough}{\textsc{Tough}\xspace}
\newcommand{\viewpoint}{\textsc{Viewpt}\xspace}
\newcommand{\illum}{\textsc{Illum}\xspace}
\newcommand{\sameseq}{\textsc{SameSeq}\xspace}
\newcommand{\diffseq}{\textsc{DiffSeq}\xspace}
\newcommand{\degrees}{\ensuremath{^{\circ}}\xspace}
\newcommand{\dataset}{HPatches\xspace}

\definecolor{mycolor1}{rgb}{0.82031,0.41016,0.11719}%
\definecolor{mycolor2}{rgb}{0.00000,0.53125,0.21484}%
\definecolor{mycolor3}{rgb}{0.36719,0.23438,0.59766}%
\definecolor{mycolor4}{rgb}{0.78906,0.00000,0.12500}%
\definecolor{mycolor5}{rgb}{1.00000,0.62500,0.47656}%
\definecolor{mycolor6}{rgb}{0.62500,0.32031,0.17578}%
\definecolor{mycolor7}{rgb}{0.86719,0.71875,0.52734}%
\definecolor{mycolor8}{rgb}{0.97917,0.50000,0.44531}%
\definecolor{mycolor9}{rgb}{0.17969,0.54297,0.33984}%
\definecolor{mycolor10}{rgb}{0.27344,0.50781,0.70312}%
\definecolor{mycolor11}{rgb}{0.52734,0.80469,0.91797}%
\definecolor{mycolor12}{rgb}{0.00000,0.54297,0.54297}%
\definecolor{mycolor13}{rgb}{0.43750,0.50000,0.56250}%
\definecolor{mycolor14}{rgb}{0.43750,0.50000,0.56250}%
\definecolor{mycolor15}{rgb}{0.43750,0.50000,0.56250}%
\definecolor{mycolor16}{rgb}{0.43750,0.50000,0.56250}%
\definecolor{mycolor17}{rgb}{0.43750,0.50000,0.56250}%
\definecolor{mycolor18}{rgb}{0.43750,0.50000,0.56250}%
\definecolor{mycolor19}{rgb}{0.43750,0.50000,0.56250}%
\definecolor{mycolor20}{rgb}{0.43750,0.50000,0.56250}%

\newlength\figH
\newlength\figW

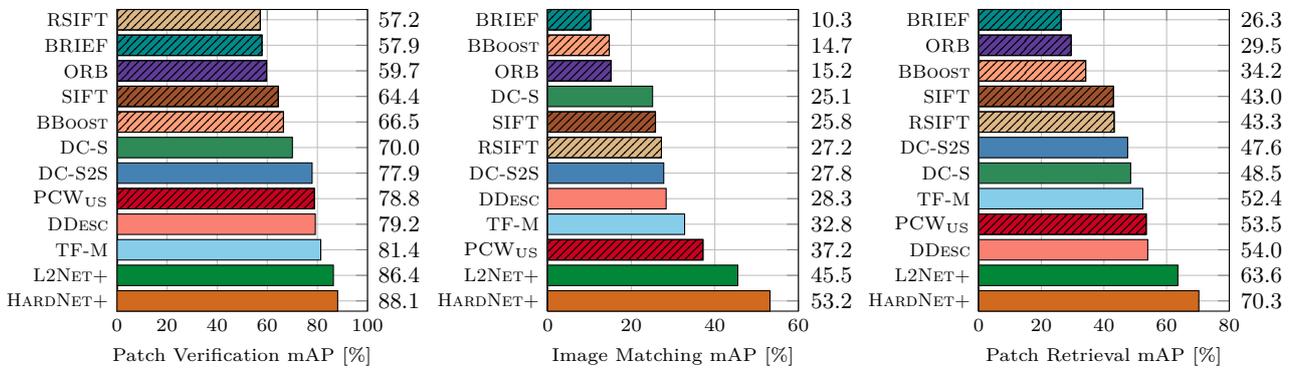
\begin{figure*}
  % \vspace{-1em}
  \centering \footnotesize   \hspace{2.5em}
  % \vspace{-0.5em}
\begin{tikzpicture}
\definecolor{easy}{rgb}{ 0, 0.5312, 0.2148}%
\definecolor{hard}{rgb}{0.3672,0.2344,0.5977}%
\definecolor{tough}{rgb}{0.7891, 0, 0.1250}%
\end{tikzpicture}\\
  \setlength{\figH}{4cm}
  \setlength{\figW}{0.42\columnwidth}
  \begin{minipage}{0.32\linewidth} \centering \hspace{2em}
    \vspace{-0.5em}
    \input{tables/table_verif}
  \end{minipage}
  \begin{minipage}{0.32\linewidth} \centering \hspace{2em}
    \vspace{-0.5em}
    \input{tables/table_match}
  \end{minipage}
  \begin{minipage}{0.32\linewidth} \centering  \hspace{2em}
    \vspace{-0.5em}
    \input{tables/table_retrv}
  \end{minipage}
\vspace{-15pt}
    \caption{Performance comparison on HP benchmark. The learning, whenever applicable, is performed on Liberty of PT dataset.
    Descriptors that do not require any supervision in the form of labeled patches, \ie hand-crafted or unsupervised, are
    shown in striped bars. Our descriptor is denoted by PC\pcaws (P=\baseline, C=\xya).\label{fig:hp_bars}}
\end{figure*}

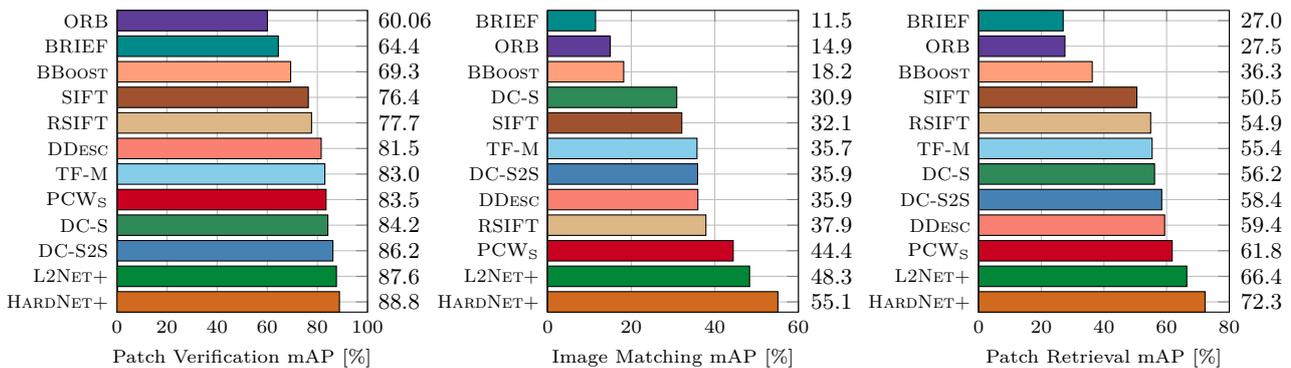
\begin{figure*}
  % \vspace{-1em}
  \centering \footnotesize   \hspace{2.5em}
  % \vspace{-0.5em}
\begin{tikzpicture}
\definecolor{easy}{rgb}{ 0, 0.5312, 0.2148}%
\definecolor{hard}{rgb}{0.3672,0.2344,0.5977}%
\definecolor{tough}{rgb}{0.7891, 0, 0.1250}%
\end{tikzpicture}\\
  \setlength{\figH}{4cm}
  \setlength{\figW}{0.42\columnwidth}
  \begin{minipage}{0.32\linewidth} \centering \hspace{2em}
    \vspace{-0.5em}
    \input{tables/table_verif_proj}
  \end{minipage}
  \begin{minipage}{0.32\linewidth} \centering \hspace{2em}
    \vspace{-0.5em}
    \input{tables/table_match_proj}
  \end{minipage}
  \begin{minipage}{0.32\linewidth} \centering  \hspace{2em}
    \vspace{-0.5em}
    \input{tables/table_retrv_proj}
  \end{minipage}
  \vspace{-15pt}
    \caption{Performance comparison on HP benchmark when post-processing all descriptors with supervised whitening \lw which is learned on HP. The initial learning of the descriptor, whenever applicable, is performed on Liberty of PT dataset.
    Our descriptor uses the whitening learned on HP and does not use the PT dataset at all.
    Our descriptor is denoted by PC\lw (P=\baseline, C=\xya).
    \label{fig:hp_bars_proj}}
\end{figure*}

%% file: tables/table_verif.tex
\begin{tikzpicture}
\begin{axis}[
 width=0.935\figW,
 height=\figH,
 at={(0\figW,0\figH)},
 scale only axis,
 clip=false,
 xmin=0,
 xmax=100,
 xlabel={Patch Verification mAP [\%]},
 y dir=reverse,
 ymin=0.6,
 ymax=12.4,
 ytick={1, 2, 3, 4, 5, 6, 7, 8, 9, 10, 11, 12},
 yticklabels={\rootsift, \brief, \orb, \sift, \binboost, \dcsiam, \dcsiamts, PC\pcaws, \deepdesc, \tfmargin, \lnet, \hardnet},
 axis background/.style={fill=white},
 xmajorgrids,
 ymajorgrids
 ]

\addplot[xbar, bar width=0.8, fill=mycolor7, draw=black, postaction={pattern=north east lines}, area legend] table[row sep=crcr] {%
57.211 1\\
};
\addplot[forget plot, color=white!15!black] table[row sep=crcr] {%
0	0.6\\
0	12.4\\
};
\node[right, align=left]
at (axis cs:101,1) {57.2};

\addplot[xbar, bar width=0.8, fill=mycolor12, draw=black, postaction={pattern=north east lines}, area legend] table[row sep=crcr] {%
57.899  2\\
};
\addplot[forget plot, color=white!15!black] table[row sep=crcr] {%
0	0.6\\
0	12.4\\
};
\node[right, align=left]
at (axis cs:101,2) {57.9};

\addplot[xbar, bar width=0.8, fill=mycolor3, draw=black, postaction={pattern=north east lines}, area legend] table[row sep=crcr] {%
59.744  3\\
};
\addplot[forget plot, color=white!15!black] table[row sep=crcr] {%
0	0.6\\
0	12.4\\
};
\node[right, align=left]
at (axis cs:101,3) {59.7};

\addplot[xbar, bar width=0.8, fill=mycolor6, draw=black, postaction={pattern=north east lines}, area legend] table[row sep=crcr] {%
64.386  4\\
};
\addplot[forget plot, color=white!15!black] table[row sep=crcr] {%
0	0.6\\
0	12.4\\
};
\node[right, align=left]
at (axis cs:101,4) {64.4};

\addplot[xbar, bar width=0.8, fill=mycolor5, draw=black, postaction={pattern=north east lines}, area legend] table[row sep=crcr] {%
66.462  5\\
};
\addplot[forget plot, color=white!15!black] table[row sep=crcr] {%
0	0.6\\
0	12.4\\
};
\node[right, align=left]
at (axis cs:101,5) {66.5};

\addplot[xbar, bar width=0.8, fill=mycolor9, draw=black, area legend] table[row sep=crcr] {%
70.027  6\\
};
\addplot[forget plot, color=white!15!black] table[row sep=crcr] {%
0	0.6\\
0	12.4\\
};
\node[right, align=left]
at (axis cs:101,6) {70.0};

\addplot[xbar, bar width=0.8, fill=mycolor10, draw=black, area legend] table[row sep=crcr] {%
77.900  7\\
};
\addplot[forget plot, color=white!15!black] table[row sep=crcr] {%
0	0.6\\
0	12.4\\
};
\node[right, align=left]
at (axis cs:101,7) {77.9};

\addplot[xbar, bar width=0.8, fill=mycolor4, draw=black, postaction={pattern=north east lines}, area legend] table[row sep=crcr] {%
78.81  8\\
};
\addplot[forget plot, color=white!15!black] table[row sep=crcr] {%
0	0.6\\
0	12.4\\
};
\node[right, align=left]
at (axis cs:101,8) {78.8};

\addplot[xbar, bar width=0.8, fill=mycolor8, draw=black, area legend] table[row sep=crcr] {%
79.178  9\\
};
\addplot[forget plot, color=white!15!black] table[row sep=crcr] {%
0	0.6\\
0	12.4\\
};
\node[right, align=left]
at (axis cs:101,9) {79.2};

\addplot[xbar, bar width=0.8, fill=mycolor11, draw=black, area legend] table[row sep=crcr] {%
81.358  10\\
};
\addplot[forget plot, color=white!15!black] table[row sep=crcr] {%
0	0.6\\
0	12.4\\
};
\node[right, align=left]
at (axis cs:101,10) {81.4};

\addplot[xbar, bar width=0.8, fill=mycolor2, draw=black, area legend] table[row sep=crcr] {%
86.381  11\\
};
\addplot[forget plot, color=white!15!black] table[row sep=crcr] {%
0	0.6\\
0	12.4\\
};
\node[right, align=left]
at (axis cs:101,11) {86.4};

\addplot[xbar, bar width=0.8, fill=mycolor1, draw=black, area legend] table[row sep=crcr] {%
88.092  12\\
};
\addplot[forget plot, color=white!15!black] table[row sep=crcr] {%
0	0.6\\
0	12.4\\
};
\node[right, align=left]
at (axis cs:101,12) {88.1};
]

\end{axis}
\end{tikzpicture}

%% file: tables/table_match.tex
\begin{tikzpicture}
\begin{axis}[
 width=0.935\figW,
 height=\figH,
 at={(0\figW,0\figH)},
 scale only axis,
 clip=false,
 xmin=0,
 xmax=60,
 xlabel={Image Matching mAP [\%]},
 y dir=reverse,
 ymin=0.6,
 ymax=12.4,
 ytick={1, 2, 3, 4, 5, 6, 7, 8, 9, 10, 11, 12},
 yticklabels={\brief, \binboost, \orb, \dcsiam, \sift, \rootsift, \dcsiamts, \deepdesc, \tfmargin, PC\pcaws, \lnet, \hardnet},
 axis background/.style={fill=white},
 xmajorgrids,
 ymajorgrids
 ]
  \addplot[xbar, bar width=0.8, fill=mycolor12, draw=black, postaction={pattern=north east lines}, area legend] table[row sep=crcr] {%
10.288200  1\\
};
\addplot[forget plot, color=white!15!black] table[row sep=crcr] {%
0	0.6\\
0	12.4\\
};
\node[right, align=left]
at (axis cs:61,1) {10.3};
\addplot[xbar, bar width=0.8, fill=mycolor5, draw=black, postaction={pattern=north east lines}, area legend] table[row sep=crcr] {%
14.722800  2\\
};
\addplot[forget plot, color=white!15!black] table[row sep=crcr] {%
0	0.6\\
0	12.4\\
};
\node[right, align=left]
at (axis cs:61,2) {14.7};
\addplot[xbar, bar width=0.8, fill=mycolor3, draw=black, postaction={pattern=north east lines}, area legend] table[row sep=crcr] {%
15.164300  3\\
};
\addplot[forget plot, color=white!15!black] table[row sep=crcr] {%
0	0.6\\
0	12.4\\
};
\node[right, align=left]
at (axis cs:61,3) {15.2};
\addplot[xbar, bar width=0.8, fill=mycolor9, draw=black, area legend] table[row sep=crcr] {%
25.098800  4\\
};
\addplot[forget plot, color=white!15!black] table[row sep=crcr] {%
0	0.6\\
0	12.4\\
};
\node[right, align=left]
at (axis cs:61,4) {25.1};
\addplot[xbar, bar width=0.8, fill=mycolor6, draw=black, postaction={pattern=north east lines}, area legend] table[row sep=crcr] {%
25.780300  5\\
};
\addplot[forget plot, color=white!15!black] table[row sep=crcr] {%
0	0.6\\
0	12.4\\
};
\node[right, align=left]
at (axis cs:61,5) {25.8};
\addplot[xbar, bar width=0.8, fill=mycolor7, draw=black, postaction={pattern=north east lines}, area legend] table[row sep=crcr] {%
27.212400  6\\
};
\addplot[forget plot, color=white!15!black] table[row sep=crcr] {%
0	0.6\\
0	12.4\\
};
\node[right, align=left]
at (axis cs:61,6) {27.2};
\addplot[xbar, bar width=0.8, fill=mycolor10, draw=black, area legend] table[row sep=crcr] {%
27.763900  7\\
};
\addplot[forget plot, color=white!15!black] table[row sep=crcr] {%
0	0.6\\
0	12.4\\
};
\node[right, align=left]
at (axis cs:61,7) {27.8};
\addplot[xbar, bar width=0.8, fill=mycolor8, draw=black, area legend] table[row sep=crcr] {%
28.340800  8\\
};
\addplot[forget plot, color=white!15!black] table[row sep=crcr] {%
0	0.6\\
0	12.4\\
};
\node[right, align=left]
at (axis cs:61,8) {28.3};
\addplot[xbar, bar width=0.8, fill=mycolor11, draw=black, area legend] table[row sep=crcr] {%
32.761500  9\\
};
\addplot[forget plot, color=white!15!black] table[row sep=crcr] {%
0	0.6\\
0	12.4\\
};
\node[right, align=left]
at (axis cs:61,9) {32.8};
\addplot[xbar, bar width=0.8, fill=mycolor4, draw=black, postaction={pattern=north east lines}, area legend] table[row sep=crcr] {%
37.160000  10\\
};
\addplot[forget plot, color=white!15!black] table[row sep=crcr] {%
0	0.6\\
0	12.4\\
};
\node[right, align=left]
at (axis cs:61,10) {37.2};
\addplot[xbar, bar width=0.8, fill=mycolor2, draw=black, area legend] table[row sep=crcr] {%
45.491100  11\\
};
\addplot[forget plot, color=white!15!black] table[row sep=crcr] {%
0	0.6\\
0	12.4\\
};
\node[right, align=left]
at (axis cs:61,11) {45.5};
\addplot[xbar, bar width=0.8, fill=mycolor1, draw=black, area legend] table[row sep=crcr] {%
53.198800  12\\
};
\addplot[forget plot, color=white!15!black] table[row sep=crcr] {%
0	0.6\\
0	12.4\\
};
\node[right, align=left]
at (axis cs:61,12) {53.2};
]
\end{axis}
\end{tikzpicture}

%% file: tables/table_retrv.tex
\begin{tikzpicture}
\begin{axis}[
 width=0.935\figW,
 height=\figH,
 at={(0\figW,0\figH)},
 scale only axis,
 clip=false,
 xmin=0,
 xmax=80,
 xlabel={Patch Retrieval mAP [\%]},
 y dir=reverse,
 ymin=0.6,
 ymax=12.4,
 ytick={1, 2, 3, 4, 5, 6, 7, 8, 9, 10, 11, 12},
 yticklabels={\brief, \orb, \binboost, \sift, \rootsift, \dcsiamts, \dcsiam, \tfmargin, PC\pcaws, \deepdesc, \lnet, \hardnet},
 axis background/.style={fill=white},
 xmajorgrids,
 ymajorgrids
 ]
\addplot[xbar, bar width=0.8, fill=mycolor12, draw=black, postaction={pattern=north east lines}, area legend] table[row sep=crcr] {%
26.317700  1\\
};
\addplot[forget plot, color=white!15!black] table[row sep=crcr] {%
0	0.6\\
0	12.4\\
};
\node[right, align=left]
at (axis cs:81,1) {26.3};
\addplot[xbar, bar width=0.8, fill=mycolor3, draw=black, postaction={pattern=north east lines}, area legend] table[row sep=crcr] {%
29.518600  2\\
};
\addplot[forget plot, color=white!15!black] table[row sep=crcr] {%
0	0.6\\
0	12.4\\
};
\node[right, align=left]
at (axis cs:81,2) {29.5};
\addplot[xbar, bar width=0.8, fill=mycolor5, draw=black, postaction={pattern=north east lines}, area legend] table[row sep=crcr] {%
34.180100  3\\
};
\addplot[forget plot, color=white!15!black] table[row sep=crcr] {%
0	0.6\\
0	12.4\\
};
\node[right, align=left]
at (axis cs:81,3) {34.2};
\addplot[xbar, bar width=0.8, fill=mycolor6, draw=black, postaction={pattern=north east lines}, area legend] table[row sep=crcr] {%
43.040300  4\\
};
\addplot[forget plot, color=white!15!black] table[row sep=crcr] {%
0	0.6\\
0	12.4\\
};
\node[right, align=left]
at (axis cs:81,4) {43.0};
\addplot[xbar, bar width=0.8, fill=mycolor7, draw=black, postaction={pattern=north east lines}, area legend] table[row sep=crcr] {%
43.264200  5\\
};
\addplot[forget plot, color=white!15!black] table[row sep=crcr] {%
0	0.6\\
0	12.4\\
};
\node[right, align=left]
at (axis cs:81,5) {43.3};
\addplot[xbar, bar width=0.8, fill=mycolor10, draw=black, area legend] table[row sep=crcr] {%
47.566500  6\\
};
\addplot[forget plot, color=white!15!black] table[row sep=crcr] {%
0	0.6\\
0	12.4\\
};
\node[right, align=left]
at (axis cs:81,6) {47.6};
\addplot[xbar, bar width=0.8, fill=mycolor9, draw=black, area legend] table[row sep=crcr] {%
48.528400  7\\
};
\addplot[forget plot, color=white!15!black] table[row sep=crcr] {%
0	0.6\\
0	12.4\\
};
\node[right, align=left]
at (axis cs:81,7) {48.5};
\addplot[xbar, bar width=0.8, fill=mycolor11, draw=black, area legend] table[row sep=crcr] {%
52.413000  8\\
};
\addplot[forget plot, color=white!15!black] table[row sep=crcr] {%
0	0.6\\
0	12.4\\
};
\node[right, align=left]
at (axis cs:81,8) {52.4};
\addplot[xbar, bar width=0.8, fill=mycolor4, draw=black, postaction={pattern=north east lines}, area legend] table[row sep=crcr] {%
53.500000  9\\
};
\addplot[forget plot, color=white!15!black] table[row sep=crcr] {%
0	0.6\\
0	12.4\\
};
\node[right, align=left]
at (axis cs:81,9) {53.5};
\addplot[xbar, bar width=0.8, fill=mycolor8, draw=black, area legend] table[row sep=crcr] {%
53.981000  10\\
};
\addplot[forget plot, color=white!15!black] table[row sep=crcr] {%
0	0.6\\
0	12.4\\
};
\node[right, align=left]
at (axis cs:81,10) {54.0};
\addplot[xbar, bar width=0.8, fill=mycolor2, draw=black, area legend] table[row sep=crcr] {%
63.587500  11\\
};
\addplot[forget plot, color=white!15!black] table[row sep=crcr] {%
0	0.6\\
0	12.4\\
};
\node[right, align=left]
at (axis cs:81,11) {63.6};
\addplot[xbar, bar width=0.8, fill=mycolor1, draw=black, area legend] table[row sep=crcr] {%
70.288800  12\\
};
\addplot[forget plot, color=white!15!black] table[row sep=crcr] {%
0	0.6\\
0	12.4\\
};
\node[right, align=left]
at (axis cs:81,12) {70.3};
]
\end{axis}
\end{tikzpicture}

%% file: tables/table_verif_proj.tex
\begin{tikzpicture}
\begin{axis}[
 width=0.935\figW,
 height=\figH,
 at={(0\figW,0\figH)},
 scale only axis,
 clip=false,
 xmin=0,
 xmax=100,
 xlabel={Patch Verification mAP [\%]},
 y dir=reverse,
 ymin=0.6,
 ymax=12.4,
 ytick={1, 2, 3, 4, 5, 6, 7, 8, 9, 10, 11, 12},
 yticklabels={\orb, \brief, \binboost, \sift, \rootsift, \deepdesc, \tfmargin, PC\lw, \dcsiam, \dcsiamts, \lnet, \hardnet},
 axis background/.style={fill=white},
 xmajorgrids,
 ymajorgrids
 ]

\addplot[xbar, bar width=0.8, fill=mycolor3, draw=black, area legend] table[row sep=crcr] {%
60.058  1\\
};
\addplot[forget plot, color=white!15!black] table[row sep=crcr] {%
0	0.6\\
0	12.4\\
};
\node[right, align=left]
at (axis cs:101,1) {60.06};

\addplot[xbar, bar width=0.8, fill=mycolor12, draw=black, area legend] table[row sep=crcr] {%
64.421  2\\
};
\addplot[forget plot, color=white!15!black] table[row sep=crcr] {%
0	0.6\\
0	12.4\\
};
\node[right, align=left]
at (axis cs:101,2) {64.4};

\addplot[xbar, bar width=0.8, fill=mycolor5, draw=black, area legend] table[row sep=crcr] {%
69.340  3\\
};
\addplot[forget plot, color=white!15!black] table[row sep=crcr] {%
0	0.6\\
0	12.4\\
};
\node[right, align=left]
at (axis cs:101,3) {69.3};

\addplot[xbar, bar width=0.8, fill=mycolor6, draw=black, area legend] table[row sep=crcr] {%
76.386  4\\
};
\addplot[forget plot, color=white!15!black] table[row sep=crcr] {%
0	0.6\\
0	12.4\\
};
\node[right, align=left]
at (axis cs:101,4) {76.4};

\addplot[xbar, bar width=0.8, fill=mycolor7, draw=black, area legend] table[row sep=crcr] {%
77.673  5\\
};
\addplot[forget plot, color=white!15!black] table[row sep=crcr] {%
0	0.6\\
0	12.4\\
};
\node[right, align=left]
at (axis cs:101,5) {77.7};

\addplot[xbar, bar width=0.8, fill=mycolor8, draw=black, area legend] table[row sep=crcr] {%
81.528  6\\
};
\addplot[forget plot, color=white!15!black] table[row sep=crcr] {%
0	0.6\\
0	12.4\\
};
\node[right, align=left]
at (axis cs:101,6) {81.5};

\addplot[xbar, bar width=0.8, fill=mycolor11, draw=black, area legend] table[row sep=crcr] {%
82.979  7\\
};
\addplot[forget plot, color=white!15!black] table[row sep=crcr] {%
0	0.6\\
0	12.4\\
};
\node[right, align=left]
at (axis cs:101,7) {83.0};

\addplot[xbar, bar width=0.8, fill=mycolor4, draw=black, area legend] table[row sep=crcr] {%
83.48  8\\
};
\addplot[forget plot, color=white!15!black] table[row sep=crcr] {%
0	0.6\\
0	12.4\\
};
\node[right, align=left]
at (axis cs:101,8) {83.5};

\addplot[xbar, bar width=0.8, fill=mycolor9, draw=black, area legend] table[row sep=crcr] {%
84.175  9\\
};
\addplot[forget plot, color=white!15!black] table[row sep=crcr] {%
0	0.6\\
0	12.4\\
};
\node[right, align=left]
at (axis cs:101,9) {84.2};

\addplot[xbar, bar width=0.8, fill=mycolor10, draw=black, area legend] table[row sep=crcr] {%
86.192  10\\
};
\addplot[forget plot, color=white!15!black] table[row sep=crcr] {%
0	0.6\\
0	12.4\\
};
\node[right, align=left]
at (axis cs:101,10) {86.2};

\addplot[xbar, bar width=0.8, fill=mycolor2, draw=black, area legend] table[row sep=crcr] {%
87.598  11\\
};
\addplot[forget plot, color=white!15!black] table[row sep=crcr] {%
0	0.6\\
0	12.4\\
};
\node[right, align=left]
at (axis cs:101,11) {87.6};

\addplot[xbar, bar width=0.8, fill=mycolor1, draw=black, area legend] table[row sep=crcr] {%
88.801  12\\
};
\addplot[forget plot, color=white!15!black] table[row sep=crcr] {%
0	0.6\\
0	12.4\\
};
\node[right, align=left]
at (axis cs:101,12) {88.8};

]
\end{axis}
\end{tikzpicture}

%% file: tables/table_match_proj.tex
\begin{tikzpicture}
\begin{axis}[
 width=0.935\figW,
 height=\figH,
 at={(0\figW,0\figH)},
 scale only axis,
 clip=false,
 xmin=0,
 xmax=60,
 xlabel={Image Matching mAP [\%]},
 y dir=reverse,
 ymin=0.6,
 ymax=12.4,
 ytick={1, 2, 3, 4, 5, 6, 7, 8, 9, 10, 11, 12},
 yticklabels={\brief, \orb, \binboost, \dcsiam, \sift, \tfmargin, \dcsiamts, \deepdesc, \rootsift, PC\lw, \lnet, \hardnet},
 axis background/.style={fill=white},
 xmajorgrids,
 ymajorgrids
 ]
\addplot[xbar, bar width=0.8, fill=mycolor12, draw=black, area legend] table[row sep=crcr] {%
11.450000  1\\
};
\addplot[forget plot, color=white!15!black] table[row sep=crcr] {%
0	0.6\\
0	12.4\\
};
\node[right, align=left]
at (axis cs:61,1) {11.5};
\addplot[xbar, bar width=0.8, fill=mycolor3, draw=black, area legend] table[row sep=crcr] {%
14.950000  2\\
};
\addplot[forget plot, color=white!15!black] table[row sep=crcr] {%
0	0.6\\
0	12.4\\
};
\node[right, align=left]
at (axis cs:61,2) {14.9};
\addplot[xbar, bar width=0.8, fill=mycolor5, draw=black, area legend] table[row sep=crcr] {%
18.180000  3\\
};
\addplot[forget plot, color=white!15!black] table[row sep=crcr] {%
0	0.6\\
0	12.4\\
};
\node[right, align=left]
at (axis cs:61,3) {18.2};
\addplot[xbar, bar width=0.8, fill=mycolor9, draw=black, area legend] table[row sep=crcr] {%
30.880000  4\\
};
\addplot[forget plot, color=white!15!black] table[row sep=crcr] {%
0	0.6\\
0	12.4\\
};
\node[right, align=left]
at (axis cs:61,4) {30.9};
\addplot[xbar, bar width=0.8, fill=mycolor6, draw=black, area legend] table[row sep=crcr] {%
32.090000  5\\
};
\addplot[forget plot, color=white!15!black] table[row sep=crcr] {%
0	0.6\\
0	12.4\\
};
\node[right, align=left]
at (axis cs:61,5) {32.1};
\addplot[xbar, bar width=0.8, fill=mycolor11, draw=black, area legend] table[row sep=crcr] {%
35.730000  6\\
};
\addplot[forget plot, color=white!15!black] table[row sep=crcr] {%
0	0.6\\
0	12.4\\
};
\node[right, align=left]
at (axis cs:61,6) {35.7};
\addplot[xbar, bar width=0.8, fill=mycolor10, draw=black, area legend] table[row sep=crcr] {%
35.870000  7\\
};
\addplot[forget plot, color=white!15!black] table[row sep=crcr] {%
0	0.6\\
0	12.4\\
};
\node[right, align=left]
at (axis cs:61,7) {35.9};
\addplot[xbar, bar width=0.8, fill=mycolor8, draw=black, area legend] table[row sep=crcr] {%
35.940000  8\\
};
\addplot[forget plot, color=white!15!black] table[row sep=crcr] {%
0	0.6\\
0	12.4\\
};
\node[right, align=left]
at (axis cs:61,8) {35.9};
\addplot[xbar, bar width=0.8, fill=mycolor7, draw=black, area legend] table[row sep=crcr] {%
37.860000  9\\
};
\addplot[forget plot, color=white!15!black] table[row sep=crcr] {%
0	0.6\\
0	12.4\\
};
\node[right, align=left]
at (axis cs:61,9) {37.9};
\addplot[xbar, bar width=0.8, fill=mycolor4, draw=black, area legend] table[row sep=crcr] {%
44.400000  10\\
};
\addplot[forget plot, color=white!15!black] table[row sep=crcr] {%
0	0.6\\
0	12.4\\
};
\node[right, align=left]
at (axis cs:61,10) {44.4};
\addplot[xbar, bar width=0.8, fill=mycolor2, draw=black, area legend] table[row sep=crcr] {%
48.310000  11\\
};
\addplot[forget plot, color=white!15!black] table[row sep=crcr] {%
0	0.6\\
0	12.4\\
};
\node[right, align=left]
at (axis cs:61,11) {48.3};
\addplot[xbar, bar width=0.8, fill=mycolor1, draw=black, area legend] table[row sep=crcr] {%
55.090000  12\\
};
\addplot[forget plot, color=white!15!black] table[row sep=crcr] {%
0	0.6\\
0	12.4\\
};
\node[right, align=left]
at (axis cs:61,12) {55.1};
]
\end{axis}
\end{tikzpicture}

%% file: tables/table_retrv_proj.tex
\begin{tikzpicture}
\begin{axis}[
 width=0.935\figW,
 height=\figH,
 at={(0\figW,0\figH)},
 scale only axis,
 clip=false,
 xmin=0,
 xmax=80,
 xlabel={Patch Retrieval mAP [\%]},
 y dir=reverse,
 ymin=0.6,
 ymax=12.4,
 ytick={1, 2, 3, 4, 5, 6, 7, 8, 9, 10, 11, 12},
 yticklabels={\brief, \orb, \binboost, \sift, \rootsift, \tfmargin, \dcsiam, \dcsiamts, \deepdesc, PC\lw, \lnet, \hardnet},
 axis background/.style={fill=white},
 xmajorgrids,
 ymajorgrids
 ]
\addplot[xbar, bar width=0.8, fill=mycolor12, draw=black, area legend] table[row sep=crcr] {%
26.990000  1\\
};
\addplot[forget plot, color=white!15!black] table[row sep=crcr] {%
0	0.6\\
0	12.4\\
};
\node[right, align=left]
at (axis cs:81,1) {27.0};
\addplot[xbar, bar width=0.8, fill=mycolor3, draw=black, area legend] table[row sep=crcr] {%
27.540000  2\\
};
\addplot[forget plot, color=white!15!black] table[row sep=crcr] {%
0	0.6\\
0	12.4\\
};
\node[right, align=left]
at (axis cs:81,2) {27.5};
\addplot[xbar, bar width=0.8, fill=mycolor5, draw=black, area legend] table[row sep=crcr] {%
36.260000  3\\
};
\addplot[forget plot, color=white!15!black] table[row sep=crcr] {%
0	0.6\\
0	12.4\\
};
\node[right, align=left]
at (axis cs:81,3) {36.3};
\addplot[xbar, bar width=0.8, fill=mycolor6, draw=black, area legend] table[row sep=crcr] {%
50.510000  4\\
};
\addplot[forget plot, color=white!15!black] table[row sep=crcr] {%
0	0.6\\
0	12.4\\
};
\node[right, align=left]
at (axis cs:81,4) {50.5};
\addplot[xbar, bar width=0.8, fill=mycolor7, draw=black, area legend] table[row sep=crcr] {%
54.910000  5\\
};
\addplot[forget plot, color=white!15!black] table[row sep=crcr] {%
0	0.6\\
0	12.4\\
};
\node[right, align=left]
at (axis cs:81,5) {54.9};
\addplot[xbar, bar width=0.8, fill=mycolor11, draw=black, area legend] table[row sep=crcr] {%
55.380000  6\\
};
\addplot[forget plot, color=white!15!black] table[row sep=crcr] {%
0	0.6\\
0	12.4\\
};
\node[right, align=left]
at (axis cs:81,6) {55.4};
\addplot[xbar, bar width=0.8, fill=mycolor9, draw=black, area legend] table[row sep=crcr] {%
56.150000  7\\
};
\addplot[forget plot, color=white!15!black] table[row sep=crcr] {%
0	0.6\\
0	12.4\\
};
\node[right, align=left]
at (axis cs:81,7) {56.2};
\addplot[xbar, bar width=0.8, fill=mycolor10, draw=black, area legend] table[row sep=crcr] {%
58.440000  8\\
};
\addplot[forget plot, color=white!15!black] table[row sep=crcr] {%
0	0.6\\
0	12.4\\
};
\node[right, align=left]
at (axis cs:81,8) {58.4};
\addplot[xbar, bar width=0.8, fill=mycolor8, draw=black, area legend] table[row sep=crcr] {%
59.370000  9\\
};
\addplot[forget plot, color=white!15!black] table[row sep=crcr] {%
0	0.6\\
0	12.4\\
};
\node[right, align=left]
at (axis cs:81,9) {59.4};
\addplot[xbar, bar width=0.8, fill=mycolor4, draw=black, area legend] table[row sep=crcr] {%
61.790000  10\\
};
\addplot[forget plot, color=white!15!black] table[row sep=crcr] {%
0	0.6\\
0	12.4\\
};
\node[right, align=left]
at (axis cs:81,10) {61.8};
\addplot[xbar, bar width=0.8, fill=mycolor2, draw=black, area legend] table[row sep=crcr] {%
66.440000  11\\
};
\addplot[forget plot, color=white!15!black] table[row sep=crcr] {%
0	0.6\\
0	12.4\\
};
\node[right, align=left]
at (axis cs:81,11) {66.4};
\addplot[xbar, bar width=0.8, fill=mycolor1, draw=black, area legend] table[row sep=crcr] {%
72.270000  12\\
};
\addplot[forget plot, color=white!15!black] table[row sep=crcr] {%
0	0.6\\
0	12.4\\
};
\node[right, align=left]
at (axis cs:81,12) {72.3};
]
\end{axis}
\end{tikzpicture}

%% file: nomixparam.tex
\begin{appendices}
\section{Regularized concatenation}
\label{sec:proof}
When combining the descriptors of different parametrization by concatenation
we use both with equal contribution, \ie the final similarity is equal to
$\kp\hspace{-1pt}\kr\hspace{-1pt}\ktt + \kx\hspace{-1pt}\ky\hspace{-1pt}\kt$.
In the case of the raw descriptors this is clearly suboptimal. One would 
rather regularize by $\kp\hspace{-1pt}\kr\hspace{-1pt}\ktt + w\kx\hspace{-1pt}\ky\hspace{-1pt}\kt$
and search for the optimal value of scalar $w$. We are about to prove
that this is not necessary in the case of post-processing by supervised whitening,
where the optimal regularization is included in the projection matrix.

We denote a set of descriptors without regularized concatenation by $V_\Pb$ when $w=1$, while
the $V_\Pb^{(w)}$ when $w \neq 1$. It holds that $V_\Pb^{(w)} = \{W V(\P),~\P \in \Pb\}$, where $W$ is a diagonal
matrix with ones on the dimensions corresponding to the first descriptors (for \kp\hspace{-1pt}\kr\hspace{-1pt}\ktt), and has all the rest elements 
of the diagonal equal to $w$. The covariance matrix of $V_\Pb$ is $C$, while of $V_\Pb^{(w)}$ it is
$C^{(w)} = W C W^\top$.

Learning the supervised whitening on $V_\Pb$ as in (\ref{eq:lw}) produces projection matrix
\begin{equation}
\label{eq:matrixa}
A = C_\Mb^{-\nicefrac{1}{2}}\mbox{eig}( C_\Mb^{-\nicefrac{1}{2}} C C_\Mb^{-\nicefrac{1}{2}}),
\end{equation}
while learning it on $V_\Pb^{(w)}$ produces projection matrix
\begin{equation}
\label{eq:matrixaw}
A^{(w)} = {C_\Mb^{(w)}}^{-\nicefrac{1}{2}}\mbox{eig}( {C_\Mb^{(w)}}^{-\nicefrac{1}{2}} C^{(w)} {C_\Mb^{(w)}}^{-\nicefrac{1}{2}}).
\end{equation}
Cholesky decomposition of $C$ gives
\begin{equation}
\label{eq:cholc}
C = U^\top U = L L^\top,
\end{equation}
which leads to the Cholesky decomposition
\begin{equation}
\label{eq:cholw}
C^{(w)} = W U^\top U W^\top = W L L^\top W^\top.
\end{equation}
Using (\ref{eq:cholw}) allows us to rewrite (\ref{eq:matrixaw}) as
\begin{equation}
\small
\begin{aligned}
% \label{eq:}
A^{(w)} &= {(L^\top W^\top)}^{-1}\mbox{eig}( {(W U^\top)}^{-1} W C W^\top {(L^\top W^\top)}^{-1})\\
        &= {(W^\top)}^{-1}{(L^\top)}^{-1} \mbox{eig}( {U^\top}^{-1} W^{-1} W C W^\top (W^\top)^{-1}{(L^\top)}^{-1})\\
        &= {(W^\top)}^{-1}{(L^\top)}^{-1} \mbox{eig}( {U^\top}^{-1} C {(L^\top)}^{-1})\\
        &= {(W^\top)}^{-1}{(L^\top)}^{-1} \mbox{eig}( C_\Mb^{-\nicefrac{1}{2}} C C_\Mb^{-\nicefrac{1}{2}})\\
        &= {(W^\top)}^{-1}A\\.
\end{aligned}
\end{equation}
Whitening descriptor $V(\P) \in V_\Pb$ with matrix $A$ is performed by
\begin{equation}
\hat{V}(\P) = A^\top (V(\P) - \mu), 
\end{equation}
while whitening descriptor ${V(\P)}^{(w)} \in V_\Pb^{(w)}$ with matrix $A^{(w)}$ is performed by
\begin{equation}
\begin{aligned}
{\hat{V}(\P)}^{(w)} &= {A^{(w)}}^\top (WV(\P) - W\mu)\\
                    &= A^\top {W}^{-1} (WV(\P) - W\mu)\\
                    &= \hat{V}(\P). 
\end{aligned}
\end{equation}
No matter what the regularization parameter is, the descriptor is identical after whitening. 
We conclude that there is no need to perform such regularized concatenation.
\end{appendices}